\documentclass[Afour,sageh,times]{sagej}

\usepackage{moreverb,url}
\usepackage{caption}
\usepackage{multirow}
\usepackage{adjustbox}
\usepackage{threeparttable}
\usepackage{colortbl}
\usepackage{color}
\usepackage{xcolor}
\usepackage[colorlinks,bookmarksopen,bookmarksnumbered,citecolor=red,urlcolor=red]{hyperref}
\definecolor{mygray}{gray}{.9}
\usepackage{float}
\usepackage{tabularx}
\usepackage{enumitem}
\newcommand\BibTeX{{\rmfamily B\kern-.05em \textsc{i\kern-.025em b}\kern-.08em
T\kern-.1667em\lower.7ex\hbox{E}\kern-.125emX}}
\usepackage[utf8]{inputenc}
\usepackage{booktabs}
\usepackage{tabularx}
\usepackage{geometry}
\usepackage{longtable}
\usepackage{graphicx}
\usepackage{float}
\usepackage{titlesec}

\def\volumeyear{2016}

\begin{document}

\runninghead{Foundation Models for Manipulation}

\title{What Foundation Models can Bring for Robot Learning in Manipulation : A Survey}

\author{Dingzhe Li\affilnum{1}, Yixiang Jin\affilnum{1}, YuHao Sun\affilnum{2}, Yong A\affilnum{1}, Hongze Yu\affilnum{1}, Jun Shi\affilnum{1}, Xiaoshuai Hao\affilnum{1}, Peng Hao\affilnum{1}, Huaping Liu\affilnum{3}, Xiang Li\affilnum{3}, Xinde Li\affilnum{4}, Fuchun Sun\affilnum{3}, Jianwei Zhang\affilnum{5}, Bin Fang\affilnum{2}}

\affiliation{\affilnum{1}Samsung R\&D Institute China-Beijing, China\\
\affilnum{2}Beijing University of Posts and Telecommunications, China\\
\affilnum{3}Tsinghua University, China\\
\affilnum{4}Southeast University, China\\
\affilnum{5}Universität Hamburg, Germany\\}
\corrauth{Bin Fang, Beijing University of Posts and Telecommunications, China.}

\email{\tt\small \{fangbin1120\}@bupt.edu.cn}

\begin{abstract}
The realization of universal robots is an ultimate goal of researchers. However, a key hurdle in achieving this goal lies in the robots’ ability to manipulate objects in their unstructured environments according to different tasks. The learning-based approach is considered an effective way to address generalization. The impressive performance of foundation models in the fields of computer vision and natural language suggests the potential of embedding foundation models into manipulation tasks as a viable path toward achieving general manipulation capability. However, we believe achieving general manipulation capability requires an overarching framework akin to auto driving. This framework should encompass multiple functional modules, with different foundation models assuming distinct roles in facilitating general manipulation capability. This survey focuses on the contributions of foundation models to robot learning for manipulation. We propose a comprehensive framework and detail how foundation models can address challenges in each module of the framework. What's more, we examine current approaches, outline challenges, suggest future research directions, and identify potential risks associated with integrating foundation models into this domain.
\end{abstract}


\maketitle

\section{Introduction}
Researchers aim to create universal robots that can seamlessly integrate into human life to boost productivity, much like those depicted in the movie `I, Robot'. However, a key hurdle in achieving this lies in the robots' ability to manipulate objects in their unstructured environments according to different tasks. There is abundant literature available for improving the general manipulation capability of robots, which can be roughly categorized into model-based and learning-based approaches (\cite{zarrin2023hybrid}). The real world is too diverse for universal robots and they must adapt to unstructured environments and arbitrary objects to manipulate effectively. Therefore, learning-based methods are crucial for manipulation tasks (\cite{kleeberger2020survey}).

The predominant methodologies in learning-based approaches are deep learning, reinforcement learning and imitation learning. Learning-based methods have spanned from acquiring specific manipulation skills through labeled datasets like human demonstration, to acquiring abstract representations of manipulation tasks conducive to high-level planning, to exploring an object's functionalities through interaction and encompassing various objectives in between (\cite{kroemer2021review}). However, challenges persist, including 1) unnatural interaction with humans; 2) high-cost data collection; 3) limited perceptual capability; 4) non-intelligent hierarchy of skills; 5) inaccurate pre- and post-conditions \& post-hoc correction; 6) unreliable skill learning; 7) poor environment transition (\cite{hu2023toward}).

Foundation models are primarily pretrained on vast internet-scale datasets, enabling them to be fine-tuned for diverse tasks. Their significant advancements in vision and language processing contribute to mitigating the aforementioned challenges. Based on \cite{firoozi2023foundation} and considering the different input modalities and functionalities of the models, we categorize foundation models into the following six types.
\begin{enumerate}
    \item \textbf{Large Language Models (LLMs)} like BERT (\cite{devlin2018bert}), GPT-3 (\cite{brown2020language}) demonstrate the capability to generate coherent chains of thought. 
    \item \textbf{Visual Foundation Models (VFMs)} like SAM (\cite{kirillov2023segment}) demonstrate strong segmentation capability for open-set objects.
    \item \textbf{Visual Generative Models (VGMs)} like DALL-E (\cite{ramesh2021zero}), Zero-1-to-3 (\cite{liu2023zero}) and Sora (\cite{videoworldsimulators2024}), demonstrate the capability to generate 2D images, videos or 3D meshes through text or images.
    \item \textbf{Visual-Language Models (VLMs)} like GPT-4V (\cite{achiam2023gpt}), CLIP (\cite{radford2021learning}) showcase robust comprehension of both vision and language, such as open-set image classification and visual question answering. 
    \item \textbf{Large Multimodal Models (LMMs)} expand their scope beyond vision and language to create novel categories of foundation models incorporating additional modalities, such as ULIP (\cite{xue2023ulip}) aligns point cloud representation to the pre-aligned image-text feature space. VLMs are a type of LMMs (\cite{firoozi2023foundation}). However, due to the current literature focusing more on VLMs, this paper will treat VLMs as a separate category. To avoid confusion, LMMs in this paper refer to those that include images, language, and more modalities.
    \item \textbf{Robotic-specific Foundation Models (RFMs)}, like RT-X (\cite{rt_x_2023}). Internet-scale dataset, such as images and text data, are suitable for pre-training visual and language models, but lack task level manipulation data. Therefore, researchers aim to train end-to-end RFMs by collecting task-level manipulation datasets to enable observations-to-action mapping.
\end{enumerate}

\begin{figure}[t]
  \includegraphics[width=\linewidth]{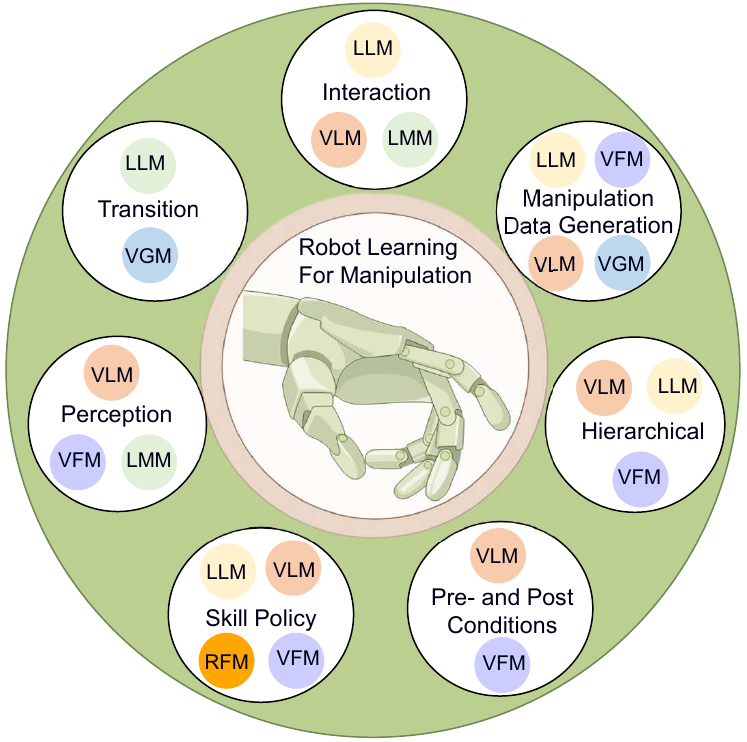}
  \centering
  \caption{LLMs help address challenges in Interaction, Manipulation Data Generation, Hierarchy of Skills, Skill Policy Learning, and Environment Transition Model. VLMs assist in tackling challenges in Interaction, Manipulation Data Generation, Hierarchy of Skills, Pre- and Post-conditions Detection, Skill Policy Learning, and Perception. LMMs aid in addressing challenges in Interaction and Perception. VGMs tackle the challenge of Manipulation Data Generation and Environment Transition. VFMs help address challenges in Manipulation Data Generation, Hierarchy of Skills, Pre- and Post-conditions Detection, Skill Policy Learning, and Perception. RFMs assist in addressing the challenge of Skill Policy Learning.}
  \label{fig:teaser}
\end{figure}

In this survey, we investigate how foundation models are utilized in robot learning for manipulation, like Fig. \ref{fig:teaser}:

\begin{enumerate}
    \item \textbf{LLMs} enable the direct generation of policy codes or action sequences and facilitate natural interaction with the environment.
    \item \textbf{VFMs} enhance open-world perception.
    \item \textbf{VLMs} serve as the cornerstone for alignment between vision and language, facilitating understanding of multimodality.
    \item \textbf{LMMs} expand their modalities to include 3D point cloud and haptic data, among others.
    \item \textbf{VGMs} generate 2D images or 3D meshes based on prompting, aiding in scene generation within simulation environments.
    \item \textbf{RFMs} serve as an end-to-end policy model, directly outputting actions based on input observations.
\end{enumerate}

These findings underscore the potential of embedding foundation models into manipulation tasks as a viable path toward achieving general manipulation capability. However, we do not believe that a single foundation model alone can achieve general manipulation capability. Although RFMs currently represent a single-model end-to-end training approach, ensuring safety and stability, particularly in achieving an over 99\% success rate in manipulation tasks, remains a challenge. Achieving over a 99\% success rate in manipulation tasks is crucial, as human manipulation success rates are around 99\%. Without this level of accuracy, robots can't replace humans (\cite{nishanth2023}). Therefore, drawing inspiration from the development of autonomous driving systems (\cite{hu2023planning}), achieving general manipulation capability necessitates an overarching framework that encompasses multiple functional modules, with different foundation models assuming distinct roles in facilitating general manipulation capability.

The ultimate general manipulation framework should be able to interact with human or other agent and control whole-body to manipulate arbitrary objects in open-world scenarios and achieve diverse manipulation tasks (\cite{mccarthy2024towards}). Drawing from \cite{kroemer2021review} and this general manipulation definition, we propose a comprehensive framework for general manipulation. However, the interaction between robot and human involves not only recognizing intentions but also learning new skills or improving old skills from human experts in the external world. Open-world scenarios may be static or dynamic. Objects can be either rigid or deformable. Task objectives can vary from short-term to long-term. Furthermore, tasks may necessitate different degrees of precision with respect to contact points and applied forces/torques. Although there are many challenges, achieving general manipulation can be accomplished through multiple stages. We designate the restriction of the robot's learning capability to improving old skills and to manipulating rigid objects in static scenes in order to achieve short-horizon task objectives with low precision requirements for contact points and forces/torques as Level 0 (L0). At the same time, we believe that improving the algorithm performance of different modules in the framework can support the transition from the L0 stage to the final general manipulation. Hence, we aim to use this survey not only to enlighten scholars on the issues that foundation models can address in robot learning for manipulation but also to stimulate their exploration of the general manipulation framework and the role various foundation models can play in the general manipulation framework. 

\cite{di2023towards} and \cite{firoozi2023foundation}  provide detailed descriptions of the application of foundation models in navigation and manipulation, but these lack thoughtful consideration of the relationship between foundation models across different applications. The survey most closely related to this paper is  \cite{xiao2023robot}. Compared to this survey, our survey focuses on the contributions of foundation models to robot learning for manipulation, proposing a comprehensive framework and detailing how foundation models can address challenges in each module of the framework.

This paper is structured as follows: In Sec. \ref{sec:framework}, we present a  comprehensive framework of robot learning for general manipulation, based on the developmental history of robot learning for manipulation and general manipulation definition. We elaborate on the impact of foundation models on each module in the framework in the following sections. Sec. \ref{sec:Interaction} is Human/Agent Interaction module, Sec. \ref{sec:prepost} is Pre- and Post-conditions Detection module, Sec. \ref{sec:Hierarchy} is Hierarchy of Skills module, Sec. \ref{sec:state}  is State Perception module, Sec. \ref{sec:policy} is Policy module, Sec. \ref{sec:generation} is Manipulation Data Generation module. In Sec. \ref{sec:discussion}, we discuss several issues of particular concern to us. In Sec. \ref{sec:conclusion}, we summarize the contributions of this survey and identify the limitations of the current framework as well as the challenges in each module.

\section{Framework of Robot Learning for General Manipulation}\label{sec:framework}
Over the past decade, there has been a significant expansion in research concerning robot manipulation, with a focus on leveraging the growing accessibility of cost-effective robot arms and grippers to enable robots to interact directly with the environment in pursuit of their objectives. As the real world encompasses extensive variation, a robot cannot expect to possess an accurate model of its unstructured environment, the objects within it, or the skills necessary for manipulation in advance (\cite{kroemer2021review}). 

\begin{figure*}[htbp]
\centering
\includegraphics[width=\textwidth]{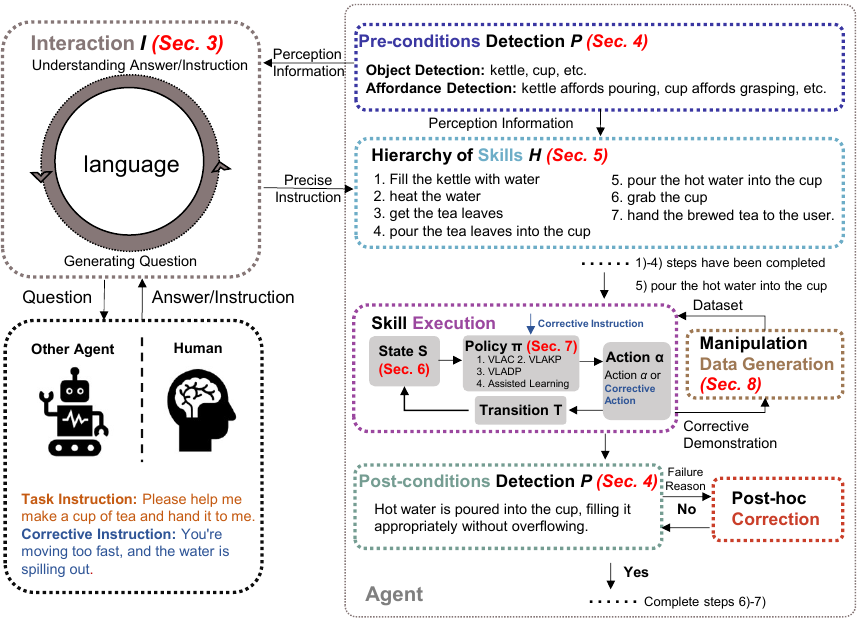}
\caption{Framework of Robot Learning for General Manipulation. The Pre-conditions Detection module $P$ perceives the environment to identify objects and the affordances objects support. The Interaction module $I$ receives instruction from a human or other agent. It uses perception information from the Pre-conditions Detection module $P$ to check for ambiguities in the instruction. If there are any ambiguities, it generates a question to clarify the instruction by asking the human or other agent. The Hierarchy of Skills module $H$ generates subgoals by using precise instruction from the Interaction module $I$ and perception information from the Pre-conditions Detection module $P$. Each subgoal is then passed to the Skill Execution module. 
In the Skill Execution module, Policy module $\Pi$ generates Action $\alpha$ based on the State $S$. To obtain the next state after executing the current action, State $S$ can either perceive it from the environment or use the Transition module $T$. To train the Skill Execution module, including the State module $S$, the Policy module $\Pi$ and the Transition module $T$, the Manipulation Data Generation module is required. This module provides a task-level manipulation dataset. When issues arise during execution, corrective instruction is sent to the Policy module $\Pi$ for manual adjustment. Policy module $\Pi$ modifies the current action to corrective action and saves corrective demonstration to the dataset for self-improvement of Policy module $\Pi$. After skill execution, Post-conditions Detection module $P$ determines the success of execution. If successful, proceed to the next subgoal; if not, the failure reason is conveyed to Post-hoc Correction module for self-correction.}
\label{fig:robot_learning_framework}
\end{figure*}

Early stage, robot manipulation is defined as learning a policy $\Pi$ through deep learning, reinforcement learning, or imitation learning, etc. This policy controls the robot's joint movements and executes tasks based on observations of the environment and the robot's state $S$, mapping to actions $\alpha$. such as Rlafford (\cite{geng2023rlafford}) and Graspnet (\cite{fang2020graspnet}) take point cloud as input and output the target pose. This process is represented by the Skill Execution module, as shown in Fig. \ref{fig:robot_learning_framework}.

In the mid-term, many tasks in robotics require a series of correct actions, which are often long-horizon tasks. For example, making a cup of tea with a robot involves multiple sequential steps such as boiling water, adding a tea bag, pouring hot water, etc. Learning to plan for long-horizon tasks is a central challenge in episodic learning problems (\cite{wang2020long}). Decomposing tasks has several advantages. It makes learning individual skills more efficient by breaking them into shorter-horizon, thus aiding exploration. Reusing skills in multiple settings can speed up learning by avoiding the need to relearn elements from scratch each time. Researchers train a hierarchy model to decompose the task into a sequence of subgoals (\cite{ahn2022can}), and observe pre- and post-conditions to ensure that the prerequisites and outcomes of each subgoals are met (\cite{cui2022can}). These three processes are represented as the Hierarchy of Skills module $H$, the Pre-conditions Detection module $P$, and the Post-conditions Detection module $P$ in Fig. \ref{fig:robot_learning_framework}. However, detecting only task success with post-conditions detection is insufficient. It should also identify the reasons for task failure to help the robot self-correct and improve success rates. Therefore, we add a Post-hoc Correction module, as shown in Fig. \ref{fig:robot_learning_framework}.

Recently, researchers have realized that training policies require real-world interaction between the robot and environments, which inevitably increases the probability of unforeseen hazardous situations. Therefore, researchers aim to train the environment's transition model $T$. Once the model is fitted, robot can generate samples based on it, significantly reducing the frequency of direct interaction between the robot and environments (\cite{liu2024learning}). This process is represented as the Transition module $T$ in Fig. \ref{fig:robot_learning_framework}.

The modules described above are summarized from the development of robot learning for manipulation. However, they are still insufficient for a comprehensive framework for general manipulation. The ultimate general manipulation framework should be able to interact with human or other agent and control whole-body to manipulate arbitrary objects in open-world scenarios, achieving diverse manipulation tasks. When interacting with human or other agent to understand task objectives, the transmitted instruction may sometimes be unclear, such as when there are two cups in the environment, it needs to determine which cup to pour water. Therefore, we add the Interaction module $I$ in Fig. \ref{fig:robot_learning_framework} to understand the precise task objective.

The aforementioned modules all require datasets for learning. The data collection process for the Hierarchy of skills $H$ and Pre- and Post-conditions detection modules $P$ is similar to that in the fields of CV and NLP. Compared to data collection in CV and NLP domains, gathering datasets for manipulation tasks requires the robot's trajectory to train the policy. Therefore, we include the Manipulation Data Generation module in Fig. \ref{fig:robot_learning_framework}.

We organize the framework of robot learning for general manipulation according to its development history and definition, as shown in Fig. \ref{fig:robot_learning_framework}. In the caption of Fig. \ref{fig:robot_learning_framework}, we outline the flow of the entire framework. To better illustrate the role of each module, we list the inputs and outputs of each module below, along with their specific functions. 

\begin{enumerate}
     \item \textbf{Pre-conditions Detection.} This module takes raw information observed by the robot as input. It outputs perception information about objects in the environment and affordances of those object. Perception information helps ensures that requirements are met and helps select the execution method based on object affordances. For instance, when placing a tea bag in a teacup, perception information can help determine whether there are tea bags and teacups and chooses between pick-place or pushing based on their affordances, such as, a tea bag is spherical, and it has the affordance of rolling when pushed. 
     
    \item \textbf{Human/Agent Interaction.} The input to the Human/Agent Interaction module $I$ consists of an instruction or answer from the collaborating agent or human, and perception information from the Pre-conditions Detection module $P$. The output includes a question if the instruction or answer has ambiguities and provides a precise instruction to the Hierarchy of Skills module $H$. The main function of this module is to understand the exact task objectives. 
    \item \textbf{Hierarchy of skills.} This module takes as input the perception information about objects in the environment and their affordances for the task from the Pre-conditions Detection module $P$, as well as the precise instruction from the Interaction module $I$. It then produces a sequence of subgoals as output. The concept of  `Hierarchy of skills' often involves creating a sequence of subgoals (\cite{song2023llm}). Each subgoal necessitates a skill, which may consist of one or multiple primitive actions (\cite{zhang2023bootstrap}). For instance, tasks like filling the kettle with water, heating the water, and getting the tea leaves are examples of subgoals that robot needs to achieve in a specific order to fulfill the final goal as instructed.

    \item \textbf{State.} The input to the State module is the current environment, objects and robot observation. States require the use of multiple sensors for perception. The output is the features of states. The states consists of robot proprioception $S_{robot}$, environment state $S_{e}$, and objects states $S_{o}$. The difference between $S_{e}$ and $S_{o}$ is analogous to the foreground and background of an image. $S_{robot}$ generally relates to the mechanical structure of the robot. Currently, there are limited studies focusing on the improvement of robot mechanical structure using foundation models, with \cite{stella2023can} being one of them. However, researches in this direction are scarce and still in their initial stages.
    \item \textbf{Policy.} The Policy module takes as input features from the State module $S$ and subgoals generated by the Hierarchy of Skills module $H$. The policy outputs action to accomplish task goals based on the input states. 
    We categorize action into three types: Code, Key Pose, and Dense Pose. Code refers to the direct control code of the robot. Key Pose refers to the desired poses of the end-effector, which is input to motion planning to generate the trajectory. Dense Pose refers to the next waypoint the end-effector moves to, with continuously outputted dense pose forming the trajectory. At present, the methods for generating actions using foundation models include LLMs directly generating code for robot execution, VLMs directly generating or VLMs combined with LLMs generating corresponding key poses, RFMs directly outputting key poses or dense poses through end-to-end training, and foundation models assisting reinforcement learning in generating various actions.
    \item \textbf{Post-conditions Detection.} This module takes as input the environment, objects and robot states observed after the robot performs a task, along with the subgoals generated by the Hierarchy of Skills module H. It outputs whether the current subgoal is successful. If not, it provides the reason for failure to Post-hoc Correction module. The Post-hoc Correction module generates a sequence of actions for self-correction based on the failure reason. For example, if a teacup is knocked over during pick-and-place, inform post-hoc correction and use pick-and-place to upright the cup and reinsert the tea bag.
    
    \item \textbf{Transition.} The Transition module $T$ takes an action generated by the Policy module $P$ as input. It outputs the next state after executing this action, thus helping to reduce the interaction between the robot and the real environment. UniSim (\cite{yang2023learning}) introduces the action-in-video-out framework as an observation prediction model. It takes the current action as input and produces the subsequent observation as output.
    \item \textbf{Manipulation Data Generation.} This module functions as a database. It takes in existing manipulation data and correction data generated from robot tasks. The output is to provide task-level manipulation datasets for offline training.  
\end{enumerate}

Current research on foundation models for manipulation primarily focuses on several key modules: the Interaction module in Sec. \ref{sec:Interaction}, the Pre- and Post-conditions Detection module in Sec. \ref{sec:prepost}, the Hierarchy of Skills module in Sec. \ref{sec:Hierarchy}, the State module in Sec. \ref{sec:state}, the Policy module in Sec.\ref{sec:policy}, and the Manipulation Data Generation module in Sec. \ref{sec:generation}. The following section will provide an overview of these modules.

\section{Human/Agent Interaction}\label{sec:Interaction}
There are two ways for human or other agent to interact with robot: 1) Providing task instruction to the robot to help it understand the task objective and complete the task independently (\cite{khan2023natural}). 2) Collaborating with the human or other agent to complete tasks, sharing workspace information, and conveying corrective instruction when useful or error-correcting information is identified to optimize the robot's current action (\cite{lynch2023interactive}). 

When conveying task instruction to the robot, there may contain language ambiguity in the task goal, such as having both red and green cups in the scene, and the task instruction is `grasp the cup'. This ambiguity may confuse the robot regarding which color cup to grasp. To address this issue, the robot needs to inquire about and confirm the final task objective from the human or other agent, thus requiring enhancement of their capability in text generation and comprehension. When conveying corrective instruction to a robot, it needs to comprehend the meaning of the corrective instruction and translate corrective instruction into appropriate actions. For instance, if a robot is picking up a book from a shelf filled with books, lifting too quickly may cause other books to fall. Human or collaborating agent need to alert the robot that the current lifting action is dangerous and advise it to lift slowly. If necessary, the robot should also report its current execution state, such as its grasping speed, and inquire whether this speed is considered high. However, corrective instruction are diverse; thus, understanding them is essential.

In addressing instruction ambiguity and text generation and comprehension challenges, SeeAsk (\cite{mo2023towards}) utilizes CLIP's perceptual module to identify objects in the scene and employs a fixed questioning template to organize language to ask about which object will be manipulated. Although the use of CLIP enhances the generalization ability for object recognition, it can't generate text for asking questions and to comprehend answers from the outside world and SeeAsk (\cite{mo2023towards}) focuses solely on addressing ambiguities concerning object color and spatial relationship due to a fixed questioning template. KNOWNO (\cite{ren2023robots}) utilizes LLM to score the next action to be taken. If the score difference between the top two actions is less than a threshold, it's considered ambiguity, prompting a confirmation for the final action. This approach improves efficiency and autonomy. Matcha (\cite{zhao2023chat}) not only employs vision but also utilizes haptic and sound senses to perceive object properties, such as material. When encountering ambiguity in object attribute recognition, it leverages LLM to generate inquiry content. CoELA (\cite{zhang2023building}) utilizes LLM as both a communication module and a planning module to enhance interaction text generation and comprehension, as well as task scheduling, with collaborative agent. LLM-GROP (\cite{ding2023task}) utilizes LLM to extract latent commonsense knowledge embedded within task instruction. For example, a task instruction might be "set dinner table with plate and fork," while the latent commonsense knowledge could be "fork is on the left of a bread plate."

As for corrective instruction, LILAC (\cite{cui2023no}) utilizes GPT-3 to distinguish between task instruction and corrective instruction. It then employs Distil-RoBERTa to extract text features and input them into the network to modify the robot's original trajectory. LATTE (\cite{bucker2023latte}), on the other hand, employs BERT and CLIP to extract features from corrective instruction and observation images and input them into the network to modify the robot's original trajectory. RT-H (\cite{belkhale2024rt}) employs VLMs in a two-step operation, initially outputting abstract delta-pose representations like "move left," which are then converted into delta poses and human intervention can enable robots to adjust trajectories based on human language instruction. 

\subsection*{Summary}
Following Fig. \ref{fig:interaction_challenge}, LLMs using chain of thought efficiently identifies ambiguity, surpassing the limitations of enumerating ambiguity. LLMs' comprehension of text effectively understands corrective instruction and transforms the original trajectory into a corrective trajectory.

\begin{figure}[htbp]
\centering
\includegraphics[width=\linewidth]{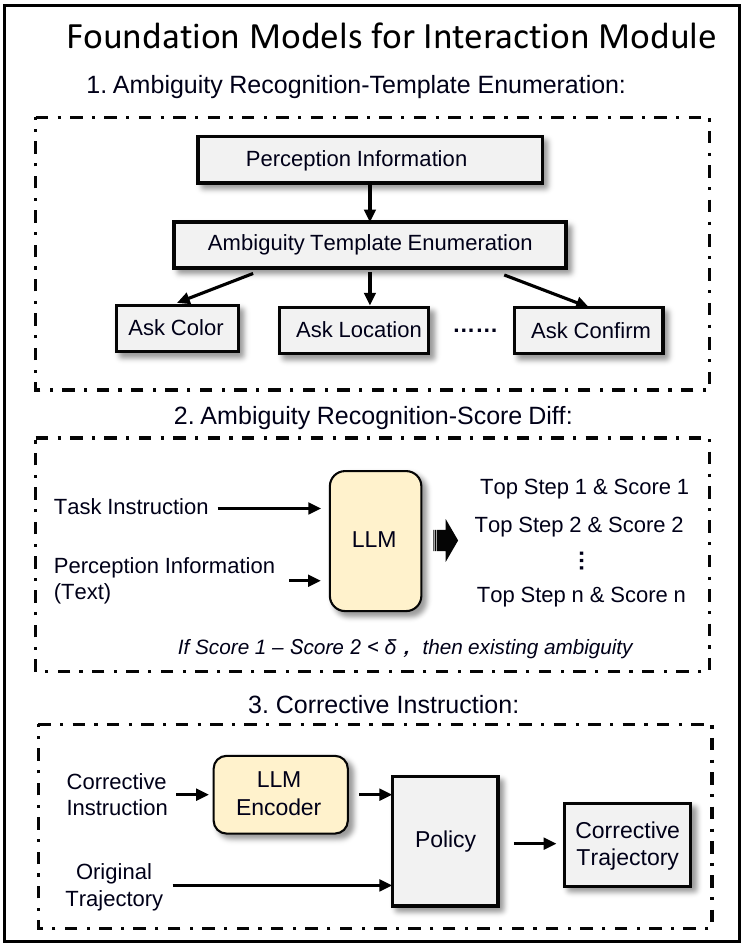}
\caption{Foundation Models for Interaction Module. Interaction mainly involves the exchange between task instruction and corrective instruction. Ambiguity often arises in task instruction interaction, hence robot needs to detect ambiguities. 1) One approach is to perceive objects in a multi-modal environment and enumerate possible ambiguities based on perception information (\cite{mo2023towards}). 2) Another approach involves using LLM to be the next step prediction module, which predicts and scores the next step; if the scores of the top 2 steps are less than $\delta$, it is considered that the task goal is ambiguous (\cite{ren2023robots}). 3) Strong comprehension skills are required during the transmission of corrective instruction, and the current mainstream approach involves using the encoder of LLM to extract tokens and input them into the policy to modify the original trajectory (\cite{bucker2023latte}).}
\label{fig:interaction_challenge}
\end{figure}

\section{Pre- and Post-conditions Detection}\label{sec:prepost}
In pre- and post-conditions detection, it is necessary to identify the initial and termination conditions. In pre-conditions detection, recognize objects and observe the affordances of objects. In post-conditions detection, identify whether a task has been successfully executed and provide reasons for task failure after skill execution. Currently, there are few papers focusing on identifying termination conditions. \cite{cui2022can} utilizes CLIP to compare the target's text or image with the termination environment to determine the success of task execution. Few articles are found in this study that address the output of task failure reasons after skill execution. RobotGPT (\cite{jin2024robotgpt}) analysis task failure utilizes the positions of manipulated objects after execution, but task failure should be determined during execution. AHA (\cite{duan2024aha}) uses a large number of robotic failure trajectories to fine-tune the VLM. The fine-tuned VLM leverages keyframe trajectory images and task descriptions from the robot's current task execution process to detect failures and provide detailed, adaptable failure explanations. Therefore, this section focuses on literature discussing foundation models in pre-conditions detection including object affordance and object recognition.

\subsection{Object Affordance}
The affordances associated with an object represent the range of manipulations that the object affords the robot (\cite{gibson2014ecological}). Early approaches addressed the issue by treating it as a supervised task (\cite{kokic2017affordance}). However, the process of annotating datasets is laborious and time-consuming, making it impractical to exhaustively cover all geometric information present in real-world environments. Consequently, researchers are exploring the application of reinforcement learning, enabling robots to collect data and train affordance perception modules through continuous exploration (\cite{wu2021vat}). Nevertheless, current reinforcement learning methods are trained in simulated environments, leading to a significant sim-to-real gap. To address these challenges, researchers propose training the affordance perception module using videos of human interactions in real-world scenarios (\cite{ye2023affordance, bahl2023affordances}).

For supervised learning methods, GraspGPT (\cite{tang2023graspgpt}) utilizes LLM outputs for object class descriptions and task descriptions. Object class descriptions detail the geometric shapes of each part of an object, while task descriptions outline the desired affordances for task execution, such as the types of manipulation actions to be taken. Integrating both components into the task-oriented grasp evaluator enhances the quality of the generated grasp pose. 3DAP (\cite{nguyen2023language}) utilizes the text encoder of LLM for feature extraction. The extracted features from desired affordances text are inputted into both the affordance detection module and pose generation module. This enhances the quality of the predicted affordance map and the generated pose.

In reinforcement learning, ATLA (\cite{ren2023leveraging}) utilizes GPT-3 to generate language descriptions of tools. These descriptions are then inputted into a pre-trained BERT model to obtain representations. The extracted features are finally fed into the SAC network module. Meta-learning techniques are employed to enhance the learning efficiency for the use of new tools. \cite{xu2023joint} employ CLIP's text and image encoders to extract features from language instruction and scene image, improving the quality of grasp pose generation in the SAC module.

The methods mentioned above utilize foundation models to assist other learning methods in improving affordance maps or grasp poses. There are also direct approaches using foundation models to generate affordance maps and grasp poses. PartSLIP (\cite{liu2023partslip}) converts 3D point clouds into 2D rendering images and inputs multi-view 2D images and textual descriptions of object parts into GLIP for object parts detection, ultimately fusing 2D bounding boxes into 3D segmentation to generate affordance maps. However, PartSLIP requires manual definition text prompts and additional algorithms to convert 2D boxes back to 3D regions. UAD (\cite{tang2025uad}) clusters object points into fine-grained semantic regions based on pixel-wise features extracted from multi-view rendered images using DINOv2. It then queries the VLM to generate a set of task instructions and associates these instructions with the most relevant clustered region to construct the affordance map. LAN-grasp (\cite{mirjalili2023lan}) inputs human instruction into LLM, utilizing its prior knowledge to output the shape of part to be grasped. These shapes, along with the object's 2D image, are then inputted into VLM to detect the bounding box for the grasping part. Finally, the bounding box and the point cloud from object 3D reconstruction are inputted into the grasp planner to generate grasp poses.

\begin{figure*}[htbp]
\centering
\includegraphics[width=0.9\linewidth,height=0.40\textheight]{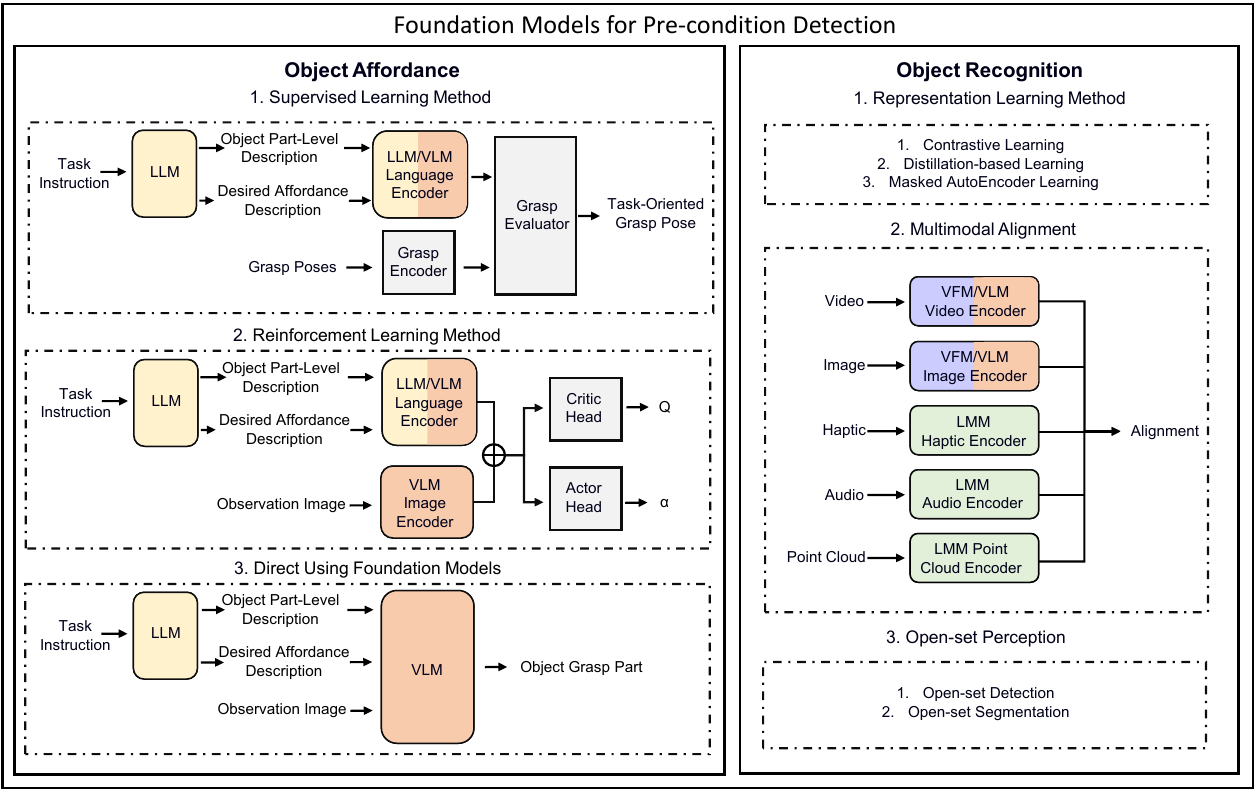}
\caption{Foundation Models for Pre-conditions Detection. As for object affordance, the main approaches of task-oriented grasp are supervised learning and reinforcement learning. Both methods utilize LLM to generate object part-level description and desired affordance description in task instruction, then fuse tokens and features into the original network through language encoder and image encoder to output task-oriented grasp pose (\cite{tang2023graspgpt, ren2023leveraging}). In reinforcement learning, it is possible to choose between a LLM language encoder with a custom-designed image encoder, or a VLM language encoder with a VLM image encoder. When selecting the LLM language encoder with a custom image encoder, the LLM language encoder should be frozen, and the custom image encoder should be trained (\cite{ren2023leveraging}). When using the VLM language encoder with the VLM image encoder, both encoders are typically frozen (\cite{xu2023joint}). Direct using foundation method utilizes LLM to generate object part-level description and desired affordance description according to task instruction. VLM marks out the part of the object to grasp in the image based on the description (\cite{liu2023partslip}). As for object recognition, the representation learning methods in state perception mainly include contrastive learning (\cite{radford2021learning}), distillation-based learning (\cite{caron2021emerging}), and masked autoencoder learning (\cite{radosavovic2023real}). Masked autoencoding methods prioritize low-level spatial aspects, sacrificing high-level semantics, whereas contrastive learning methods focus on the inverse, the fusion of masked autoencoder and contrastive learning is employed in both Voltron (\cite{karamcheti2023language}) and iBOT (\cite{zhou2021ibot}). Multimodal representation learning focuses primarily on multimodal alignment (\cite{xue2023ulip2, tatiya2023mosaic}). Training the encoder with large-scale data and parameters has facilitated open-set perception, including tasks such as open-set detection, open-set segmentation. For instance, SAM (\cite{kirillov2023segment}) utilizes the MAE (\cite{he2022masked}), ViLD (\cite{gu2021open}) employs the CLIP (\cite{radford2021learning}).}.
\label{fig:pre_conditions_detection}
\end{figure*}

\subsection{Object Recognition}
Object recognition can be categorized into two types: passive perception and active perception. Compared to passive perception, active perception adjusts the perspective to the areas of interest (\cite{kroemer2021review}). Then, modeling manipulation tasks and generalizing manipulation skills necessitate representations of both the robot's environment and the manipulated objects. These representations form the foundation for skill hierarchies, pre- and post-condition detection, skill learning, and transition model learning. 

The Vision Transformers (ViTs) and similar attention-based neural networks have recently achieved state-of-the-art performance on numerous computer vision benchmarks (\cite{han2022survey, khan2022transformers, zhai2022scaling}) and the scaling of ViTs has driven breakthrough capability for vision models (\cite{dehghani2023scaling}). The development of visual backbones not only advances pre-trained visual representations but also accelerates the progress of open-set perception tasks, such as segmentation and detection.

As for pre-trained visual representations, the algorithms mentioned have various training objectives: for instance, contrastive methods like Vi-PRoM (\cite{caron2021emerging}), R3M (\cite{nair2022r3m}), VIP (\cite{ma2022vip}), CLIP (\cite{radford2021learning}), LIV (\cite{ma2023liv}); distillation-based methods such as DINO (\cite{caron2021emerging}); or masked autoencoder methods like MAP (\cite{radosavovic2023real}), MAE (\cite{he2022masked}). The primary datasets utilized comprise the CLIP dataset (\cite{radford2021learning}), consisting of 400 million (image, text) pairs sourced from the internet, along with ImageNet (\cite{deng2009imagenet}), Ego4D (\cite{grauman2022ego4d}), and EgoNet (\cite{jing2023exploring}).

Pre-trained visual representations have high transfer ability to policy learning (\cite{xiao2022masked, yang2023moma}), but visual representation involves not just recognizing spatial features but also understanding semantic features. Masked autoencoding methods prioritize low-level spatial aspects, sacrificing high-level semantics, whereas contrastive learning methods focus on the inverse (\cite{karamcheti2023language}). The fusion of masked autoencoder and contrastive learning is employed in both Voltron (\cite{karamcheti2023language}) and iBOT (\cite{zhou2021ibot}). The loss function achieves a balanced trade-off between these two aspects. To compare different pre-trained visual representations, benchmarks are established by CORTEXBENCH (\cite{majumdar2023we}) and EmbCLIP (\cite{khandelwal2022simple}) to assess which model could provide a better "artificial visual cortex" for manipulation tasks. However, the models included in these benchmarks are still not comprehensive enough. 

The aforementioned pre-trained visual representations mainly involve the extraction of features from 2D images. The experience of learning representations on 2D images can also be extended to other modalities. For the object point cloud modality, ULIP (\cite{xue2023ulip}) and ULIP2 (\cite{xue2023ulip2}) employ contrastive learning to align features between point clouds and text-images.  Point-BERT (\cite{yu2022point}) uses the masked autoencoding method to learn point cloud features by reconstructing point clouds. GeDi (\cite{poiesi2022learning}) uses a contrastive learning approach to extract general and distinctive 3D local geometric information. In the haptic modality, MOSAIC (\cite{tatiya2023mosaic}) utilizes contrastive learning to train the haptic encoder.

As for segmentation, SAM (\cite{kirillov2023segment}) develops a transformer-based architecture and creates the largest segmentation dataset, with over 1 billion masks from 11 million images. The model is adaptable and enables zero-shot transfer to new tasks and image distributions. Fast-SAM (\cite{zhao2023fast}) and Faster-SAM (\cite{zhang2023faster}) aim to improve the training and inference speed of the network by enhancing its network structure. TAM (\cite{yang2023track}) merges SAM (\cite{yang2023track}) and XMem (\cite{cheng2022xmem}) for high-performance interactive tracking and segmentation in videos.

As for detection, traditional detection models are usually confined to a narrow range of semantic categories because of the cost and time involved in gathering localized training data within extensive or open-label domains.  However, advancements in language encoders and contrastive image-text training enable open-set detection. Researchers integrate language into a closed-set detector to generalize open-set concepts, detecting various classes through language generalization despite being trained solely on existing bounding box annotations, such as OWL-ViT (\cite{minderer2022simple}), Grounding-DINO (\cite{liu2023grounding}), OVD (\cite{zareian2021open}), ViLD (\cite{gu2021open}), DetCLIP (\cite{yao2022detclip}).

Deploying such models in open-set detection presents a significant challenge, primarily because even slight alterations in prompting can greatly impact performance. Fine-tuning can enhance a foundation model's understanding of prompting. However, foundation models are often over-parameterized, leading to slow training processes. COOP (\cite{zhou2022learning}) maps prompting to a set of learnable vectors, which can be optimized through network training. In CLIP-Adapter (\cite{gao2024clip}), two extra linear layers are appended after the final layer of either the vision or language backbone to enable efficient few-shot transfer learning through fine-tuning. 

The method for open-set detection on 2D images can be extended to the research direction of open-set detection on 3D point clouds. PointCLIP (\cite{zhang2022pointclip}) utilizes pre-trained CLIP to extract multi-view depth image features of point cloud, then compares the extracted features with textual features to identify the point cloud category.

\subsection*{Summary}
As shown in Fig. \ref{fig:pre_conditions_detection}, LLM provides object part-level knowledge via text, aiding in affordance map or grasp pose generation. Reinforcement learning can make robotic systems perform better though interaction than supervised learning trained on datasets. Direct use of foundation models avoids training. However, stability remains a concern. In object recognition, representation learning aligns multimodal features with text, improving model cognition, similar to human think with words. It also supports open-set perception tasks, like detection and segmentation.

\section{Hierarchy of skills }\label{sec:Hierarchy}
The skill hierarchy is closely related to the field of task and motion planning (TAMP). TAMP aims to address high-level instructions by organizing tasks in a sequence that ensures dynamic feasibility (\cite{guo2023recent}). There are three main types of classical TAMP methods: constraint-based TAMP, sampling-based TAMP, and optimization-based TAMP (\cite{zhao2024survey}). Constraint-based and sampling-based TAMP define the problem with goal conditions. Unlike optimization-based TAMP, these approaches often cannot assess or compare the quality of the generated plan or final state due to the lack of objective functions, such as when the goal is to pour as much water as possible into the cup (\cite{zhao2024survey,zhang2022task}). However, optimization-based TAMP is sensitive to the initial conditions and goal setup of the problem (\cite{zhao2024survey}).

The scalability of classical TAMP methods is often constrained by the tree search problem size for complex tasks and the computational cost of evaluating heuristics and optimal trajectories (\cite{zhao2024survey}). Integrating learning-based approaches into TAMP enables informed decision-making based on prior examples and experiences and improves flexibility and generalizability (\cite{guo2023recent}). Models for skill hierarchy can be trained using text or videos, similar to how humans learn assembly procedures from instructional manuals or tutorial videos. As for tutorial videos, VLaMP (\cite{patel2023pretrained} and SeeDo (\cite{wang2024vlm}) use trained models to understand human video operations and HourVideo (\cite{chandrasegaran2024hourvideo}) proposes a benchmark dataset specifically designed for hour-long video-language understanding. 

Traditional TAMP's domain representations are usually manually specified by expert users such as PDDL (\cite{silver2022pddl}). However, LLMs have been explored for processing and interpreting natural language inputs (\cite{huang2022language}). They offer a novel approach to encoding the planning domain in a more intuitive and accessible way. Furthermore, LLMs' acquisition of world knowledge and commonsense reasoning has the potential to improve the scalability and generalizability of skill hierarchy tasks (\cite{vemprala2023chatgpt,jansen2020visually,palme}). Various benchmarks such as PlanBench (\cite{valmeekam2023planbench}) can assess the planning and reasoning capability of LLMs.

LLMs possess a notable limitation: they lack practical experience, hindering their utility for decision-making within a specific context, so the output of LLMs often cannot be translated into executable actions for the robot. \cite{huang2022language} first use pre-trained causal LLM to break down high-level tasks into logical mid-level action plans. Then, a pre-trained masked LLM is employed to convert mid-level action plans into admissible actions. However, prompts usually require the context of the robot's capability, its current state, and the environment. At the same time, LLMs are considered `forgetful' and don't treat information in the system prompt as absolute. Despite efforts to reinforce task constraints in the objective prompt and extract numerical task contexts from the system prompt, storing them in data structures, errors caused by LLM forgetfulness remain unresolved (\cite{chen2023forgetful}). 

To address the aforementioned issues, SayCan (\cite{ahn2022can}) scores pre-trained tasks based on prompting and observation images, generating the task sequence with the highest score. Saycan provides a paradigm for generating action sequences using LLM, but there are still some drawbacks: 1) The generated action sequences do not incorporate user preferences. 2) Safety regulations are not adequately addressed. 3) The limitation of the skill library. 4) LLM focuses solely on reasoning when generating action sequences, neglecting feedback on action execution. 5) The limitation of scene grounding. GD (\cite{huang2023grounded}) proposes a paradigm to address the aforementioned issues by not only scoring the generated action sequence using LLM but also introducing a grounded function model for scoring the generated action sequence. The grounded function model encompasses token-conditioned robotic functions, such as affordance functions that capture the abilities of a robot based on its embodiment, safety functions, and more. This approach tackles drawbacks by designing grounded functions, avoiding fine-tuning in LLM.

Regarding user preferences, TidyBot  (\cite{wu2023tidybot})  trains LLM by collecting users' preference data, enabling the trained LLM to choose behaviors that better align with user preferences.  As for safety regulations, \cite{yang2023plug} incorporate ISO 61508, a global standard for safely deploying robots in industrial factory settings, into the constraints of the action sequence generation. As for the skill library, BOSS (\cite{zhang2023bootstrap}) suggests using LLMs' rich knowledge to guide skills chaining in the skill library, aiming to create new skills through combinations. RoboGen (\cite{wang2023robogen}) employs generative models to create new skill task scenarios, then utilizes either reinforcement learning or gradient optimization methods to automatically learn new skills based on the reward function generated by the LLM. As for action execution feedback, REACT (\cite{yao2022react}), COWP (\cite{ding2022robot}), LLM-Planner (\cite{song2023llm}), CoPAL (\cite{joublin2023copal}) and PROGPROMPT (\cite{singh2023progprompt}) provide feedback on robot action execution to LLMs. This allows LLMs to adjust action sequences based on execution status, creating a closed-loop process for generating action sequences. 

\begin{figure}[htbp]
\centering
\includegraphics[width=\linewidth]{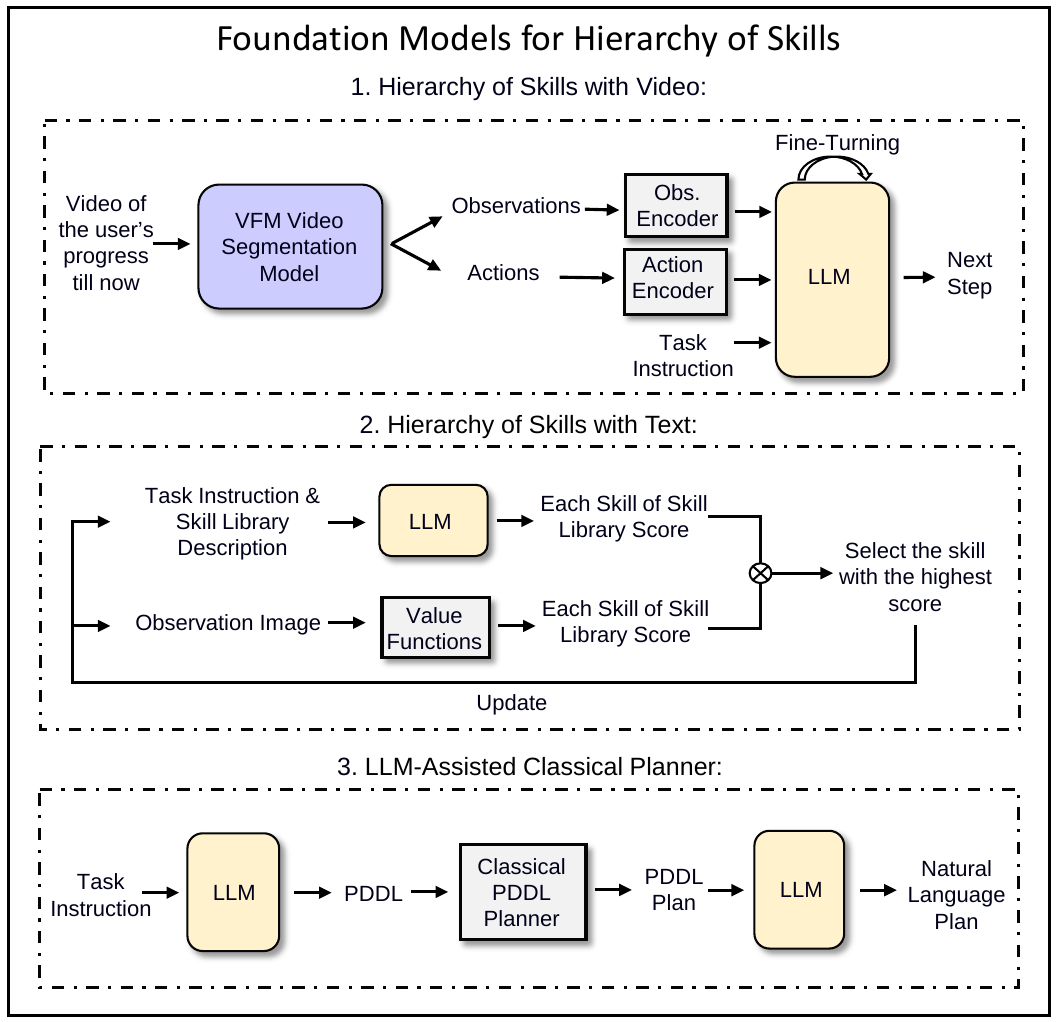}
\caption{Foundation Models for Hierarchy of Skills. 1) Utilize human operation video to learn the skill sequence for task execution, decompose the video of the user's progress so far into observations and human actions through segmentation, and input them along with task instruction into a pre-trained language model to predict the next step (\cite{patel2023pretrained}). 2) LLM scores the skills in the skill library based on task instruction and the skills already executed, and the value function also scores the skills in the skill library based on observation images. The highest-scoring skill, obtained by multiplying the two scores, is selected as the next step (\cite{ahn2022can}). The value function can consider multiple factors such as affordance, safety, user preference, and more (\cite{huang2023grounded}), and these considerations can also be fine-tuning LLM (\cite{wu2023tidybot}). 3) LLM assists the classical planner by translating task instruction into PDDL descriptions, sending them to the classical planner to generate a PDDL plan, and then translating the PDDL plan into a natural language plan using LLM (\cite{liu2023llm}).}
\label{fig:Hierarchy_of_Skills_template}
\end{figure}

As for the limitation of scene grounding, LLMs need to inquire about the scene representation to determine the availability, relationship and location of objects. NLMap (\cite{chen2023open}) proposes an open-vocabulary, queryable semantic representation map built on ViLD and CLIP. This map outputs the pose of related objects based on task instruction, which are then handed over to the LLM for planning. Text2Motion (\cite{lin2023text2motion}) incorporates a geometric value function on top of the value function, enabling the robot to select actions that adhere to geometric constraints based on scene descriptions. \cite{xu2023creative} explore the possibility of teaching robots to creatively utilize tools within scenarios, which involve implicit physical limitations and require long-term planning. 
VILA (\cite{hu2023look}) seamlessly incorporates perceptual data into ChatGPT-4V for its reasoning and planning processes, facilitating a deep comprehension of common sense knowledge within the visual domain, encompassing spatial arrangements and object characteristics. PHYSOBJECTS (\cite{gao2023physically}) fine-tunes a VLM to enhance its understanding of physical object attributes, such as material. This integration of a physically informed VLM into an interactive framework with a LLM enhances task planning performance in tasks incorporating instruction related to physical object attributes. SpatialVLM (\cite{chen2024spatialvlm}) and 3D-LLM (\cite{hong20233dllm}) utilizes a 2D pre-trained VLM to train on collected 3D datasets, enhancing capabilities related to 3D tasks while maintaining the abilities of previous tasks.


The hierarchy of skills possessed by LLMs or VLMs can be applied not only to single agent but also to multiple agents. SMART-LLM  (\cite{kannan2023smart}) utilizes LLM for the hierarchy of skills and allocates each task to every agent through the task assignment module. 

Regardless of whether the prompting input to LLMs is in natural language or PDDL format, the hierarchy of skills possessed by LLMs still exhibits instability (\cite{silver2022pddl}). Hence, researchers are exploring approaches that integrate LLMs with classical PDDL-based planning methods for the hierarchy of skills. LLM+P (\cite{liu2023llm}) utilize LLMs to translate natural language into PDDL and input into a classical planner for the hierarchy of skills. \cite{xie2023translating} indicate that LLMs exhibit greater efficacy in translation tasks as opposed to planning.

\subsection*{Summary}
According to Fig. \ref{fig:Hierarchy_of_Skills_template}, the hierarchy of skills is mainly divided into video instruction and language instruction. VFM and LLM play roles in perception and reasoning. Language instruction is further divided into methods based on foundation models and methods combining foundation models with classical TAMP. As shown in \ref{sec: appen_5} Tab. \ref{tab:comparise_5}, there is currently no significant research comparing video instruction and language instruction. However, from the modality perspective, video provides more temporal or spatial dependencies regarding tasks compared to language. This also means that video instruction requires a higher level of hierarchy of skills, not only needing to output task plans but also understanding the task and scene constraints from the video. Language instruction is more suitable for interaction and reasoning for LLMs/VLMs. However, the two share some similarities. Current research on the hierarchy of skills in language and video instruction tends to focus on SOTA VLMs, and both have similar failure modes, indicating that both face challenges in perception and reasoning. In the language instruction methods, the combination of foundation models and classical TAMP is more reliable than foundation models alone, but it also faces limitations in generalization. Therefore, how to better integrate foundation models with classical TAMP requires further research.

\section{State}\label{sec:state}

The State module focuses on perceiving the environment, objects, and robot states. Section. \ref{sec:prepost} introduces low-level perception methods. This section explains high-level approaches for 3D reconstruction and pose estimation.

\subsection{3D Reconstruction}
3D reconstruction involves capturing both the shape and appearance of an object or scene (\cite{wiki2025}). 3D reconstruction methods are divided into passive and active types (\cite{butime20063d}). Active methods involve contact or project some form of energy onto the object, such as Laser Scanning (\cite{butime20063d}), X-ray (\cite{maken20232d}). These devices have high accuracy, but they are usually expensive. Therefore, various studies focus on 3D reconstruction using consumer RGBD cameras, such as Microsoft Kinect, Intel RealSense, Google Tango, and ORBBEC Gemini (\cite{li2022high}).

These consumer cameras typically use principles such as structured light, time of flight, and traditional photometric stereo for depth estimation (\cite{zhou2024comprehensive}) and 3D representation can be generated by registering them using camera poses (\cite{huang2024surface}). However, when surfaces are shiny, bright, transparent, textureless or distant from the camera, depth images produced by consumer cameras are often noisy and incomplete. Several studies have addressed this challenging problem by learning to restore depth images (\cite{dai2022domain, fang2022transcg, sajjan2020clear}). However, the correct depth information may already be lost in the original depth. ASGrasp (\cite{shi2024asgrasp}) demonstrates that 3D reconstruction using raw multi-view images from consumer cameras is better than restoring the original depth. Many studies use a single image for 3D reconstruction (\cite{fu2021single}). However, a single image loses a significant amount of information, resulting in lower accuracy. Despite this, the zero-shot capability of current single image 3D reconstruction has led to its widespread application in simulation scene generation (in Sec. \ref{subsec:scene_generation}).

The 3D representation for 3D reconstruction can be expressed as explicit and implicit expressions (\cite{zhou2024comprehensive}). Explicit expressions include point clouds (\cite{shi2024asgrasp}), voxels (\cite{jiang2021synergies}), and meshes (\cite{wen2019pixel2mesh}). The three representations can be converted into each other (\cite{jiang2021synergies}), but each has its own advantages. A point cloud is made up of discrete points in space, providing flexibility in processing. A voxel can store spatial information inside an object but comes with high space complexity. A mesh uses triangle meshes to represent complex shapes and details accurately, such as deformation (\cite{wen2019pixel2mesh}). It ensures the projection is always convex, making it easier to rasterize (\cite{zhou2024comprehensive}). Implicit expressions represent 3D geometry using a function, such as signed distance function (SDF) (\cite{chabra2020deep}), occupancy field (\cite{jiang2021synergies}), and radiance field (\cite{mildenhall2021nerf}). They offer differentiability and efficient storage, making them a powerful tool. In contrast, explicit expressions tend to be more intuitive.

GIGA (\cite{jiang2021synergies}) points out that manipulation requires a fine-grained understanding of local geometry details. Implicit representations, due to their continuous and differentiable nature, can represent smooth surfaces at high resolution. As a result, there is a growing research using implicit representations for manipulation tasks (\cite{dai2023graspnerf, lu2024manigaussian}). Current state-of-the-art methods for representing scenes using implicit representations are mainly divided into Nerf-based (\cite{wang2023masked}) and 3DGS-based approaches (\cite{kulhanek2024nerfbaselines}). Compared to Nerf (\cite{mildenhall2021nerf}), 3DGS offers better real-time performance (\cite{kerbl20233dgaussiansplattingrealtime}). However, these implicit 3D representations currently lack scene semantics and not easily editable for 3D modifications (\cite{bai2024survey}).

As for scene semantics, semantic-NeRF (\cite{zhi2021place}) employs manually annotated semantic labels to jointly encode semantics, appearance, and geometry using NeRF. Manual annotation is time-consuming and labor-intensive. Due to the foundation model's robust open-set capability for objects, DFF (\cite{shen2023distilled}), CLIP-Fields (\cite{shafiullah2022clip}) and LERF (\cite{kerr2023lerf}) employ CLIP image encoder to extract features from multi-view 2D images for NeRF (\cite{mildenhall2021nerf}) reconstruction. These features are integrated as part of the output of the NeRF network, enriching the semantic information of the reconstructed 3D scenes. When a text prompt is provided, the features output by the CLIP text encoder can be compared with the CLIP image features output by NeRF to form a relevancy map. This relevancy map can support downstream tasks, such as semantic scene completion and object localization (\cite{ha2022semantic}). Since CLIP can only provide image-level features, the relevancy map lacks precise pixel-level object boundary information. 3DOVS (\cite{liu2023weakly}) incorporates DINO features into the NeRF output to distill object boundary information. OV-NeRF (\cite{liao2024ov}) addresses the issues of coarse relevancy maps and view-inconsistent relevancy maps through SAM and cross-view self-enhancement. FMGS (\cite{zuo2025fmgs}) transfers this concept from NeRF to 3DGS, achieving 851X faster inference. Although foundation models, such as CLIP and DINO, enable 3D open-set semantic scene understanding, the performance is limited by the foundation models themselves. For example, CLIP is constrained by the bag-of-words limitation (\cite{kerr2023lerf}).

The image features output by NeRF can be used to build a relevancy map. They can also be lifted into 3D space through multi-view images, serving as 3D features for downstream tasks (\cite{ze2023gnfactor}). 3D-LLM (\cite{hong20233d}) extracts 2D features from multi-view rendered images using the CLIP image encoder. These features are then fused into 3D features through Direct Reconstruct, gradSLAM (\cite{jatavallabhula2023conceptfusion}), or Neural Field methods (\cite{hong20233d}), endowing 3D features with semantic information.

For implicit 3D editing, some methods use human scribbles to edit 3D shape and appearance (\cite{zhang2023nerflets, schwarz2020graf, li2024interactive, liu2021editing}). However, this approach is not intuitive enough. With the development of foundation models, many methods for implicit 3D editing using text prompts have emerged. CLIP-NeRF (\cite{wang2022clip}) integrates semantic features extracted by CLIP into NeRF reconstruction to change object shape and appearance during rendering. However, CLIP-based approaches cannot precisely modify specific local regions. Instruct-NeRF2NeRF (\cite{haque2023instruct}) utilizes InstructPix2Pix (\cite{brooks2023instructpix2pix}) to iteratively edit multiview input images and optimize the underlying scene in NeRF. This process produces a refined 3D scene that adheres to the edit instruction. However, InstructPix2Pix modifies the entire image. As a result, regions that are not desired may also be altered. DreamEditor (\cite{zhuang2023dreameditor}) uses Dreambooth (\cite{ruiz2023dreambooth}) to generate 2D editing masks. These masks are then converted into 3D editing regions through back projection. This approach enables precise local editing.

\subsection{Pose Estimation}
Object pose estimation can be divided into marker-based and markerless methods (\cite{karashchuk2021anipose}). Marker-based methods require attaching passive or active markers (\cite{cassinis2007amirolos}) to the object. These methods achieve high accuracy in pose estimation. For example, the NDI Polaris Vega XT, commonly used in medical robotics, can achieve an accuracy of 0.12 mm RMS (\cite{NDI_Polaris_Vega_XT}). However, in unstructured environments, it is not feasible to attach specific markers to every object. Therefore, achieving object pose estimation in unstructured markerless environment is necessary.

From the perspective of generalization, pose estimation methods can be classified into instance-level, category-level, and unseen object approaches (\cite{liu2024deep}). Instance-level methods can estimate pose accurately for specific object instances on which they are trained. However, they struggle with novel objects. To improve the model's adaptation for pose estimation of novel objects, category-level approaches use geometric priors from objects of the same category to estimate the pose of the novel object without requiring its 3D model (\cite{wang2019normalized}). Unseen object approaches typically rely on the 3D model of the novel object to estimate its pose (\cite{caraffa2024freeze}).

Category-level and unseen object approaches can also be primarily classified into model-free and model-based approaches (\cite{liu2024deep}). Model-free methods do not require prior knowledge of the object’s 3D model. These methods typically regress the object pose using neural network (\cite{guan2024survey}). However, these methods require large amounts of data for the neural network to learn the geometric priors of the object. In contrast, model-based methods use a known 3D model of the object and usually lead the BOP benchmark for object pose estimation (\cite{burde2024comparative}). However, obtaining accurate object 3D models in the real world is not easy. The advancement of multi-view image 3D reconstruction technology bridges the gap between model-based and model-free real-world applications (\cite{burde2024comparative}).

The input modalities for the model-based approach include RGB, depth, and RGBD. Currently, the RGBD modality leads the BOP benchmark for object pose estimation. The optimization goals are primarily divided into three parts: 2D-2D correspondences followed by regression (\cite{nguyen2024gigapose}), 2D-3D correspondences followed by PnP (\cite{li2023nerf, ausserlechner2024zs6d}), and 3D-3D correspondences followed by least squares fitting (\cite{lin2024sam, caraffa2024freeze}). However, pose estimation accuracy remains a challenge when dealing with occlusion, specularity, symmetry and textureless objects (\cite{guan2024survey}). Many methods use the predicted pose as a coarse result and refine it to obtain a fine result (\cite{labbe2022megapose, wen2024foundationpose, moon2024genflow}).

\begin{figure}[htbp]
\centering
\includegraphics[width=\linewidth]{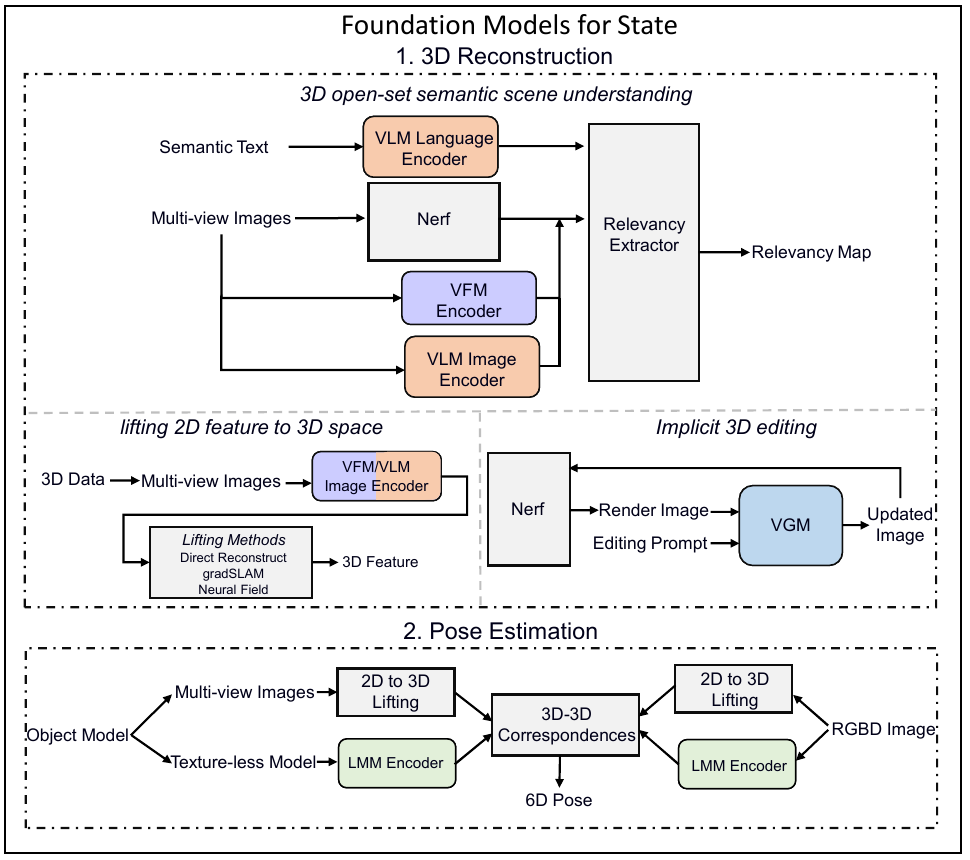}
\caption{Foundation Models for State. The foundation models have three main applications in 3D reconstruction: 3D open-set semantic scene understanding, lifting 2D features to 3D space, and implicit 3D editing. In 3D open-set semantic scene understanding, the main pipeline is to use image features extracted by the VFM encoder and VLM image encoder as input for NeRF. Then, semantic text features extracted by the VLM language encoder are used in conjunction with the image features from NeRF to generate a relevancy map through a relevancy extractor (\cite{kerr2023lerf}). This relevancy map can support downstream tasks, such as semantic scene completion and object localization (\cite{ha2022semantic}). As for lifting, using the VLM image encoder to extract features from 3D data multi-view images and lift them into 3D features can incorporate semantic information into the 3D features. The lifting methods include direct reconstruction, gradSLAM, and Neural Field (\cite{hong20233d}). For implicit 3D editing, the current mainstream pipeline is to input the image rendered by NeRF and the editing prompt into the VGM to generate the updated image. The updated image is then fed back into NeRF for training, modifying NeRF's radiance field representation of the 3D scene (\cite{haque2023instruct}). Pose estimation with foundation models achieves state-of-the-art results (\cite{caraffa2024freeze}). The main method is 2D-lifting-3D. It extracts texture features from the object model and observation RGBD image. LMM extracts geometric features from the object model and observation RGBD image. The fused features are then used to estimate the 6D pose through 3D-3D correspondences.}
\label{fig:State_Perception} 
\end{figure}

The pose estimation of moving objects mainly involves two methods. 1) Some single-image 6D pose estimation methods are fast and re-estimate poses from scratch for each frame. However, this approach is inefficient and results in less coherent estimations (\cite{wen2021bundletrack}). 2) Pose tracking utilizes temporal cues to improve pose prediction. It enhances efficiency, smoothness, and accuracy in video sequences. Current pose-tracking methods are mainly divided into probabilistic tracking (\cite{deng2021poserbpf,issac2016depth, stoiber2022iterative}) and optimization-based tracking (\cite{li2018deepim, wen2020se, lin2022keypoint, wang20206}). Pose tracking faces challenges mainly from motion blur, incremental error drift, and occlusion. To address these issues, BundleSDF (\cite{wen2023bundlesdf}) and BundleTrack (\cite{wen2021bundletrack}) use an online pose graph optimization process.

There are some research integrating foundation models into pose estimation. As for category-level, OV9D (\cite{cai2024ov9d}) utilizes DINO and VQVAE to extract visual features from images, while CLIP is used to extract text features from category prompts. These features are then fed into the Stable Diffusion UNet (\cite{rombach2022high}) to generate a normalized object coordinate space (NOCS) map. This method achieves generalizability to unseen categories and enables open-set pose estimation. In unseen object pose estimation with foundation models, FoundationPose (\cite{wen2024foundationpose}) utilizes LLM-aided synthetic data generation at scale to ensure strong generalizability for novel object pose estimation \& tracking. SAM-6D (\cite{lin2024sam}) and ZS6D (\cite{ausserlechner2024zs6d}) leverage SAM to generate valid proposals, enabling zero-shot 6D pose estimation. FreeZe (\cite{caraffa2024freeze}) employs frozen GeDi (\cite{poiesi2022learning}) and DINO (\cite{caron2021emerging}) to extract both geometric and visual features from the query object model and the target object's RGBD observation image. It then uses 3D-3D fused feature correspondences to obtain the 6D pose. Due to the foundation models' robust capability in discriminative feature extraction, FreeZe achieves state-of-the-art results without the need for any data or training. Overall, foundation models primarily improve generalization for novel object pose estimation in three areas: data , object recognition, and feature extraction. However, the performance is limited by the foundation models themselves. For example, foundation models are large in size (\cite{caraffa2024freeze}) and SAM may hallucinate in object segmentation (\cite{kirillov2023segment}).

\subsection*{Summary}
Following Fig. \ref{fig:State_Perception}, VLM and VFM assist implicit 3D reconstruction by generating relevancy maps that include semantic information (\cite{kerr2023lerf}). They can also be employed in 2D-to-3D lifting to extract 3D features, encompassing texture, semantic, and spatial information (\cite{hong20233d}). VGM aids in generating edited images and modifying 3D scenes based on these edited images (\cite{haque2023instruct}). FreeZe (\cite{caraffa2024freeze}) achieves state-of-the-art result in pose estimation by extracting discriminative features through 2D-to-3D lifting and LMM.

\section{Policy}\label{sec:policy}
The policy is divided into two categories: object/action-centric methods and end-to-end methods. Object/action-centric methods extract attributes from observations, such as bounding boxes, masks (\cite{sajjan2020clear}), or 3D spatial action-value map (\cite{shi2024asgrasp}). These extracted attributes are then transformed into either a sequence of key poses or a single key pose, which is used in motion planning to guide robot motion. End-to-end methods directly map observation to robot action (\cite{chi2023diffusion}). They eliminate the need for attribute extraction.

End-to-end methods are mainly divided into reinforcement learning (\cite{herzog2023deep}) and imitation learning (\cite{dasari2019robonet}). Recent end-to-end methods have made significant progress. ACT (\cite{zhao2023learning}) uses action chunks to reduce compounding errors. Diffusion policy (\cite{chi2023diffusion}) applies the idea of diffusion to visuo-motor control, tackling challenges such as action multimodality and sequential correlation to handle high-dimensional action sequences. 

However, the above methods are all one-model-for-one-task, lacking general-purpose capability. Due to the development of foundation models, general-purpose models have advanced. The representation of task instruction can be categorized into four types: language, human video, goal image, and multimodal prompts. BC-Z (\cite{jang2022bc}) and Vid2Robot (\cite{jain2024vid2robot}) introduce a video-conditioned policy that uses human video as task instructions. DALL-E-Bot (\cite{Kapelyukh_2023}) employs DALL-E to generate target images for tasks and generates actions for manipulation by combining the target image with the observation image. VIMA (\cite{jiang2023vima}) and MIDAS (\cite{li2023mastering}) observe that many robot manipulation tasks can be represented as multimodal prompts intertwining language and image/video frames. They construct multimodal prompts manipulation datasets and utilize pre-trained language foundation models for fine-tuning to control robot outputs. MUTEX (\cite{shah2023mutex}) extends instruction to various modalities and develops speech-conditioned, speech-goal-conditioned, image-goal-conditioned, and text-goal-conditioned.

Language-conditioned general-purpose policies remain the predominant paradigm in current research. RT-2 (\cite{rt2})  refers to this approach as \textbf{V}ision-\textbf{L}anguage-\textbf{A}ction (VLA). Following this naming convention, we divide the policy into \textbf{V}ision-\textbf{L}anguage-\textbf{A}ction-\textbf{K}ey-\textbf{P}ose (VLAKP), \textbf{V}ision-\textbf{L}anguage-\textbf{A}ction-\textbf{D}ense-\textbf{P}ose (VLADP). VLAKP is more similar to traditional object/action-centric approaches. On the other hand, VLADP resembles traditional end-to-end methods. Recent studies have explored the use of foundation models to directly synthesize low-level policy code (\cite{liang2023code, yoshida2025text}). Such policies generate executable code for robots, enabling fine-grained human inspection and debugging. We denote this paradigm as \textbf{V}ision-\textbf{L}anguage-\textbf{A}ction-\textbf{C}ode (VLAC).

The above methods integrate foundation models into policy models to guide action generation. Currently, some approaches leverage foundation models to assist in reinforcement learning training.

\subsection{VLAC}
Code generation and program synthesis have been demonstrated to be capable of developing generalizable, interpretable policy (\cite{trivedi2021learning}. However, a robot capable of generating code for multiple tasks, rich knowledge across various domains is essential (\cite{ellis2023dreamcoder}). Therefore, scholars aim to apply the prior knowledge of LLM to code generation task (\cite{chen2021evaluating,austin2021program}). Code-As-Policy (\cite{liang2023code}) demonstrates the possibility of using LLMs to directly generate code for robot execution based on prompts. The study shows that 1) code-writing LLMs enable novel reasoning capability, such as encoding spatial relationships by leveraging familiarity with third-party libraries and 2) hierarchical code-writing inspired by recursive summarization improves code generation. From-text-to-motion (\cite{yoshida2025text}) translates descriptions of human actions into humanoid robot motion code, enabling it to perform various tasks autonomously. 

\subsection{VLAKP}
The utilization of foundation models to generate key poses can be categorized into two approaches: 1) Directly using existing foundation models to output key poses. 2) Training RFMs to generate key poses through imitation learning.

Utilizing foundation models pre-trained on existing large-scale internet datasets enables the direct perception of observation images and outputting key poses. Instruct2Act (\cite{huang2023instruct2act}) utilizes CLIP and SAM to identify manipulated objects within an observation image and outputs the 3D position of these manipulated objects. Voxposer (\cite{huang2023voxposer}) utilizes LLMs to generate code that interacts with VLMs to produce affordance maps and constraint maps, collectively referred to as value maps, grounded in the robot's observation space. These composed value maps serve as objective functions for motion planners to synthesize trajectories for robot manipulation. ReKep (\cite{huang2024rekep}) uses VFM and VLM to extract relational keypoint constraints from language instructions and RGBD observations. It then applies an optimization solver to generate a series of end-effector poses.

As for imitation learning methods, CLIPort (\cite{shridhar2021cliport}) demonstrates the capability of imitation learning in language-conditioned general manipulation. However, CLIPort (\cite{shridhar2021cliport}) addresses 4-DoF end-effector pose prediction by treating it as a pixel classification problem. Keypoint-based approaches are extended to handle 6-DoF end-effector pose prediction. Due to keypoint-based methods primarily focus on 3D scene-to-action tasks, these methods become computationally expensive as resolution requirements increase (\cite{ke20243d}). To address high spatial resolution, PerAct (\cite{shridhar2023perceiver}) uses the latent set self-attention of Perceiver (\cite{jaegle2021perceiver}), which has linear complexity with voxels. Act3D (\cite{gervet2023act3d}) represents scenes as a continuous 3D feature field, transforming 2D model features into 3D feature clouds using sensed depth and learns a 3D feature field of arbitrary spatial resolution through recurrent coarse-to-fine point sampling.

Some research has extended the work on PerAct (\cite{shridhar2023perceiver}) and Act3D (\cite{gervet2023act3d}). ChainedDiffuser (\cite{xian2023chaineddiffuser}) builds upon Act3D (\cite{gervet2023act3d}) by replacing the motion planner with a diffusion model. This approach addresses the challenges of continuous interaction tasks. The 3D Diffuser Actor (\cite{ke20243d}), similar to Act3D (\cite{gervet2023act3d}), employs tokenized 3D scene representations. However, unlike Act3D (\cite{gervet2023act3d}) and 3D Diffusion Policy (\cite{ze20243d}) with 1D point cloud embeddings, 3D Diffuser Actor (\cite{ke20243d}) leverages CLIP to extract features from 2D images and aggregates them into a 3D scene representation. GNFactor (\cite{ze2023gnfactor}) improves upon PerAct (\cite{shridhar2023perceiver}) by enhancing 3D semantic features. It achieves this by distilling pre-trained semantic features from 2D foundation models into Neural Radiance Fields (NeRFs). DNAct (\cite{yan2024dnact}) builds on GNFactor (\cite{ze2023gnfactor}) and transforms the perceiver model into a diffusion head. VoxAct-B (\cite{liu2024voxact}) uses VLM to divide the task into subtasks for the left arm and the right arm and applies PerAct (\cite{shridhar2023perceiver}) to generate separate key poses for each arm.

The current imitation learning approaches also include methods using large-scale LLM/VLM. LEO (\cite{huang2023embodied}) expands upon language foundation models by incorporating modalities like images and 3D point clouds. It fine-tunes manipulation datasets using the LoRA method. This showcases the ability to transfer the original foundation model to more modalities and manipulation tasks.  \cite{xu2024manifoundation} considers the object motion produced by LLM/VLM, the object’s physical properties, and the end-effector’s design and creates a ManiFoundation model to generate the key pose. However, the key pose output by the ManiFoundation model is not 6D pose. Instead, it provides the positions of multiple contact points and the force to be applied at each contact point. 3D-VLA (\cite{zhen20243d}) can generate the final state image and point cloud based on user input. This goal state is then used to create key poses in 3D VLA.

\begin{figure*}[h]
\centering
\includegraphics[width=\linewidth,height=0.35\textheight]{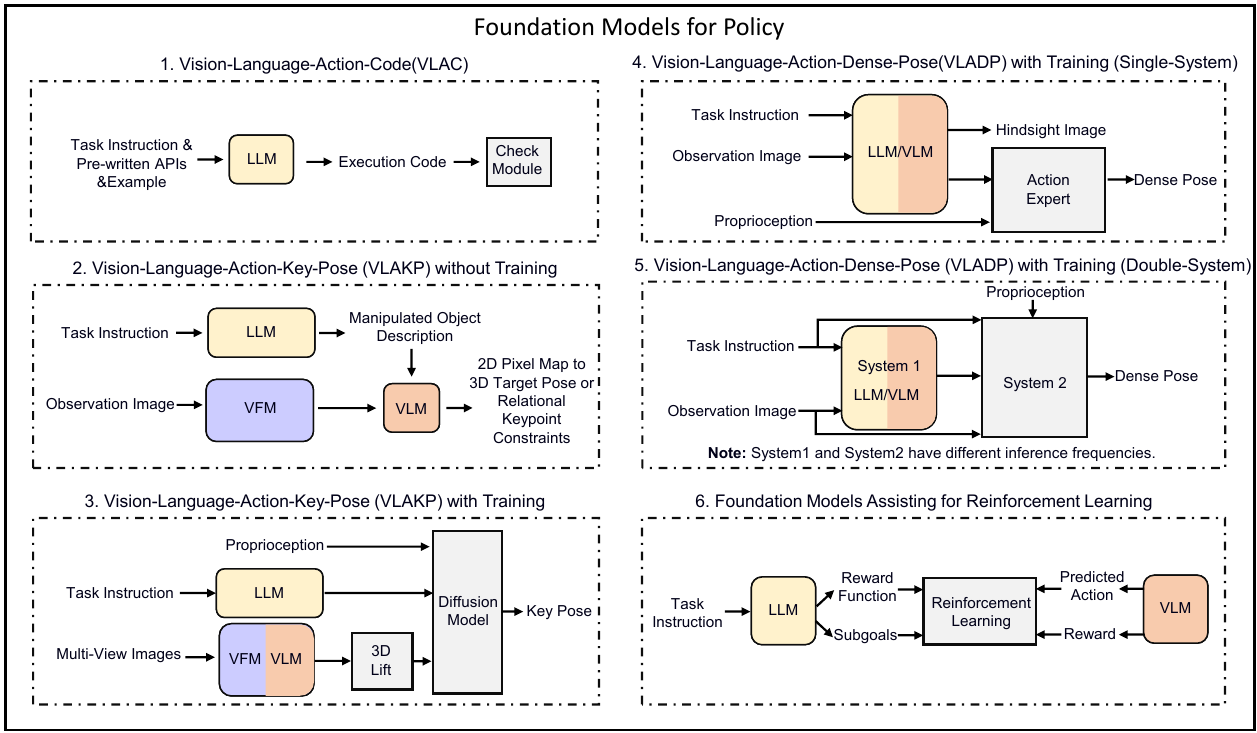}
\caption{Foundation Models for Policy. VLAC integrates task instruction, pre-written APIs, and example inputs into LLM. It generates corresponding execution code (\cite{liang2023code, yoshida2025text}). 2) VLAKP without foundation models training inputs task instruction into LLM, which specifies the manipulated object. The observation image is fed into VFM for object segmentation or keypoints, and both the manipulated object and object feature images are input into VLM, outputting the pixel mapping of the object to be manipulated into cartesian 3D key pose (\cite{huang2023instruct2act}) or relational keypoint constraints (\cite{huang2024rekep}). 3) VLAKP with training uses LLM to extract tokens from Task Instructions. VFM or VLM extract features from Multi-View Images and generate 3D features through 3D Lift. Finally, these features, along with proprioception, are used as input to the Diffusion Model to generate the key pose (\cite{ke20243d}).  4) VLADP with training (Single-System) outputs dense pose and the hindsight image by inputting task instruction and observation into a pre-trained model after training (\cite{black2410pi0}). The difference between VLADP and VLAKP with training lies in generating dense poses directly through policies, allowing for conversion into trajectory through time sequences, offering a more end-to-end approach compared to key pose. Key pose often requires subsequent motion planning. Outputting dense pose resembles more closely the paradigm of human task execution, as it does not require camera and spatial calibration or robot body configuration. However, it still necessitates extensive data training to embed the parameters of robot execution in the policy model’s hidden layer. As for imagining after the next movement, predicting both the next action and the hindsight image can improve the performance (\cite{bousmalis2023robocat}). 5) The VLADP with training (Double-System) generates dense pose by using models with different inference frequencies (\cite{helix2025}). This method can effectively leverage prior knowledge from VLM and improve inference frequency. 6)Foundation Models assisting for Reinforment Learning. LLM generates subgoals based on task instruction to transform long-horizon tasks into short-horizon ones (\cite{di2023towards}), facilitating RL learning. LLM also creates a reward function for RL according to task instruction (\cite{ma2023eureka}), while VLM can utilize prior knowledge to provide predicted action and sparse/dense reward, enhancing the effective exploration in reinforcement learning.(\cite{ye2023foundation}).}
\label{fig:Foundation_Models_For_Policy} 
\end{figure*}

\begin{table*}[htbp]
\centering
\caption{\textbf{Strengths and Limitations of VLAC, VLAKP, and VLADP}}
\label{tab:policy_comparison}
\renewcommand{\arraystretch}{0.9} 
\scriptsize
\begin{tabularx}{\textwidth}{>{\centering\arraybackslash}m{0.12\textwidth} X X}
\hline
\textbf{Policy} & \textbf{Strengths} & \textbf{Limitations} \\
\hline
\multirow{2}{*}{\textbf{VLAC}} &
\begin{itemize}[leftmargin=*, nosep, itemsep=0pt]
    \item  There are many Internet code datasets available. These can serve as priors for code generation. This helps improve the generalization of VLAC-style methods (\cite{liang2023code}).
    \item VLAC facilitates human attention to fine-grained details, enhancing the debugging process.
    \item VLAC is able to generate more creative actions, as shown in \cite{yoshida2025text}.
\end{itemize} &
\begin{itemize}[leftmargin=*, nosep, itemsep=0pt]
    \item The use of third-party libraries in VLAC can constrain the extensibility of policy tasks.
    \item Syntax errors and similar simple faults tend to appear frequently..
\end{itemize} \\

\textbf{VLAKP} &
\vspace{-0.8\baselineskip}
\begin{itemize}[leftmargin=*, nosep, itemsep=0pt]
    \item Compared to VLADP, VLAKP has a key advantage. It can leverage rich prior knowledge from foundation models in CV and NLP. This supports object-centric and action-centric action generation. As a result, VLAKP requires less training data and generalizes better than VLADP (\cite{ke20243d}).
    \item VLAKP improves upon VLAC by replacing sequential chaining with spatial composition under joint optimization. This flexibility supports more diverse manipulation tasks and ensures more stable execution (\cite{huang2023voxposer}).
\end{itemize}
\vspace{-0.8\baselineskip} &
\vspace{-0.8\baselineskip}
\begin{itemize}[leftmargin=*, nosep, itemsep=0pt]
    \item Mostly relies on 3D inputs, sensitive to pose errors. Requires precise scene reconstruction and calibration.
    \item VLAKP often requires motion planning. However, motion planning is not always reliable.
\end{itemize} \vspace{-0.8\baselineskip} \\

\textbf{VLADP} &
\vspace{-0.8\baselineskip}
\begin{itemize}[leftmargin=*, nosep, itemsep=0pt]
    \item Compared with VLAKP, it does not require calibration or motion planning.
    \item It is more end-to-end, making it easier to transfer learned priors to other tasks.
    \item Compared with other methods, VLADP has a smaller data annotation burden.
\end{itemize} \vspace{-0.8\baselineskip} & \vspace{-0.8\baselineskip}
\begin{itemize}[leftmargin=*, nosep, itemsep=0pt]
    \item It relies on large-scale training data, but faces data scarcity issues.
    \item It typically has a large number of parameters, requiring significant computational resources and leading to high latency.
    \item It shows poor generalization across different embodiments.
\end{itemize} \vspace{-0.8\baselineskip} \\
\hline
\end{tabularx}
\label{tab:comparise}
\end{table*}

\subsection{VLADP}
The policy model for outputting dense pose resembles more closely the paradigm of human task execution, as it does not require camera and spatial calibration or robot body configuration. Instead, it takes observation images as input and directly outputs the direction and magnitude of the next waypoint. While this approach is more end-to-end, it still necessitates extensive data training to embed the parameters of robot execution in the policy model's hidden layers.

Effective robotic multi-task learning necessitates a high-capacity model, hence Gato (\cite{reed2022generalist}) and RT-1 (\cite{rt1}) devise transformer-based architectures. Nonetheless, RT-1 and Gato differ; RT-1's input lacks proprioception from the robot body, while Gato incorporates proprioception. Building upon Gato, RoboCat  (\cite{bousmalis2023robocat}) demonstrates that a large sequence model can learn unseen tasks through few-shot learning. It proposes a simple but effective self-improvement process. Additionally, it shows that predicting both the next action and the hindsight image after executing that action can enhance performance. Building upon RT-1, RoboAgent (\cite{bharadhwaj2023roboagent}) enhances model generalization and stability through data augmentation and action-chunking. MOO (\cite{stone2023open}) leverages Owl-ViT to extract object locations from observation images, enhancing RT-1's open-set detection capability. 

As for multi-task reinforcement learning. PI-QT-Opt (\cite{lee2023pi}) leverages a large-scale, multi-task dataset and employs a model-free off-policy reinforcement learning approach for training. Q-Transformer (\cite{q-transformer}) facilitates training high-capacity sequential architectures on mixed-quality data by applying transformer models to RL.

Utilizing pre-trained VLMs (\cite{zhang2024vision}) for fine-tuning to construct RFMs is considered efficient. RT-2 (\cite{rt2}) collects manipulation trajectory data and fine-tunes manipulation datasets using VLM models like PaLI-X (\cite{chen2023pali}) and PaLM-E (\cite{palme}) after treating pose as tokens. However, democratizing such an expensive framework for all robotics practitioners proves challenging as it relies on private models and necessitates extensive co-fine-tuning on vision-language data to fully exhibit effectiveness. Consequently, there is an urgent need within robot communities for a low-cost alternative solution, hence RoboFlamingo (\cite{li2023vision}) and OpenVLA (\cite{kim2024openvla}) emerge, effectively enabling a robot manipulation policy with VLMs.

However, this approach necessitates a lot of data for the hidden layers to learn parameters related to the robot body, objects, and environment. Open X-Embodiment (\cite{RTX}) assembles a dataset from 22 different robots, demonstrating 527 skills (160266 tasks). However, the current Open X-Embodiment dataset faces the heterogeneity dataset challenge. Octo (\cite{team2024octo}) and HPT (\cite{wang2024scaling}) propose multi-module networks to address this issue.  RDT-1B (\cite{liu2024rdt}) and PI (\cite{black2024pi_0}) propose unified action space to address this issue. The Open X-Embodiment dataset lacks in-the-wild scenes. DROID (\cite{khazatsky2024droid}) introduces an “in-the-wild” robot manipulation dataset. It contains 76k trajectories, or 350 hours of interaction data. This data is collected from 564 scenes, 86 tasks, and 52 buildings over 12 months.

Internet videos contain information on the physics and dynamics of the world, some studies have explored training foundation models using both video datasets and manipulation data. SuSIE (\cite{black2023zero}) uses an image-editing diffusion model. This model is fine-tuned on human videos and robot rollouts. It acts as a high-level planner by proposing intermediate subgoals. These subgoals can be accomplished by a low-level controller. The two training steps of SuSIE do not share weights. GR-1 (\cite{wu2023unleashing}) is initially trained on a large-scale video dataset for video prediction, and then seamlessly fine-tuned with manipulation data. GR-2 (\cite{cheang2024gr}) uses VQGAN to convert each image into discrete tokens and is trained with a larger text-video dataset than GR-1. LAPA (\cite{ye2024latent}) begins by extracting the latent delta action between video frames. It then labels the video dataset with this information. These labeled datasets are used to train a VLM network. Finally, a small-scale robot manipulation dataset is applied for fine-tuning, enabling the mapping of latent actions to robot actions. Go-1 (\cite{bu2025agibot}) trains latent actions using human videos and combines those with manipulation data.  Then, it trains the VLA on the merged dataset to boost the model’s generalization.

Previous studies, such as GR-1 (\cite{wu2023unleashing}) and GR-2 (\cite{cheang2024gr}), train the policy head using MSE regression. In contrast, OpenVLA (\cite{kim2024openvla}) and RT-2 (\cite{rt2}) apply next-token prediction for their policy head. Building on the success of diffusion policy (\cite{chi2023diffusion}), PI0 (\cite{black2410pi0}) and TinyVLA (\cite{wen2024tinyvla}) adopt diffusion head as its policy head, achieving better performance than OpenVLA. In order to address higher degree of multi-modality in the distribution of feasible actions for bimanual manipulation, RDT-1B (\cite{liu2024rdt}) utilizes Diffusion Transformers (DiTs) as its scalable backbone network. 

Although the diffusion policy can represent complex continuous action distributions, OpenVLA-OFT (\cite{kim2025fine}) has shown in dual-arm tasks that fine-tuning the VLA with an L1 regression objective achieves performance similar to diffusion-based fine-tuning. However, it offers faster training convergence and inference speed. FAST (\cite{pertsch2025fast}) proposes a new compression-based tokenization scheme for next-token prediction. This method matches the performance of diffusion VLAs, while reducing training time by up to 5x across multiple dexterous manipulation tasks.

Previous methods face a basic tradeoff: VLM backbones are general but slow, while robot visuomotor policies are fast but not general. Synchronizing both does not improve inference speed. Helix (\cite{helix2025}) and Groot N1 (\cite{bjorck2025gr00t}) overcome this tradeoff with two asynchronous complementary systems, trained end-to-end to communicate. However, Groot N1 makes more use of human video latent action and simulated data compared to Helix.

\subsection{Foundation Models assisting for Reinforcement Learning}\label{subsec:assistRF}
Reinforcement learning has garnered widespread attention from researchers due to its ability to explore the environment by not requiring extensive annotated data. However, it also faces numerous challenges, such as dealing with long-horizon sequences, effectively exploring, reusing experience data, and designing reward functions (\cite{kober2013reinforcement}). Foundation models have demonstrated the emergence of common sense reasoning, the ability to sequence sub-goals and visual understanding. Due to the strong capability of foundation models, many studies aim to leverage the unprecedented capability of foundation models to address the challenges faced by reinforcement learning. RobotGPT (\cite{jin2024robotgpt}) aims to distill the knowledge of the brain ChatGPT into the mind of a small brain trained with reinforcement learning. At the same time, many studies explore the use of foundation models to solve challenges like long-horizon problems and effectively exploring and designing reward functions.

Norman (\cite{di2023towards}) employs LLMs to decompose tasks into subgoals and utilizes CLIP to identify the completion of each subgoal, serving as a signal generator for sparse rewards. ROBOFUME (\cite{yang2024robot}) employs a fine-tuned VLM as the sparse reward function for the RL algorithm, tackling the issue of the extensive human supervision needed for training or fine-tuning a policy in the real world. Eureka  (\cite{ma2023eureka}) utilizes LLM to craft a reward function for five-fingered hand pen spinning. Subsequently, it engages in a cyclic process encompassing reward sampling, GPU-accelerated reward evaluation, and reward reflection to progressively refine its reward outputs. In contrast to Eureka's self-iteration and sparse reward function design, TEXT2REWARD  (\cite{xie2023text2reward}) incorporates human feedback into the iterative updating of the reward function, yielding a dense reward function. FAC (\cite{ye2023foundation}) proposes using knowledge from foundation models as policy prior knowledge to improve sampling action efficiency, as value prior knowledge to measure the values of states and as success-reward prior knowledge to provide final feedback on task success.

\subsection*{Summary}
According to Fig. \ref{fig:Foundation_Models_For_Policy}, policies can be classified into VLAC, VLAKP, VLADP, and Foundation Models assisting for Reinforcement Learning. As shown in \ref{sec: appen_a} Tab. \ref{tab:comparasion_1} and Tab. \ref{tab:comparasion_2}, comparison with baseline approaches reveals key distinctions. The strengths and limitations of VLAC, VLAKP, and VLADP are as described in Tab. \ref{tab:comparise}.

\section{Manipulation Data Generation}\label{sec:generation}
Demonstration data plays a crucial role in robotic manipulation, particularly in the context of imitation learning (\cite{RTX}). A common approach for gathering such demonstrations is human teleoperation in the real world. However, collecting real-world data often requires significant human labor and specialized teleoperation equipment. Recently, there has been a growing number of excellent developments in low-cost teleoperation hardware, which enables the collection of high-quality demonstration data (\cite{fang2024airexo,cheng2024open}).

To collect data in real environments, human effort is required for scene setup and data annotation (\cite{sermanet2023robovqa}). There are currently two methods for data collection: the bottom-up approach and the top-down step-by-step approach. The bottom-up approach focuses on selecting a task to perform based on the current scene. Then, it uses methods like crowd-sourcing to label the data. The top-down approach involves a step-by-step process where decision-makers assign task labels and manage overheads, such as resets and scene preparations (\cite{sermanet2023robovqa}). The robot then performs tasks according to these labels. RoboVQA (\cite{sermanet2023robovqa}) shows that the bottom-up approach is more efficient in data collection compared to the top-down step-by-step approach.  DIAL (\cite{xiao2022robotic}) uses a fine-tuned CLIP to replace humans in labeling robot trajectories during bottom-up data collection. This transforms the robot manipulation dataset on the internet into the robot-language manipulation dataset. PAFF (\cite{ge2023policy}) points out that incorrect robot trajectories can be linked to new tasks and uses fine-tuned CLIP to label the incorrect robot trajectories with appropriate task labels. The above methods demonstrate that high-level cognitive models can assist in data annotation. SOAR (\cite{zhou2024autonomous}) shows that integrating a high-level cognitive model with a low-level control policy can result in a fully autonomous data-collection system in  varied real-world environments. 

Generating lots of data in simulation is a cheaper solution. However, it still requires human effort to create both scene generation and task execution code for specific tasks (\cite{wang2023gensim}). Moreover, the notorious sim-to-real gap issue remains a challenge in transferring policies trained in simulation to real-world applications. But there are many methods to address the sim-to-real challenge. \cite{matas2018sim} trains the policy fully in simulation through domain randomization and then successfully deployed in the real world, even though it has never encountered real deformable objects. Therefore, simulation plays an important role in manipulation and this section will analyze existing simulators, scene generation, demonstration generation and sim-to-real gap challenge. 

Compared to single-frame images and language data on the internet, internet videos contain information on the physics and dynamics of the world, as well as on human behaviors and actions (\cite{chandrasegaran2024hourvideo}). This information is precisely what is required for manipulation tasks. Therefore, in this section, we also introduce the internet-scale video data for robot learning.

Regardless of whether it's in a real or simulated environment, improving the efficiency of the existing dataset is essential. The mainstream approach is dataset augmentation.

\begin{table*}[htbp]
\caption{\textbf{Representative Low-cost Hardware Works.}}
\begin{threeparttable}
\resizebox{\textwidth}{!}{
\begin{tabular}{c c c c c c c}
\hline
\Large \textbf{Teleoperation Device} & \Large \textbf{Teleoperation Type}  & \Large \textbf{Human Motion Measurement} & \Large \textbf{Feedback} & \Large \textbf{Cost} & \Large \textbf{Embodiment Config} & \Large \textbf{Manipulation Task} \\ \hline 

\rowcolor{mygray}
\Large Aloha(\cite{zhao2023learning}) & \Large Online Teleoperation & \Large Joint Copy & \Large Third-person View & \Large \$2000(include robot) & \Large Dual-Arm; 2F Gripper & \Large  \begin{tabular}[c]{@{}c@{}}Slide Ziploc;\\ Slot Battery;\\ Open Cup; etc;\end{tabular} \\ 

\Large GELLO(\cite{wu2023gello}) & \Large Online Teleoperation & \Large Joint copy & \Large Third-person View & \Large \$300 & \Large Dual-Arm; 2F Gripper & \Large \begin{tabular}[c]{@{}c@{}}Place Hat on Rack;\\ Hand Over;\\ Fold Towel; etc;\end{tabular} \\ 

\rowcolor{mygray}
\Large Human Plus(\cite{fu2024humanplus}) & \Large Online Teleoperation & \Large End-effectors mapping & \Large Third-person View & \Large \$30(A RGB Camera) & \Large Humanoid Robot; Dexterous Hand & \Large \begin{tabular}[c]{@{}c@{}}Wear Shoe and Walk;\\ Fold Clothes;\\ Warehouse; etc;\end{tabular} \\

\Large Transteleop(\cite{li2020mobile})\ & \Large Online Teleoperation &\Large  End-effectors mapping & \Large Third-person View & \Large \$170 & \Large Dual-Arm; Dexteours Hand & \Large \begin{tabular}[c]{@{}c@{}}Pick-Place; \\ Handover; Push; etc;\end{tabular} \\

\rowcolor{mygray}
\Large OPEN TEACH(\cite{iyer2024open}) & \Large Online Teleoperation & \Large End-effectors mapping & \Large First-person View  & \Large \$500 & \Large Dual-Arm; Dexteours Hand/2F Gripper; & \Large \begin{tabular}[c]{@{}c@{}}Make Sandwich; \\ Ironing Cloth; Open Cabinet; etc;\end{tabular} \\

\Large Open-TeleVision(\cite{cheng2024open}) & \Large Online Teleoperation & \Large End-effectors mapping & \Large First-person View & \Large \$3499 & \Large Humanoid Robot; Dexteours Hand & \Large \begin{tabular}[c]{@{}c@{}}Can Sorting;\\ Can Insertion;\\ Folding;\\ Unloading\end{tabular} \\

\rowcolor{mygray}
\Large AirExo (\cite{fang2024airexo}) & \Large Offline Teleoperation & \Large Joint copy & \Large - & \Large \$600 & \Large Dual-Arm; 2F Gripper & \Large \begin{tabular}[c]{@{}c@{}}Gather Balls;\\ Grasp from the Curtained Shelf;\end{tabular} \\ 

\Large UMI(\cite{chi2024universal}) & \Large Offline Teleoperation & \Large End-effectors mapping & \Large - & \Large \$371 & \Large Dual-Arm; 2F Gripper & \Large \begin{tabular}[c]{@{}c@{}}Dish Washing;\\ Dynamic Tossing; Cloth Folding; etc;\end{tabular} \\

\rowcolor{mygray}
\Large DexCap(\cite{wang2024dexcap}) & \Large Offline Teleoperation & \Large End-effectors mapping & \Large - & \Large \$4000 & \Large Dual-Arm; Dexterous Hand & \Large \begin{tabular}[c]{@{}c@{}}Scissor Cutting;\\ Tea Preparing; Sponge Picking; etc;\end{tabular} \\

\Large VideoDex(\cite{shaw2023videodex}) & \Large Offline Teleoperation & \Large End-effectors mapping & \Large - & \Large - & \Large Single-Arm; Dexterous Hand & \Large \begin{tabular}[c]{@{}c@{}}Pick-Place;\\ Cover and Uncover;etc;\end{tabular} \\

\end{tabular}}
\end{threeparttable}
\label{tab:low_cost}
\end{table*}

\subsection{Low-cost Teleoperation Device}
The current low-cost teleoperation can be categorized into two types: online teleoperation and offline teleoperation. The distinction is similar to the difference between SLAM and SFM. Online teleoperation is a closed-loop interaction between a demonstrator and a robot (\cite{darvish2023teleoperation}). In the forward process, human motion is measured using devices that combine various sensors, such as vision, IMUs, or multi-joint encoders. The motion data from the demonstrator is then retargeted to the robot's space. This allows the robot to accurately follow the demonstrator's demonstrated trajectory. During the backward process, sensor data from the robot, such as forces, torques, and tactile information, should be retargeted to the demonstrator's space. As a result, the demonstrator can experience an immersive teleoperation environment by sensor data feedback. At the same time, the synchronization and real-time performance between the forward and backward processes are also crucial (\cite{darvish2023teleoperation}). Offline teleoperation remove the reliance on real robots during data collection compared to online teleoperation (\cite{chi2024universal}). Demonstrators directly perform tasks using handheld or wearable devices (\cite{fang2024airexo, chi2024universal, wang2024dexcap}) or using cameras to record the task execution process (\cite{shaw2023videodex}). They do not need to supervise real robots to complete the tasks and operate tasks using human's direct view perspective. Therefore, offline teleoperation lacks the backward feedback process. Without relying on real robots, the devices become more portable and intuitive. However, this increases the precision requirements for the retargeting algorithm.

The differences among current low-cost teleoperation devices lie primarily in two aspects. One is human motion measurement on both online teleoperation and offline teleoperation. The other is visual feedback on online teleoperation. Human motion measurement component can be broadly categorized into two classes: one aimed at capturing and mapping the pose of end-effectors (\cite{cheng2024open, liu2022robot, fu2024humanplus, chi2024universal, li2020mobile}), and one exploited devices for joint copy (\cite{zhao2023learning, wu2023gello, fang2024airexo}). Visual feedback can be generally classified into two types third-person view and first-person view (\cite{cheng2024open}). The third-person view shows the demonstrator from an external position, offering a broader perspective of surroundings. In contrast, the first-person view mimics the robot’s perspective, providing an immersive and realistic experience such as teleoperation with VR/AR headset.

For approaches capturing and mapping the pose of end-effectors, the common low-cost capturing devices include SpaceMouse (\cite{liu2022robot, zhu2023viola}), cameras (\cite{cheng2024open, fu2024humanplus, iyer2024open, shaw2023videodex, li2019vision, fang2020vision}), VR controllers (\cite{de2021leveraging, nakanishi2020towards}) and IMU sensors (\cite{chi2024universal, li2020mobile, fang2017novel, fang2017robotic}). The SpaceMouse based method passes the position and orientation of the SpaceMouse as action commands of end-effectors. This method is low-cost, easy operation, and easy implementation, but it is not suitable for dual-arm operations. In contrast, methods based on cameras and VR are well suited for bimanual teleoperation and VR offers the advantage of visual feedback compared to cameras. However, teleoperation methods based on cameras and VR heavily relies on the accuracy of pose estimation algorithms and often affected by occlusion (\cite{fu2024humanplus, pavlakos2024reconstructing, iyer2024open, cheng2024open}). The main advantage of teleoperation devices based on IMU sensors lies in their wearability (\cite{li2020mobile, chi2024universal, wang2024dexcap}). Due to this advantage, UMI (\cite{chi2024universal}) and DexCap (\cite{wang2024dexcap}) develope wearable devices capable of in-the-wild teleoperation and offline data collection.

Above systems work in cartesian space, which needs inverse kinematic (IK) solver and off-the-shelf IK often suffering from fails when operating near singularities of the robot. Although some bilateral teleoperation systems use haptic feedback to provide a tangible sense of the robot’s kinematic constraints, they do not address the challenges of very tight operational spaces (\cite{silva2009phantom}). Therefore, multi-joint encoder teleoperation devices can solve the IK problem by working in the joint space. The current design of multi-joint teleoperation devices is mainly divided into isomorphic and non-isomorphic devices (\cite{wu2023gello}). Isomorphic devices refer to teleoperation systems using standard servo-based robotic arms to control manipulators with similar size and kinematics (\cite{zhao2023learning}), while non-isomorphic devices use such arms to control manipulators with different size and kinematic properties. Non-isomorphic devices use kinematically equivalent structures based on DH parameters to map joint spaces between different properties (\cite{wu2023gello}). AirExo (\cite{fang2024airexo}) expands this low-cost and scalable platform into a wearable device to collect cheap in-the-wild demonstrations at scale.

As for teleoperation visual feedback, most of methods (\cite{liu2022robot, zhu2023viola, fu2024humanplus, li2020mobile, zhao2023learning, wu2023gello}) are use third side view that observe the robot task with the operator’s own eyes directly. However, this observation involves some visual errors. For example, there may be inaccuracies in the distance between the gripper and the object being manipulated. While for first-person view, due to wearing VR head (\cite{cheng2024open, iyer2024open, de2021leveraging, nakanishi2020towards}), it allows operators to perceive the robot's surroundings immersively. However, long time to use VR headset can lead to fatigue.

To collect large-scale real-world manipulation data, teleoperation devices need trajectory following, intuitive, low-cost, portable and in-the-wild capabilities. In Tab. \ref{tab:low_cost}, we summarize several representative works on low-cost hardware teleoperation. For online teleoperation, it is important to ensure synchronization and real-time performance between the forward and backward processes and the backward process should provide forces and torques feedback, as well as tactile feedback. This is essential for dexterous hand manipulation tasks. For offline teleoperation, hardware development and retargeting algorithms are critical. Once these two aspects are well-executed, the offline teleoperation devices facilitate large-scale manipulation data collection from experts in specific industries. For instance, chefs can wear exoskeleton devices while cooking to gather relevant data.

\subsection{Simulator}
The current mainstream simulators (\cite{zhou2023language}) include PyBullet (\cite{coumans2016pybullet}), MuJoCo (\cite{todorov2012mujoco}), CoppeliaSim (\cite{rohmer2013v}), NVIDIA Omniverse and Unity. Pybullet is easy to use and integrate, but its graphics are quite basic. It is not suitable for applications that require complex visual effects. Therefore, Pybullet is often used together with Blender (\cite{shi2024asgrasp}). Mujoco offers a high-precision physics engine. It is suitable for simulating articulated and deformable object manipulation. However, it has a high entry barrier for beginners. CoppeliaSim offers a wide range of ready-made environments, objects, and prototyping robotic systems for users. However, when dealing with many robots or complex scenes, CoppeliaSim may encounter performance issues. NVIDIA Omniverse provides real-time physics simulation and realistic rendering. However, it requires significant computational resources. NVIDIA Omniverse offers many interfaces. Users can use these to develop various applications. For example, Issac Gym is a platform for robot reinforcement learning, developed using Omniverse. Unity offers rich visual effects and a user-friendly interface. It allows for the creation of highly interactive applications. However, its physics engine is still not precise enough. The basic components of a simulator are the physics engine and the renderer. Improvements in these components can enhance the capability of sensors in simulations, such as optical tactile sensors (\cite{chen2023tacchi}). Learning-based simulators also show great potential. For example, Sora (\cite{videoworldsimulators2024}) and UniSim (\cite{yang2023learning}) use vast amounts of data from the internet to simulate the visual effects of many different actions.

\subsection{Scene and Demonstration Generation}\label{subsec:scene_generation}
Simulation scenes can be created manually. However, this approach is time-consuming and labor-intensive. As a result, automated or semi-automated scene generation methods are more preferred (\cite{deitke202}). Two methods can be used. Real-to-Sim method converts real scenes to simulation. Automated generation method automatically generates simulation scenes without real-world observation. Real-to-Sim method can accurately mimic the real world, but it limits the diversity of scenes. The automated generation method can create more diverse scenes and increase the variety of collected demonstrations. 

The Real-to-Sim method directly refers to a digital twin. The Real-to-Sim method utilizes 3D-reconstruction technology (in Sec. \ref{sec:state}) or inverse graphics (\cite{chen2024urdformer}) to create the real-world scene in a virtual environment (\cite{torne2024reconciling}). But, 3D reconstruct scene is static environment where objects lack real-world physical properties, such as material, mass and friction coefficients, and are non-interactive (\cite{torne2024reconciling}). The inverse graphics method, such as URDFormer (\cite{chen2024urdformer}), directly generates interactive simulation environment and articulated objects from input RGB image. Compared to 3D-reconstruction methods, it reduces human involvement and produces interactive simulation environment. However, it lacks physical plausibility and fails to address the mismatch between the generated object's physical properties and the real world.

As for the application of foundation models in Real-to-Sim methods, GRS (\cite{zook2024grs}) employs SAM2 for object segmentation from RGBD image and utilizes VLMs to describe and match objects with simulation-ready assets. This approach combines the strengths of 3D-reconstruction and inverse graphics methods. It ensures the credibility of 3D-reconstruction methods and allowing objects in the scene to interact. However, it is impossible for the assert dataset to fully cover objects in the real world. Constructing an interact assert dataset often requires manual design by the creator or human-assisted interactive object generation. ACDC (\cite{dai2024automated}) defines a digital cousin concept. Unlike a digital twin, it does not directly replicate a real-world counterpart. However, it retains similar geometric and semantic features by using similar asserts when the assert dataset does not include real-world objects. As for object pose, depth cameras are commonly used, but they struggle to capture reflective surfaces accurately. This limits their use in the wild. To address this issue, ACDC uses Depth-Anything-v2 (\cite{yang2024depth}), a state-of-the-art monocular depth estimation model, to estimate the depth map.

Scene diversity primarily includes the diversity of scene
layouts, such as floor plans and object placements, as
well as the diversity of objects. The automated generation methods are more
effective for producing large-scale diverse scenes. The automated generation methods can be categorized into rule-based and learning-based approaches. For instance, ProcTHOR (\cite{deitke202}) introduces a procedural generation pipeline for interactive scenes using rule-based constraints and statistical priors. However, the generated scenes often rely on pre-defined priors, resulting in unrealistic outcomes that hinder agent learning (\cite{khanna2024habitat}). To address this, PHYSCENE (\cite{yang2024physcene}) incorporates physical collision avoidance, object layouts, interactivity, and reachability metrics into a diffusion model. This approach enhances the physical plausibility and interactivity of generated scenes. 

Due to the prior knowledge of foundation models, there are current efforts to use foundation models for scene construction. RoboGen (\cite{wang2023robogen}) utilizes LLM to generate relevant assets, asset sizes, asset configuration, scene configuration based on the task proposals and use text-to-image-to-3D generation to create the corresponding assets. These assets are imported into the simulator to generate the appropriate scene. Finally, using VLM for task-specific scene verification. GenSim (\cite{wang2023gensim}) uses LLMs to generate new task and task scenario codes based on the pre-cached scene codes in a task library. However, using foundation models to automate the generation of scene's physical plausibility still relies on VLM for judgment. At the same time, the above research also uses LLMs to generate diverse instructions to ensure task diversity. However, generating diverse task instructions with LLMs presents challenges in ensuring rationality for the current environment.

The Real-to-Sim method and the Automated generation method both rely on 3D assets. The diversity of 3D assets determines the variety of scenes  (\cite{nasiriany2024robocasa}). Although there are many existing 3D object assets (\cite{chang2015shapenet, deitke2023objaverse, li2023behavior, geng2023gapartnet, xiang2020sapien, liu2022akb, calli2017yale}), their quantity is far from sufficient to cover the variety of real-world objects. As a result, many studies focus on the automatic generation of assets, such as zero-1-to-3 (\cite{wang2023robogen}), Luma.ai (\cite{nasiriany2024robocasa}), LLaMA-Mesh (\cite{wang2024llama}), Trellis (\cite{xiang2024structured}). However, the performance of generative models is also limited by the shortage of current 3D training data. To address this issue, data cleaning techniques or manual supervision are needed to filter and select high-quality generated object assets.

The modeling of the interaction environment above primarily focuses on articulated object modeling. Articulated objects can be created manually by designers or generated using procedural (\cite{jiang2022ditto, liu2023paris, zhang2023part}) or human-assisted interactive methods (\cite{torne2024reconciling}) after 3D scanning. They can also be generated automatically through inverse graphics (\cite{chen2024urdformer}) or generative model (\cite{xiang2024structured}). However, current automated methods for generating articulated object assets are limited to objects with few rotational joints. Real2Code (\cite{mandi2024real2code}) fine-tunes a CodeLlama model to process visual observation descriptions and then outputs joint predictions. This enables Real2Code to reconstruct complex articulated objects with up to 10 parts. At the same time, generative models mainly focus on rigid and articulated objects and research on deformable objects remains insufficient (\cite{sundaresan2022diffcloud}).

To collect demonstrations in simulations, different approaches can be used based on task complexity. For simple tasks, like a two-finger gripper picking up a cube, a hard-coding method (\cite{wang2022bulletarm}) can be used. However, for more complex tasks, remote teleoperation (\cite{chen2024teleoperation}) or skill library (\cite{ha2023scaling}) should be employed. Building skill library can be done using reinforcement learning or gradient optimization methods. RoboGen (\cite{wang2023robogen}) shows that gradient-based trajectory optimization is better for fine-grained manipulation tasks with soft bodies, like shaping dough into a specific form. On the other hand, reinforcement learning and evolutionary strategies are more effective for contact-rich tasks and continuous interactions with other components in the scene.

\subsection{Sim-to-Real Gap Solutions}
The sim-to-real problem is a widespread issue across machine learning, not limited to manipulation (\cite{zhao2020sim}). The goal is to successfully transfer the policy from the simulation (source domain) to the real world (target domain). The gap in the manipulation tasks between the simulation and the real-world includes two main types: visual gap and dynamic gap. Visual gap refers to the difference between the vision information produced by the renderer and the vision information in the real world. In terms of rendering realism, BEHAVIOR-1K (\cite{li2023behavior}) highlights that Omniverse offers the highest rendering performance. The dynamic gap consists of several factors. First, there is a difference between the physics engine used in simulations and real-world physics. Second, the properties of objects, including robots, contribute to the object dynamic gap. Lastly, there is a control gap in robots, such as variations in static errors caused by different PID parameters. Currently, there are three main approaches to address sim-to-real gap: system identification, domain randomization, and transfer learning (\cite{zhao2020sim}).

Most of the system identification research (\cite{kristinsson1992system}) aims to create an accurate mathematical model of a physical system to make the simulator more realistic. However, it is impossible to accurately build models of complex environments in simulators. The primary methods for physical parameter identification include estimation from interaction (\cite{seker2024estimating, bohg2017interactive, xu2019densephysnet}), estimation from demonstrations (\cite{torne2024reconciling}), and estimation from observations using foundation models (\cite{gao2023physically}). Among these, estimation from demonstrations appears more effective. Demonstrations inherently contain interaction information and can also assist policy training. However, improving the hardware performance for collecting demonstrations remains essential.

Domain randomization (\cite{ramos2019bayessim}) involves adding random disturbances to the parameters in simulation. This can include various elements, generally divided into visual and dynamic randomization. Visual randomization covers visual parameters like lighting, object textures, and camera positions. Dynamic randomization covers dynamic parameters like object sizes, surface friction coefficients, object masses, and actuator force gains. By experiencing diverse simulated environments, the policy can adapt to a broad range of real-world conditions. For the policy, the real world is essentially just another disturbed environment. However, parameter randomization requires human expertise. \cite{ma2024dreureka} demonstrates that LLM excels in selecting randomized parameters and determining the randomization distribution. This makes domain randomization more automated.

Transfer learning (\cite{yu2022dexterous, tan2018survey}) involves using limited real-world data to adapt a policy trained on a abundant simulation data to the real world. Treat policies in the real-world and in the simulation as different tasks. We can use task transfer methods for transfer learning. For example, \cite{rusu2017sim} uses the progressive network to apply knowledge from a policy trained in simulation to a new policy trained with limited real-world data, without losing the previous knowledge. Treat the policies in the real-world and in the simulation as the same task, even though the data distributions differ. We can use domain adaptation methods to address this issue. Three common methods for domain adaptation are discrepancy-based (\cite{lyu2024cross}), adversarial-based (\cite{eysenbach2020off}), and reconstruction-based methods (\cite{bousmalis2016domain}). Discrepancy-based methods measure the feature distance between the source and target domains using predefined statistical metrics. This helps to align their feature spaces. Adversarial-based methods use a domain classifier to determine whether features come from the source or target domain. Once trained, the extractor can produce features that are invariant across both domains. Reconstruction-based methods also aim to find shared features between domains through setting up an auxiliary reconstruction task and using the shared features to recover the original input.

The methods discussed above assume that the target domain remains unchanged. However, many physical parameters of the same robot can change significantly. Factors like temperature, humidity, positioning, and wear and tear over time can all affect these parameters. This makes it harder to bridge the sim-to-real gap. To address this issue, DORA (\cite{zhang2024debiased}) uses an information bottleneck principle. It aims to maximize the mutual information between the dynamics encoding and environmental data. At the same time, it minimizes the mutual information between the dynamics encoding and the behavior policy actions. Transic (\cite{jiang2024transic}) proposes a data-driven approach that enables successful sim-to-real transfer using a human-in-the-loop framework.

\subsection{Internet-Scale Video Dataset}
Extensive and diverse video datasets are available from online repositories. The collection process requires querying and searching for videos with relevant content. After that, low-quality video data is removed through data cleansing. However, the raw video data cannot be directly transferred into the manipulation model due to the absence of (1) action or reward labels; (2) distribution shifts including physical embodiments, camera viewpoints, and environments. Although AVID (\cite{smith2019avid}) and LbW (\cite{xiong2021learning}) translate human action images from videos into robot action images, this type of translation remains limited to the pixel level; (3) essential low-level information like tactile feedback, force data, proprioceptive information, and depth perception (\cite{mccarthy2024towards}). However, these raw videos contain extensive visual information, such as objects, spatial information, human activities, and sequences of interactions between humans and objects (\cite{eze2024learning}). At the same time, language annotations are essential to support learning of semantic features in this visual information.

\begin{figure}[htbp]
\centering
\includegraphics[width=\linewidth]{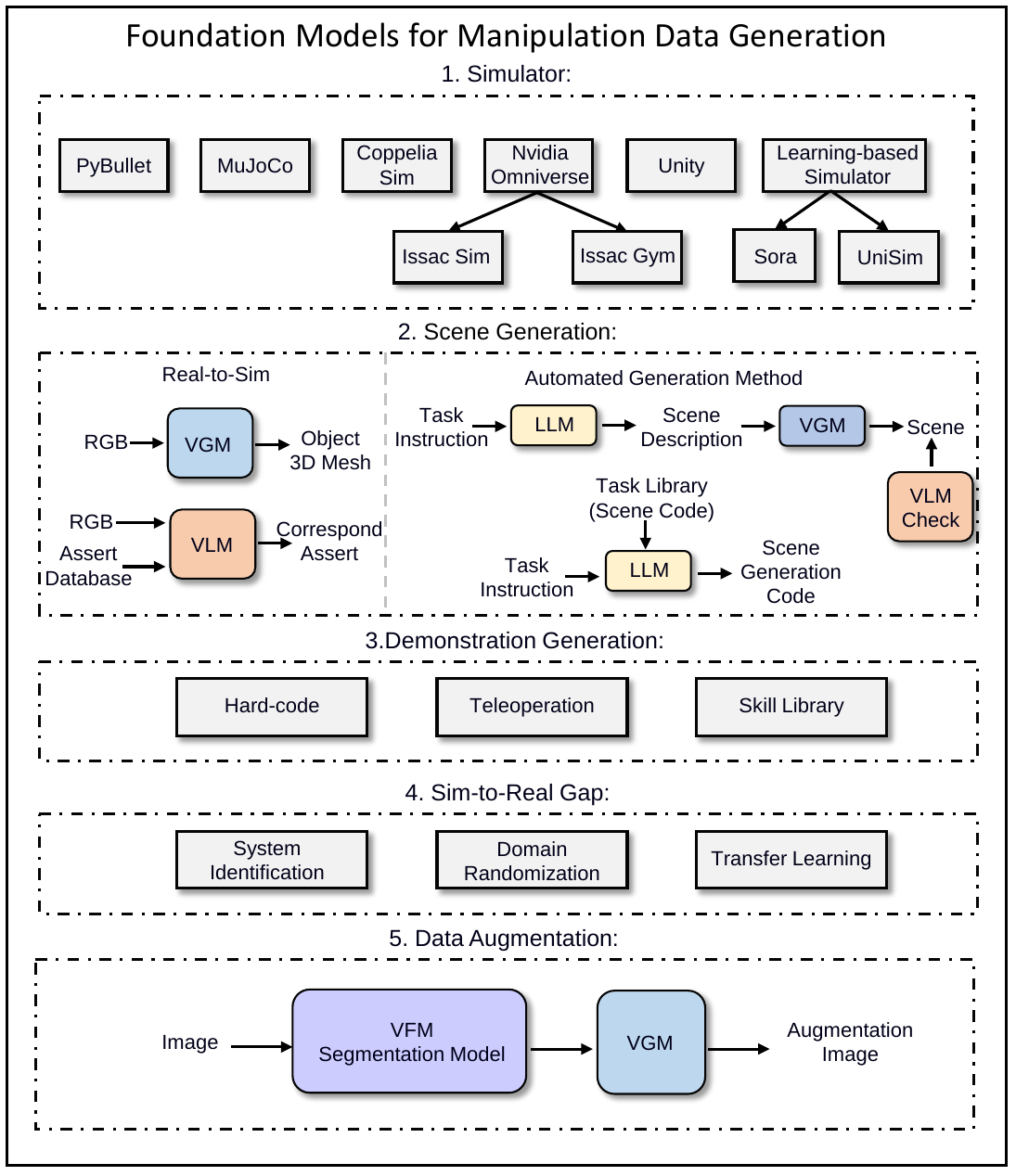}
\caption{Foundation Models for Manipulation Data Generation. Current mainstream simulators include Pybullet, MuJoCo, CoppeliaSim, NVIDIA Omniverse, and Unity. Meanwhile, learning-based generative models used as simulators have shown potential. Simulation environment generation can be classified into Real-to-Sim and Automated Generation methods. In the Real-to-Sim methods, assuming the object's position is known, the main challenge lies in constructing the object's 3D mesh. This can be achieved through scanning technique or by using VGM to generate the 3D mesh directly from RGB image (\cite{chen2024urdformer}). Additionally, GRS (\cite{zook2024grs}) utilizes VLM to extract 3D object meshes corresponding to real-world object from assert database based on RGB image. In the Automated Generation methods, LLM can output scene descriptions or scene code based on task instruction.  When the output is a scene description, VGM generates the objects and arranges them according to the description. Meanwhile, the generated scene need to be evaluated by VLM. (\cite{wang2023robogen}). When the output is scene code, it directly generates the corresponding scene (\cite{wang2023gensim}). However, this requires substantial prior knowledge of scene code within the task library. There are three methods for generating demonstrations in a scene: Hard-code, Teleoperation, and Skill Library. When building skill library, gradient optimization is effective in training skill for deformable tasks and reinforcement learning works better for contact-rich tasks (\cite{wang2023robogen}). Solutions for the Sim-to-Real gap include System Identification, Domain Randomization, and Transfer Learning. For data augmentation, VFM is used to segment images first, and then VGM renders the object's texture on the masked image.} 
\label{fig:Manipulation_Data_Generation} 
\end{figure}

Methods to obtain language annotations are divided into manual and automated captions. Manual captions are created by humans labeling video content. Automated captions include four types: (1) Automatic Speech Recognition (ASR), which converts audio in videos to text (\cite{xue2022advancing}). (2) Alt-text, which collects captions from HTML alt-text attributes in web images and videos, like descriptions, tags, and titles (\cite{bain2021frozen}). (3) Transfer, which starts with a set of image-caption pairs. Then, captions are matched to video clips with similar frames (\cite{nagrani2022learning}). (4) Foundation Models, which use pre-trained models to get captions. For example, VLMs provide single-frame image captions, while LLMs filter out inconsistent captions across frames (\cite{blattmann2023stable}). Owing to recent advancements in language annotation techniques, most widely used internet video datasets incorporate language annotations, such as InternVid (\cite{wang2023internvid}), HD-VILA-100M (\cite{xue2022advancing}), YT-Temporal-180M (\cite{zellers2021merlot}), WTS-70M (\cite{shvetsova2025howtocaption}), HowTo100M (\cite{miech2019howto100m}), WebVid-10M (\cite{nan2024openvid}), and VideoCC3M (\cite{yan2022videococa}).  At the same time, various off-the-shelf models can be used to annotate the current video with additional labels, such as pose (\cite{shaw2023videodex}), affordance (\cite{mendonca2023structured}), key points trajectory (\cite{wen2023any}), latent action (\cite{ye2024latent}), mask and bounding boxes (\cite{shan2020understanding}).

The task information contained in internet video data may not be highly relevant to the specific tasks performed by robots. Additionally, internet video data often suffers from issues such as missing action labels, low-level information, and distribution shifts. Therefore, manually recording custom videos can be an effective approach to collecting videos that are directly relevant to specific robot tasks or embodiments. This method can also help avoid the issue of re-annotating. By incorporating sensors such as IMUs, tactile sensors, and depth sensors during the recording process, manually recorded custom videos can exhibit lower noise compared to internet video data. However, the scale and diversity of manually recorded videos still cannot match the internet video data (\cite{mccarthy2024towards}). Currently, there are several commonly used manually recorded video datasets, such as Ego-4D (\cite{grauman2022ego4d}), Ego-Exo-4D (\cite{grauman2024ego}), RoboVQA (\cite{sermanet2024robovqa}), Epic-Kitchens-100 (\cite{damen2022rescaling}) and ActionSense (\cite{delpreto2022actionsense}).

\subsection{Dataset Augmentation}
Current data augmentation can be mainly divided into scene-level and object-level. Scene-level refers to changing the layout of objects in the scene. For example, MimicGen (\cite{mandlekar2023mimicgen}) and DexMimicGen (\cite{jiang2024dexmimicgen}) change the positions and orientations of objects, while CACTI (\cite{mandi2022cacti}) adds new, artificial objects to the scene. However, the
reliability of data augmentation still needs validation.  For example, MimicGen (\cite{mandlekar2023mimicgen}) filters data generation attempts based on task success. Current foundation models for dataset augmentation methods primarily operate at the object level. The main idea is to use semantic segmentation to extract masks for each object, and then employ generative rendering methods to alter the object's texture. GenAug (\cite{chen2023genaug}) leverages language prompts with a generative model to modify object textures and shapes, adding new distractors and background scenes. ROSIE (\cite{yu2023scaling}) localizes the augmentation region with an open vocabulary segmentation model and then runs image editor to perform text-guided image editing.

\subsection*{Summary}
Following Fig. \ref{fig:Manipulation_Data_Generation}, LLMs can generate credible descriptions or code for task scenes. VGMs produce 3D object meshes and render textures. Nonetheless, the validity of the generated task scenes must be ultimately assessed by VLMs. For scene generation, Automated Generation Method ensures higher diversity than Real-to-Sim. The realism of simulation depends on the simulator. Omniverse provides the best rendering performance.

\section{Discussion}\label{sec:discussion}
In this survey, we aim to outline the opportunities brought by foundation models for general manipulation. We believe the potential of embedding foundation models into manipulation tasks as a viable path towards achieving general manipulation. However, the primary applications of LLMs, VFMs, VLMs, LMMs and VGMs focus only on certain aspects of general manipulation capability, such as reasoning, perception, multimodal understanding, and data generation. The current framework for RFMs demands extensive data for learning, posing a crucial issue of constructing a data close-loop, and ensuring over a 99\% success rate remains an unresolved concern. Therefore, this paper proposes a framework of robot learning for manipulation towards achieving general manipulation capability and detailing how foundation models can address challenges in each module of the framework. However, there are still many open questions in this survey. In this section, we delve into several open questions that we are particularly concerned about.

\subsection{What is the framework for general manipulation?} 
\subsubsection{Definition of general manipulation.}
The ultimate general manipulation framework should be able to interact with human or other agent and control whole-body to  manipulate arbitrary objects in open-world scenarios and achieve diverse manipulation tasks. However, the interaction between robot and human involves not only recognizing intentions but also learning new skills or improving old skills from human experts in the external world. Open-world scenarios may be static or dynamic. Objects can be either rigid or deformable. Task objectives can vary from short-term to long-term. Furthermore, tasks may necessitate different degrees of precision with respect to contact points and applied forces/torques. We designate the restriction of the robot's learning capability to improving old skills and to manipulating rigid objects in static scenes in order to achieve short-horizon task objectives with low precision requirements for contact points and forces/torques as Level 0 (L0), the current research has a high probability of achieving L0. However, safety and accuracy remain paramount concerns.

\subsubsection{The design logic of the framework in this survey.}
Based on the general manipulation definition and robot learning development history, this paper proposes a framework for a general manipulation capability. Given that the scenarios are static, the framework is designed in a modular, sequential manner. To facilitate module migration, it is preferable for each module to be plug-and-play. Given the current reliance on human-in-the-loop mechanisms in autonomous driving and medical robotics to ensure safety, this framework aims for human-robot interaction through corrective instruction to ensure the safety of manipulation actions. The corrective action can be collected into the dataset and then improve old skills through offline training.

\subsubsection{The proposed framework limitations.}
(1) The framework is designed with a sequential structure, which contrasts with the parallel execution in human operation. (2) Both the proposed framework and the surveyed literature are based on learning-based approaches. While model-based methods may not generalize as well, they tend to significantly outperform learning-based methods in terms of success rates, precision and safety for specific tasks (\cite{pang2023global}). Therefore, investigating the integration of learning-based and model-based approaches remains an important research. (3) The framework proposed in this paper is based on the development of learning-based methods and the definition of general manipulation. The framework of brain-like cognitive research should also be explored.

\subsubsection{Product implementation strategy.}
During robot execution, continuous human supervision is not always feasible. Hence, integrating real-time monitoring through parallel surveillance videos during robot execution could enhance safety. The framework in this paper does not explicitly denote this parallel safety monitoring module, as it resembles the post-conditions detection module. The post-conditions detection module analyzes the robot's execution video to identify reasons for task failure, facilitating post-hoc correction to ensure task success. If the algorithm's task execution safety is 80\%, and the monitoring module predicts safety at 80\% as well, the probability of risky movements reduces to 4\%. Of course, for household robots, ensuring an over 99\% safety rate is imperative.
Initially, cloud-based monitoring of multiple robots by a single operator, with human intervention to correct erroneous behaviors, appears to be the best approach. This strategy not only reduces labor requirements but also ensures safety. Later, by gathering extensive data to improve model accuracy.

\subsection{What kind of learning capability should a general manipulation framework possess?} 
\subsubsection{The importance of learning ability.}
As an intelligent robot for general manipulation, it is inevitable that one cannot learn all the skills of an open-world during offline development, hence possessing a certain learning capability is necessary (\cite{wang2024imperative}). Within the framework of this paper, a module of corrective instruction is introduced, enabling the robot to rectify its actions. These corrective demonstrations are incorporated into the manipulation dataset and used to improve the policy offline through fine-tuning. However, this approach still focuses on learning old task skills and cannot acquire new ones.
\subsubsection{Definition of learning ability.}
The model of Policy should possess the capability of interactive, few-shot, continue, online learning to acquire a new skill and reinforce the policy's mastery of the newly learned skill through corrective instruction offline. Interactive refers to the ability to learn through human demonstration or by observing instructional videos. Learning through demonstration often requires physical control or teleoperation, which is less natural. Learning through observation of instructional videos aligns better with human learning patterns. However, when humans learn from teachers, they often do not predict the teacher's trajectory but rather understand the high-level description of the actions, akin to VLaMP (\cite{patel2023pretrained}). Few-shot continue learning enables the robot to learn new skills with minimal demonstrations without forgetting previously learned skills. Online learning entails processing observed data instantly and enabling the model to learn as quickly as possible.

\subsection{What foundation models bring for general manipulation?}

The emergence of foundation models can promote the progress of general manipulation. Meanwhile, for each section, we summarize the contributions of foundation models. As for Interaction, compared to traditional methods that use fixed questioning templates to eliminate instruction ambiguity, foundation models can provide the following for ambiguous instructions and corrective instructions: 1) more natural language communication, 2) multimodal perception to detect more types of ambiguity, and 3) powerful prior knowledge to better understand user intent. As for Object Affordance and Object Recognition in Pre- and Post-conditions Detection, foundation models bring several advantages. 1) They provide the perception capabilities for open-set affordance, detection, and segmentation, enabling zero-shot recognition of novel cases. 2) The powerful prior knowledge of foundation models accelerates the learning of tool affordance. 3) Foundation models assist methods in better understanding affordance and selecting more accurate poses.

As for the hierarchy of skills: 1) Foundation models can assist in processing and interpreting natural language inputs. 2) The acquisition of world knowledge and commonsense reasoning by foundation models enhances their perception and reasoning abilities. This has the potential to improve the scalability and generalizability of tasks within the skill hierarchy. As for 3D Reconstruction and 6D Pose Estimation in State. 1) Foundation models assist in reconstructing scenes with semantic information. 2) Foundation models' powerful 2D feature extraction ability can be applied to 3D lifting, aiding in the extraction of 3D features. 3) Foundation models enable open-set pose estimation.

As for policy. 1) Foundation models can help the model follow instructions better. 2) Foundation models can enhance the model's generalization ability and assist reinforcement learning. 3) Foundation models trained on large manipulation data can transfer prior knowledge to new task, such as PI0 transferring the mistake correction ability of pre-trained datasets to new task. As for manipulation data generation, the main contributions of foundation models are in simulation data and data augmentation. 1) Foundation models can generate 3D mesh assets in a zero-shot manner. 2) Foundation models help create diverse simulation scene layouts. 3) The vast priors of foundation models can be applied to data augmentation.

\subsection{How to use internet-scale video data for RFMs?}
As for what information from video dataset can be used, there are six main types of information to convert from video datasets: (1) Pose, such as capturing human hand poses and retargeting them to dexterous hand poses (\cite{shaw2023videodex, qin2022dexmv}). (2) Affordance, including grasp locations on objects and post-grasp waypoints (\cite{mendonca2023structured}). (3) Motion information, explicitly includes keypoints trajectories of objects and human hand during actions (\cite{xiong2021learning, yuan2024general, wen2023any}) and implicitly includes using VQ-VAE (\cite{van2017neural}) to generate a codebook for latent delta action (\cite{ye2024latent}). (4) Environment transition dynamic information, such as predicting hindsight images after completing the current action (\cite{wu2023unleashing, cheang2024gr, yang2023learning}). (5) Semantic information, such as descriptions of current task steps (\cite{wang2024vlm}) and task instruction (\cite{jain2024vid2robot}). (6) Spatial and texture information, such as MVP (\cite{radosavovic2023real}) suggests using masked autoencoding (\cite{he2022masked}) for improving visual reconstruction.

As for how to extract these useful information, various off-the-shelf models can be used to annotate the current video with additional labels, such as pose (\cite{shaw2023videodex}), affordance (\cite{mendonca2023structured}), key points trajectory (\cite{wen2023any}), latent action (\cite{ye2024latent}), mask and bounding boxes (\cite{shan2020understanding}). When adding various labels to the video dataset, different training objectives can be used to extract features from the video dataset, such as MAE (\cite{radosavovic2023real}), contrastive learning (\cite{ma2022vip}), time contrastive learning (\cite{ma2023liv}), temporal-difference learning (\cite{bhateja2023robotic}), video prediction objective(\cite{du2024learning}), affordance prediction objective(\cite{mendonca2023structured}), video-language alignment objective (\cite{nair2022r3m}), action motion objective(\cite{yuan2024general}) or combinations of these objectives (\cite{karamcheti2023language, zhou2021ibot}).

As for how to utilize the extracted information to enhance or train robotic foundation models, the current robotic foundation models primarily use two learning methods: imitation learning and reinforcement learning. Therefore, the discussion on the third issue focuses on leveraging prior knowledge from video datasets in these two methods. As for imitation learning, when the robotic foundation model outputs pose and the video dataset annoated with pose label, the video dataset can be directly used as training data for the robotic foundation model (\cite{shaw2023videodex,qin2022dexmv, kareer2024egomimic}). When leveraging affordance information, motion information, environment transition dynamics information, semantic information, spatial and texture information, it is essential to employ GMM \& CEM (\cite{mendonca2023structured}), Inverse Dynamic Model (IDM) (\cite{du2024learning, ye2024latent, wen2023any}), and Decoder (\cite{wang2023mimicplay,xiao2022masked, cheang2024gr, wu2023unleashing}) to transform these information into actions. Compared to other types of information, using semantic information treats the video as task instruction rather than observation (\cite{jain2024vid2robot, shah2023mutex, jang2022bc}). At the same time, semantic information can also be used to organize tasks into a hierarchy of skills (\cite{wang2024vlm}).

As for reinforcement learning, the environment transition dynamics can be used as a transition model (\cite{yang2023learning}). The encoder, trained on a video dataset with various objectives, can measure the distance between cross-embodiment actions, which then serves as the reward function or value function (\cite{bhateja2023robotic}). For example, \cite{guzey2024bridging} and \cite{xiong2021learning} use key points motion information to construct the encoder, which serves as the reward function for reinforcement learning. Since distribution shifts exist between cross-embodiment actions, AVID (\cite{smith2019avid}) and LbW (\cite{xiong2021learning}) translate human action images from videos into robot action images. However, this translation is limited to the pixel level. 

Current research focuses on different types of information in video datasets. The methods for extracting and using this information vary. It is important to consider which information from video datasets should be robustly applied to robotic foundation models. Video is similar to how humans perceive the world. Humans can improve their skills by watching experts. Similarly, using video datasets to construct a reinforcement learning from human feedback (RLHF) system in robotic foundation models is worth exploring (\cite{luo2024serl}).

\subsection{How to uses foundation models for post-conditions detection and post-hoc correction?}

The current data collection mostly focuses on gathering successful task execution data, ignoring the collection of data related to failed task executions. However, if data on failed task executions are collected and annotated with corresponding error reasons, it would be possible to train a model to both determine task execution success and analyze the reasons for task execution failure. AHA (\cite{duan2024aha}) trains a VLM to assess failures and output the reasons for these failures. However, the categories of failure modes are still limited, and it cannot output more open-ended failures, such as collaboration errors in dual-arm tasks. Many current studies use internet video data to improve the generalization of policies. Exploring the use of internet video data to enhance post-condition detection and employing multimodal perception to more accurately identify the reasons for failures is a promising direction. Post-hoc correction could then generate corrective action sequences based on the reasons for task execution failure and the task objectives, which would be handed over to a policy to generate corresponding corrective actions.

\subsection{How to use foundation models for End-effector Design?}
Currently, there are two primary approaches to designing end-effector. The first approach customizes the end-effector for specific tasks. The second approach makes the multi-fingered end-effector resemble a human hand. The end-effector designed with the first approach is usually easier to control because it has fewer degrees of freedom compared to the end-effector designed with the second approach. In \cite{billard2019trends}, dexterity is divided into two types: extrinsic dexterity and intrinsic dexterity. Extrinsic dexterity involves using external support, such as friction, gravity, and contact surfaces, to compensate for the lack of degrees of freedom. Intrinsic dexterity refers to the hand's ability to manipulate objects using its own degrees of freedom. Therefore, the first approach still has certain limitations for general manipulation.

The first approach requires manual design, extensive testing, and continual adjustments. In \cite{stella2023can}, LLMs are used for designing end-effector. However, this area is still in its early exploration stages. Using LLMs for end-effector design generates text descriptions, which still need to be manually translated into designs. This process is not fully automated. If we could develop modules for rotational and translational joints, and use something like protein structure prediction networks (\cite{jumper2021highly}), training a foundation model to output graph including these joints could help reduce the challenges of manual design. As for the second approach, the human hand has many sensors and actuators. This makes it nearly impossible to design a robotic hand that closely resembles the human hand. Therefore, it's essential to design the sensors and actuators carefully.

\begin{table*}[htbp]
\caption{\textbf{Representative Benchmarks.}}
\begin{threeparttable}
\resizebox{\textwidth}{!}{
\centering
\begin{tabular}{c c c c c c c c}
\hline
\Large \textbf{Benchmark} & \Large \textbf{Assert Categories} & \Large \textbf{Assert Number} & \Large \textbf{Room Layout Number} & \Large \textbf{Task Number} & \Large \textbf{Long-horizon Task} & \Large \textbf{Demonstration Instances} & \Large \textbf{Simulator}
 \\ \hline

\Large RLBench (\cite{james2020rlbench}) & \Large Rigid/Articulated/Deformable & \Large 28
& \Large Table-top & \Large 100 & \checkmark & \Large 90 & \Large CoppeliaSim \\

\rowcolor{mygray}
\Large Behavior-1K(\cite{li2023behavior}) & \Large Rigid/Articulated/Deformable & \Large 9318
& \Large 50 & \Large 1000 & \checkmark & - & \Large Omniverse\\

\Large VirtualHome (\cite{puig2018virtualhome}) & \Large Rigid/Articulated/Deformable & \Large 1138
& \Large 50 & \Large 8014 & \checkmark & \Large 5193 & \Large Unity \\

\rowcolor{mygray}
\Large RoboCasa (\cite{nasiriany2024robocasa}) & \Large Rigid/Articulated & \Large 2509
& \Large 10 & \Large 100 & $\times$ & \Large 100K+ & \Large Mujoco \\

\Large LIBERO-1K(\cite{liu2024libero}) & \Large Rigid/Articulated & \Large 67 & \Large Table-top & \Large 130 & \checkmark & \Large 6500 & \Large Mujoco\\

\rowcolor{mygray}
\Large Robosuite (\cite{zhu2020robosuite}) & \Large Rigid/Articulated & \Large 20
& \Large Table-top &  \Large 9 & $\times$ & - & \Large Mujoco \\

\Large Sapien(\cite{xiang2020sapien}) & \Large Rigid/Articulated & \Large 2346 & \Large Table-top & \Large 5 & $\times$  & - & \Large Sapien(NVIDIA PhysX+OpenGL)\\

\rowcolor{mygray}
\Large Maniskill2 (\cite{gu2023maniskill2}) & \Large Rigid/Articulated/Deformable & \Large 2144
& - & \Large 20 & $\times$ & \Large 30K+ & \Large Sapien \\

\Large CALVIN(\cite{mees2022calvin}) & \Large Rigid/Articulated & \Large 28 & \Large 4 & \Large 34 & \checkmark & \Large 40M & \Large Pybullet\\

\end{tabular}}
\end{threeparttable}
\label{tab:bechmark}
\end{table*}

\subsection{How to use foundation models for dexterous manipulation?}
One major challenge in data collection for dexterous manipulation lies in gathering data from multi-fingered end-effectors. Although model-based hard code method (\cite{zhu2024should}) can collect data on dexterous manipulation, they still require data analysis such as mutual information (\cite{hejna2025robot}) and entropy (\cite{zhu2024should}) to assess the quality of the data. Additionally, for multi-scenario and multi-task data collection, teleoperation methods are less dependent on algorithm performance compared to model-based hard code methods. However online teleoperation requires a real-robot system, which is not portable and cannot achieve in-the-wild data collection. Therefore, current mainstream research focuses on directly tracking human hand motions during manipulation without controlling the real-robot (\cite{wang2024dexcap}).

Two main learning-based methods for dexterous manipulation are imitation learning (\cite{ze2024h}) and reinforcement learning (\cite{ma2023eureka}).  Imitation learning can use a visual encoder (in Sec. \ref{sec:state}) for visuo-motor control. Diffusion policy (\cite{chi2023diffusion}) adapts the concept of diffusion to visuo-motor control. It addresses challenges in visuo-motor control such as action multimodality, sequential correlation to accommodate high-dimensional action sequences. It can also use an existing RFMs for fine-tuning (in Sec. \ref{sec:policy}). Fine-tuning with a RFMs allows a skill to work in an open world. This often performs better on unseen objects compared to visuo-motor control (\cite{rt2}).

Reinforcement learning offers exploration capability, which address suboptimal issues. This advantage distinguishes it from imitation learning. However, reinforcement learning is primarily trained in simulation. It still has limitations in addressing the sim-to-real challenge of complex tasks, such as pen-spinning. In Sec. \ref{sec:policy}, the use of foundation models to assist reinforcement learning is introduced.  FAC (\cite{ye2023foundation}) offers potential for training reinforcement learning in real-world environment, but it still lacks consideration of environment resets (\cite{gupta2021reset}) and safety. Therefore, using foundation models to assist reinforcement learning in real-world training requires further exploration.

Current learning methods each have their strengths and weaknesses (\cite{zhang2024hirt}). Therefore, learning approaches for dexterous manipulation should integrate different methods. For example, diffusion policy can assist reinforcement learning in addressing high-dimensional action spaces issue, while reinforcement learning can help diffusion policy overcome issues with suboptimal and negative data. Additionally, the learning models should consider both inputs and outputs. The factors necessary for achieving dexterous manipulation are summarized in the \ref{sec: appen_d}.

\subsection{How to use foundation models for whole-body control?}
The above discussion primarily focuses on the contact between the end-effector and the object. However, whole-body control is still needed in dexterous manipulation. For example, in a polishing robot, force-position hybrid control of the robotic arm is often required to manage the trajectory of contact points and  forces/torques. Mobile manipulation is essential for dexterous manipulation reachability. This idea is inspired by how humans handle objects. For example, when playing badminton, people use their waists, shoulders, elbows, and wrists together to hit the shuttlecock further. This aspect is often overlooked by current foundational models for manipulation. Although LEO (\cite{huang2023embodied}) can provide poses for both navigation and manipulation, it still does not address the synchronization issue between the two.

For whole-body control, the focus is on low-level control issues. A straightforward idea is to expand the action space of the policy model to include all joints of the robot. However, as the output dimensions increase, end-to-end training methods are more likely to diverge. Therefore, most current models output cartesian space poses and force/torques. These outputs are then optimized and converted into position or torque for each joint through a post-processing module (\cite{haviland2021neo}). To address end-to-end whole-body control issues, principal research is needed to facilitate network training and deployment.

\subsection{How to establish a benchmark?}
Current research on foundation models for manipulation focuses on various tasks, including interaction, hierarchical tasks, perception, detecting pre- and post-conditions, policy, and manipulation data generation. Therefore, a benchmark for foundation models for manipulation should include a comprehensive framework with diverse tasks. This framework should test individual tasks and tasks that involve connecting different modules. Since different simulators have unique physics engines and renderers, the benchmark should include standardized simulators and datasets.

Tab. \ref{tab:comparasion_1} lists the benchmarks used in current RFMs and we list some representative benchmarks in Tab. \ref{tab:bechmark}, highlighting a lack of standardization. This inconsistency hinders the development of RFMs for three main reasons. Firstly, current RFMs are tied to the specific parameters of each robot, such as the choice of sensors, camera pose, and the robot's degrees of freedom. These factors prevent RFMs from being easily transferred across different robots. Secondly, testing the generalization and success rate of general manipulation capability requires a wide range of scenes and tasks. Thirdly, there is no standardized metric for assessing general manipulation capability.

As for the RFMs are not transferable between different robots. The issue arises from focusing solely on testing RFM algorithms without considering hardware, which is an ineffective approach. General manipulation requires whole-body control. Thus, evaluating the generalization and success rate of RFMs should involve both algorithms and hardware, unlike in computer vision where only algorithms are considered. To address this, the simulation benchmark should include an easy interface for importing various robot hardware configurations.

As for the requirment of a wide range of scenes and tasks. Although iGibson (\cite{li2021igibson}) and BEHAVIOR-1K (\cite{li2023behavior}) support simulating a variety of household tasks with high realism, they are still manually constructed. In Sec. \ref{sec:generation}, we discuss how foundation models can automate the generation of scenes and tasks. Using foundation models to create numerous scenes, combined with VLMs for accuracy checking and minimal human intervention, could be a valuable approach to explore.

As for the metric for assessing general manipulation. The current evaluation standards mainly focus on success rates. However, in real-world applications, other metrics should also be considered. For instance, the system's real-time performance is important. Most algorithms focus on building the generalization of skills. They often overlook the amount of data and speed required for RFMs to learn a new skill. Therefore, evaluation metric should also include the learning ability of RFMs.

Overall, to assess the ability for general manipulation, methods used for testing medical robots can be referenced. Start with extensive testing in simulation environments, followed by limited tests in real-world settings. Continue evaluating the general manipulation capability during the product's application phase.

\section{Conclusion}\label{sec:conclusion}
The impressive performance of foundation models in the fields of computer vision and natural language suggests the potential of embedding foundation models into manipulation tasks as a viable path toward achieving general manipulation capability. However, current research lacks consideration of a general manipulation framework. Thus, this paper proposes a general manipulation framework based on the development of robot learning for manipulation and the definition of general manipulation. It also describes the opportunities that foundation models bring to each module of the framework.

We designate the restriction of the robot’s learning capability to improving old skills and to manipulating rigid objects in static scenes in order to achieve short-horizon task objectives with low precision requirements for contact points and forces/torques as Level 0 (L0), the current research  has a high probability of achieving L0.

Then, we discuss the following points: (1) the logic and implementation strategies of the designed framework, (2) the learning capability required for general manipulation, (3) what foundation models bring for general manipulation, (4) how to use internet-scale video data for RFMs, (5) how to uses foundation models for post-conditions detection and post-hoc correction, (6) how to use foundation models for end-effector design, (7) how to use foundation models for dexterous manipulation,  (8) how to use foundation models for whole-body control, and (9) how to establish a benchmark.

Additionally, the proposed framework has certain limitations: (1) The framework is designed with a sequential structure, which contrasts with the parallel execution in human operation. (2) Both the proposed framework and the surveyed literature are based on learning-based approaches. While model-based methods may not generalize as well, they tend to significantly outperform learning-based methods in terms of success rates, precision and safety for specific tasks (\cite{pang2023global}). Therefore, investigating the integration of learning-based and model-based approaches remains an important research. (3) The framework proposed in this paper is based on the development of learning-based methods and the definition of general manipulation. The framework of brain-like cognitive research should also be explored.

Finally, foundation models present opportunities for each module of the framework, but many challenges still remain:
\begin{enumerate}
    \item \textbf{Interaction} Human interaction involves not only language but also gestures and actions. Incorporating multimodal inputs into interaction modules can enhance recognition capability.
    \item \textbf{Hierarchical of skills} The hierarchy of skills still has many unconsidered factors, such as achieving tasks in the shortest time with the highest efficiency, and how to generate strategies for dynamic scenes.
    \item \textbf{Pre- and post-conditions detection} Current research on post-condition detection primarily focuses on detection after robot execution. However, this delay is unacceptable. Therefore, it is necessary to implement failure detection and analysis of failure reasons during the robot execution.
    \item \textbf{State} The representation of state requires integration of multiple modalities, such as touch and hearing. Additionally, it's important to consider the opportunities that foundation models can bring to active perception.
    \item \textbf{Policy} Current research on RFMs primarily involves fine-tuning VLMs. This approach deprives RFMs of the ability to self-explore. The extensive parameters of RFMs require significant computational resources for training and real-time reference, and model training also needs abundant data. Additionally, there is a lack of a unified benchmark for evaluating different RFMs.
    \item \textbf{Environment Transition Module} The foundation models inherently contain abundant physical priors. Applying foundation models to build a highly realistic physical model assist reinforcement learning training is a direction worth exploring.
    \item \textbf{Data Generation} The accuracy of data generated by LLMs and VGMs remains insufficient, necessitating appropriate check module and data cleaning algorithms.
\end{enumerate}

\begin{acks}
The authors thank the editors and anonymous reviewers for the constructive feedback and in-depth review of this work. This paper pays tribute to the researchers and engineers who tirelessly advance the field of robotics — let’s change the world.
\end{acks}

\begin{dci}
The author(s) declared no potential conflicts of interest with respect to the research, authorship, and/or publication of this article.
\end{dci}

\begin{funding}
The author(s) disclosed receipt of the following financial support for the research, authorship, and/or publication of this article: This work was supported by the National Natural Science Foundation of China under Grants 62536001 and 62173197, and by the National Natural Science Foundation of China for Key International Collaboration under Grant 62120106005.
\end{funding}

\renewcommand{\thesection}{Appendix.A}
\titleformat{\section}[block]{\normalfont\bfseries\large}{\thesection}{1em}{}
\section{Analysis of Policy Work} \label{sec: appen_a}

As shown in Tab. \ref{tab:comparasion_1} and Tab. \ref{tab:comparasion_2}. The types of datasets currently used can be categorized into internet image-language pairs data, human video data, and robot demonstration data in real-world and simulation environments. RT-2 (\cite{rt2}) demonstrates that co-training with internet image-language pairs improves the model's generalization. GR-1 (\cite{wu2023unleashing}), GR-2 (\cite{cheang2024gr}), and LAPA (\cite{ye2024latent}) show that training with human video data and robot demonstrations also enhances task generalization. GR-1 and GR-2 extract priors from human video data via image prediction, while LAPA uses latent-action for prior extraction. The advantages and disadvantages between these two methods are not clearly evident.

The Input Modality shows that VLAKP primarily uses language and 3D representations as inputs, while VLADP relies on language, images, and proprioception. At the same time, adding a mask from the open-set visual module to the input improves the model's generalization (\cite{stone2023open}). In terms of model architecture, VLAKP tends to use a 3D Feature Lift model combined with a Diffusion Model, while VLADP adopts a dual-system approach (slow and fast systems). This allows VLADP to effectively utilize the prior knowledge from VLM and enhance inference frequency.

The current training objectives include next-token prediction, regression, diffusion, and TD learning in reinforcement learning. No research has conclusively shown which method is best. TD Learning in reinforcement learning can help robotic systems become more proficient than human teleoperators, exploiting the full potential of the hardware to perform tasks quickly, fluently, and reliably. It also enables robotic systems to improve autonomously through gathered experience, instead of relying solely on high-quality demonstrations. The Diffusion Objective can address action multi-modality issues and is widely applied in dexterous manipulation and high-dimensional action space tasks. Diffusion can be divided into Diffusion Policy-based and Flow Matching Policy-based approaches. Experiments in IMLE (\cite{rana2025imle}) show that Diffusion Policy performs better than Flow Matching Policy. Fast (\cite{pertsch2025fast}) introduces a compression-based tokenization scheme that matches the performance of diffusion VLAs. OpenVLA-OFT (\cite{kim2025fine}) shows that the regression train-objective outperforms next-token prediction and diffusion. Therefore, different data, architectures, and tasks require different training objectives. The specific situations corresponding to each train objective should be deeply researched.

From the Output Modality, it can be seen that outputting the hindsight goal image along with the action improves model stability (\cite{bousmalis2023robocat, cheang2024gr}). The Benchmark shows that current benchmarks vary across methods. A unified benchmark standard can promote progress in the field. Regarding Success Rate, there are few experiments testing adaptation ability. Adaptation ability refers not only to quickly adapting to new tasks but also to quickly adapting to new embodiment configurations. Cross-embodiment learning can be divided into unified action space ({\cite{liu2024rdt, black2410pi0}}) and changing the action head ({\cite{team2024octo, wang2024scaling}}). Currently, the comparison between these two methods for cross-embodiment is not very clear. Regarding the manipulation task, it starts with single-arm pick-and-place tasks and evolves into dexterous manipulation using humanoid robot upper bodies. However, a considerable gap remains to achieve general dexterous manipulation. From the Failure Mode, it is observed that "imprecise pose error" and "wrong object error" occur most frequently.

\renewcommand{\thesection}{Appendix.B}
\section{Comparative analysis of 2D and 3D-based methods.}

As shown in Tab. \ref{tab:comparasion_2}, current methods utilize 2D observation or 3D observation. It is worth investigating which modality is more suitable for manipulation tasks.

3D observation can be expressed in various forms. These include RGBD images, point clouds, voxels or multi-view images with camera extrinsic parameters (\cite{ze20243d}). These 3D forms has not achieved large-scale adoption on the internet. As a result, the volume of 3D data remains significantly smaller than that of 2D images (\cite{chen2024spatialvlm}). Although SpatialVLM (\cite{chen2024spatialvlm}) uses off-the-shelf models to convert 2D images into 3D forms, the quality remains uncertain. Generating large-scale 3D observations in simulation environments might be an effective approach. However, there is still a sim-to-real gap.

As for 2D and 3D representation learning, we have already introduced many pre-trained encoders for 2D representation in Sec. \ref{fig:pre_conditions_detection}. For 3D representation encoders, the main options currently include PointNet++ (\cite{qi2017pointnet++}) and PointNext (\cite{qian2022pointnext}). These encoders extract key features from point clouds. DP3 (\cite{ze20243d}) introduces a holistic 1D embedding pooled from the 3D scene point cloud, which outperforms PointNet++ and PointNext. However, 3D Diffuser Actor (\cite{ke20243d}) shows that generating 3D representations by lifting features from perspective views to a 3D robot workspace, based on sensed depth and camera extrinsics, achieves even better results than DP3 (\cite{ze20243d}). Additionally, converting 3D data into 2D allows the use of pre-trained 2D encoders that trained on large-scale datasets for feature extraction. Then, lifting 2D feature to 3D space. This method extracts texture features, spatial features and semantic features and become a notable trend in 3D representation.

The current lift techniques can be categorized into three types: Direct Reconstruction, Feature Fusion, and Neural Field (\cite{hong20233d}). Direct Reconstruction refers to features are mapped directly to the 3D space using camera extrinsics. However, this method is sensitive to noise in the camera pose. Feature Fusion combines 2D features into 3D maps using gradslam (\cite{murthy2019gradslam}). This approach is more robust to camera pose noise. However, it requires depth map rendering from 3D data. Neural Field constructs 3D compact representation using a neural voxel field (\cite{sun2022direct}). This method is more robust to noise in camera pose and does not require depth map renderings from 3D data.

Current manipulation tasks are mainly divided into high-level and low-level. High-level tasks involve decision-making, such as the hierarchy of skills. Low-level tasks focus on execution, like policies. For high-level manipulation task, 3D observation has stronger spatial reasoning capabilities compared to 2D observation (\cite{chen2024spatialvlm}). It can recognize quantitative relationships of physical objects, such as distances or size differences. For low-level manipulation task, ChainedDiffuser (\cite{xian2023chaineddiffuser}) demonstrates that 3D methods are more stable than 2D methods under varying camera viewpoints. DP3 (\cite{ze20243d}) has shown that diffusion policies with 3D input achieve higher success rates compared to 2D image and the point cloud format performs best. However, the comparison is based on a relatively small dataset. \cite{lin2024data} introduces a scaling law for object diversity and environment diversity. However, it is still unclear whether 2D or 3D observation is more suitable for the scaling law.

\renewcommand{\thesection}{Appendix.C}
\section{Analysis of Hierarchy of Skills} \label{sec: appen_5}
Regarding Tab. \ref{tab:comparise_5}, it can be seen from the Foundation Models that current methods based on Video Instruction and Language Instruction have shifted from using the previous SOTA LLM to a stronger focus on utilizing the SOTA VLM. In terms of Manipulation Tasks and Horizon Steps, most current methods design tasks with a maximum length of about 10 steps. The comparison of Success Rate between Method and Baseline shows clear differences. Considering a robot’s abilities based on its embodiment, safety measures, action execution feedback, and enhanced scene grounding capability improves its task planning performance. The Failure Modes reveal that the main issues are Wrong Object Error, Plan Error, and Spatial Relations Error. This indicates that the key challenges in current task planning lie in perception and reasoning.

\renewcommand{\thesection}{Appendix.D}
\section{Model-Based Methods for Dexterous Manipulation} \label{sec: appen_d}

\cite{bicchi2000hands} offers a thorough and widely accepted definition: dexterous manipulation is the capability of changing the position and orientation of the manipulated object from a given reference configuration to a different one, arbitrarily chosen within the hand workspace. Based on this definition, the dexterous manipulation can be described as: based on the designed end-effector, determining a sequence of contact points and the forces/torques to be exerted on the object, and control the whole-body to accomplish a specific task.

Based on this definition, the challenges of dexterous manipulation lie in the design of the end-effector, determining a sequence of contact points and forces/torques, and whole-body control. The process of determining a sequence of contact points and forces/torques can be divided into model-based approach and the learning-based approach. The model-based approach is interpretable, explicitly showing the factors to consider in dexterous manipulation. Therefore, this section explains model-based approach.

When solving simple tasks, contact points can remain fixed. However, for complex tasks, contact points need to change. Thus, a sequence of contact points and forces/torques is required, achieved through regrasping or finger gaiting. The sequence of contact points and forces/torques are positively correlated with the trajectory of the motion and wrench of the manipulated object. When the wrench and motion of the manipulated object at a given moment are determined, they can be mapped to the corresponding contact points and forces/torques between the end-effector and the manipulated object. Therefore, we will explain the pipeline of this mapping relationship for a specific moment.

\begin{figure}[t]
\centering
\includegraphics[width=\linewidth]{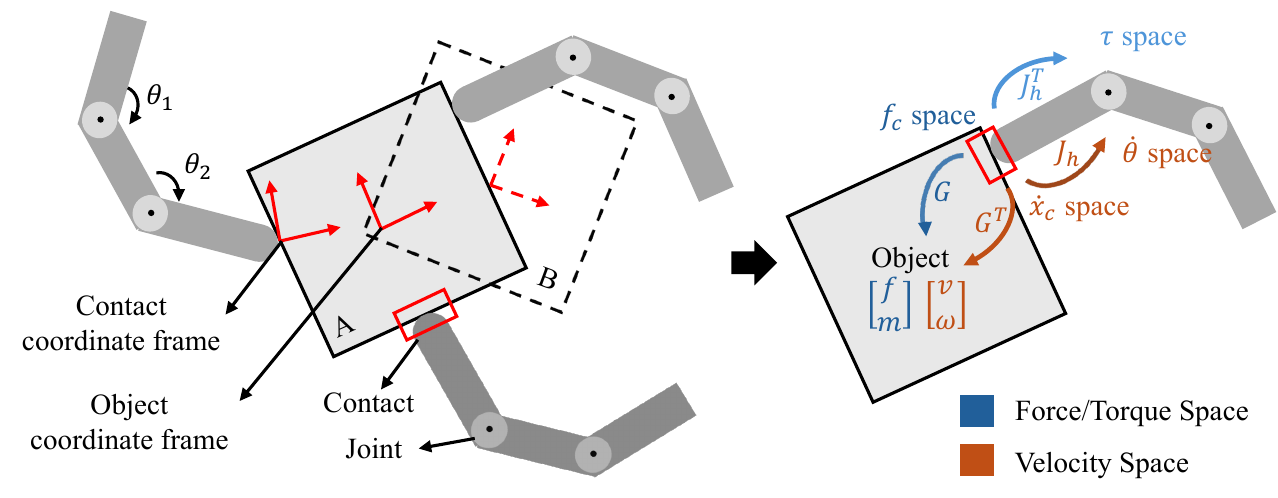}
\caption{Model-based Generation of a Sequence of Contact Points and Forces/Torques.} 
\label{fig:dexterity_framewok} 
\end{figure}

As shown in Figure \ref{fig:dexterity_framewok}, we use a simple process to explain the generation of a sequence of contact points and forces/torques. Assuming the motion trajectory of the object is already obtained, the target wrench can be calculated based on the object's mass and inertia. Contact points can be optimized based on object geometry, object material, and end-effector geometry etc, using appropriate metrics (\cite{ferrari1992planning}). Alternatively, they can be generated using a knowledge-based approach (\cite{stansfield1991robotic}). Subsequently, a coordinate system is established at the centroid of the object. Based on the location of the contact points, the target wrench is converted into fingertip force. Finally, the force of the fingertip is converted into the forces/torques required by the actuators through hand jacobian (\cite{okamura2000overview}).

The pipeline contact points and forces/torques mentioned above are derived sequentially. In practical applications, both can be optimized simultaneously (\cite{xu2024manifoundation}). However, due to the discontinuities and mode changes inherent in contact dynamics, it creates challenges for planning in contact-rich manipulation (\cite{pang2023planning}).

\begin{figure}[t]
\centering
\includegraphics[width=\linewidth]{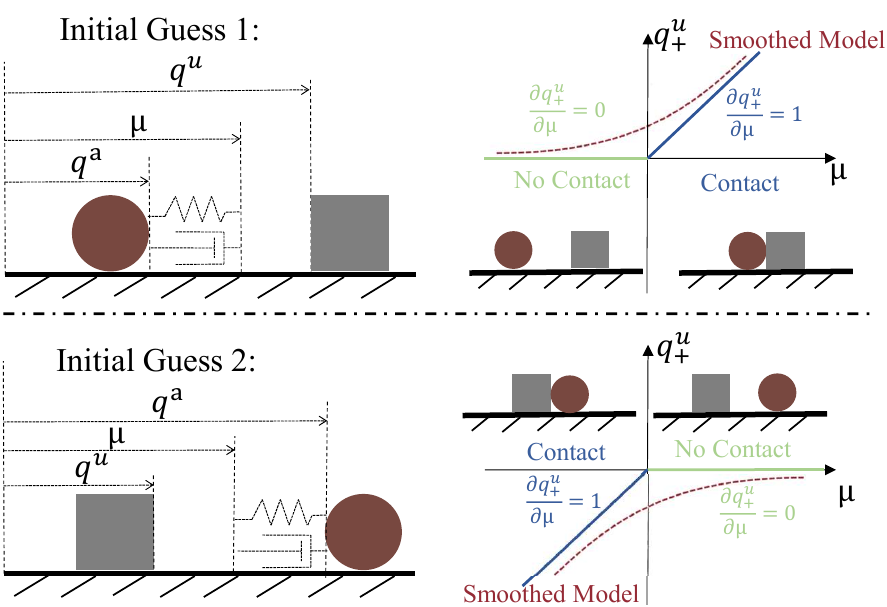}
\caption{Non-smooth dynamic system.} 
\label{fig:non-smooth} 
\end{figure}

As shown in Fig. \ref{fig:non-smooth}, $q^{a}$ and $\mu$ represent the actual and commanded positions of the actuated ball, while $q^{u}$ represents an unactuated box that slides along a rail with sufficient damping. Our goal is to push the box by the ball to transition from the current configuration $q^{u}$ to the goal configuration $q^{u}_{goal}$. Suppose $q^{u}_{+}$ is the steady-state position of the box after the ball is commanded to position $\mu$. We expect that $q^{u}_{+}$ and $q^{u}_{goal}$ are sufficiently close. Typically,  $q^{u}_{+}=f(q^{u}, q^{a}, \mu)$ represents the system's dynamic, which can be obtained through system identification (\cite{suhdoes}).

As shown in Fig. \ref{fig:non-smooth}, the gradient is non-smooth when the ball is either in contact with or separated from the box. This presents challenges for optimization-based problem-solving methods. 
Current approaches for smoothing contact dynamics are mainly divided into analytic smoothing and randomized smoothing (\cite{pang2023global}). However, it is often necessary to consider frictional contact constraints, which can make the optimization problem non-convex (\cite{tedrake5underactuated, le2024contact}). To address this issue, current methods primarily rely on Anitescu's convex relaxation(\cite{anitescu2006optimization, pang2023global, jin2024complementarity}) and Todorov's model (\cite{todorov2012mujoco, todorov2014convex}). At the same time, the contact dynamics also need to consider the robot's joint limits constraint, the penetration constraint between the hand and the object, and the self-penetration constraint (\cite{pang2021convex,xu2024manifoundation}). Although smoothed dynamics on trajectory optimization, such as iMPC (\cite{suh2022bundled} and LQR (\cite{shirai2024linear}) are considered effective, these methods are only suitable when the goal is close to the initial configuration. When the goal is farther from the initial configuration, a more global strategy is required.

As shown in Fig. \ref{fig:non-smooth}, there are two initial guesses: 1) the ball is on the left of the box, and 2) the ball is on the right of the box. Different initial guesses lead to different dynamic model. Choosing the wrong initial guess can prevent the optimization problem finding the optimal solution. To address this issue, the system dynamics need to be divided into contact modes, and a global strategy is required (\cite{pang2023global}). In Fig. \ref{fig:non-smooth}, the ball and the box experience both contact and no contact. However, for a system with $n$ rigid bodies, there can be up to $C^{2}_{n}$ contact pairs. Each contact pair may have multiple contact modes, such as sticking/rolling, separation, and sliding, which can lead to an explosion of contact modes. Although \cite{huang2021efficient} considers kinematically feasible contact modes and reduces the number to $O(n^{d^{2}/2+2.5d})$(where d is the number of degrees of freedom and n is the number of contact points), the reduced number of contact modes is still too large to handle.

To address the above issues, current model-based techniques are mainly divided into contact-explicit and contact-implicit (\cite{jiang2024contact}). The contact-explicit approach in dexterous in-hand manipulation involves explicitly representing contacts, such as their locations, modes, and reaction forces (\cite{cheng2023enhancing}). These representations form contact sequences, which are obtained through methods like enumerating (\cite{aceituno2020global, hogan2020feedback}), searching (\cite{chen2021trajectotree, zhu2023efficient, cruciani2018dexterous}), sampling (\cite{cheng2023enhancing}), or demonstration (\cite{khadivar2023adaptive}). However, these methods suffer from poor scalability (\cite{pang2023global}) and contacts easily fall into the local optimum (\cite{jiang2024contact}). The contact-implicit approach avoids the complexity of explicit contact representation. It uses methods like relaxed complementary constraints (\cite{kim2023contact, le2024fast}), smooth surrogate models (\cite{pang2023global, onol2019contact}), or direct control sampling (\cite{howell2022predictive}). However, the smoothing process introduces force-at-a-distance effects and sacrifices physical fidelity (\cite{pang2023global}). Although \cite{jiang2024contact} compensates for discrepancies between actual and planned contact modes using tactile feedback, this method assumes quasi-static contact model and struggles to handle highly dynamic actions, such as spinning a pen between fingers.

From the above analysis, it is evident that choosing appropriate contact points and forces/torques requires considering the object's geometry, mass, inertia, material, and friction parameters, as well as the end-effector's geometry, material, and actuator capability. At the same time, the hybrid process between non-smooth contact modeling and planning faces significant challenges for contact-rich tasks.

\newpage
\begin{table*}[t!]
\captionsetup{justification=justified,singlelinecheck=false}
\caption{\textbf{Representative Policy Works in Sec. \ref{sec:policy}.} The policy type is consistent with VLAC, VLAKP, VLADP in Sec. \ref{sec:policy}. Current robot foundation model training recipes include From-Scratch and Pre-Train + Fine-Tune (Post-Train). Some methods use vision or language foundation models during training. However, these foundation models are not pre-trained on large manipulation datasets. So these methods are still considered From-Scratch( \cite{shridhar2021cliport}). Pre-Train refers to training on extensive manipulation datasets. Fine-Tune involves post-training a pre-trained model on a small task-specific manipulation dataset for a particular task. The datasets used in various stages are detailed in the `Dataset’. `Benchmark’ refers to the benchmark used in the experiment. The `Success Rate’ is categorized into Seen and Unseen. `Seen' refers to test cases that have appeared in the training data, while `Unseen' refers to test cases that have not been present in the training data. However, some studies do not clearly differentiate between Seen and Unseen cases, and in our analysis, we also do not make this distinction. Meanwhile, for the pre-train + fine-tune training recipe, the success rate is further divided into the out-of-box generalization ability of the pre-trained model and the adaptation ability of the fine-tuned model. We have also provided annotations on the success rates of both the out-of-box and fine-tuned models. Due to differences in benchmarks, solely estimating the success rate is meaningless. Therefore, we also include the success rates of the baseline for comparison. `Embodiment Config' refers to the type of robot selected in the experiment. `Manipulation Task' refers to the manipulation task designed for the experiment. `Failure Mode' refers to the common failure cases of the method, as indicated in the paper.} 
\label{tab:comparasion_1}
\huge
\begin{threeparttable}
\resizebox{\textwidth}{!}{
\rowcolors{3}{mygray}{white} 
\begin{tabular}{c c c c c c c c c c c}
\hline
\textbf{Method} 
& \textbf{\begin{tabular}[c]{@{}c@{}}Policy  Type\end{tabular}} 
& \multicolumn{3}{c}{\textbf{Dataset}} 
& \textbf{Benchmark} 
& \multicolumn{2}{c}{\textbf{Success Rate}} 
& \textbf{Embodiment Config} 
& \textbf{Manipulation Task} 
& \textbf{Failure Mode} \\
& 
& \textbf{From-Scratch} 
& \textbf{Pre-Train} 
& \textbf{Fine-Tune} 
& 
& \textbf{Seen} 
& \textbf{UnSeen} 
& 
& 
& \\
\hline

\begin{tabular}[c]{@{}c@{}}Code as Policy (\cite{liang2023code}) \end{tabular} 
& VLAC 
& NA & NA & NA 
& CLIPort 
& \begin{tabular}[c]{@{}c@{}}Cap: 89.36\%\\ CLIPort: 53.23\%\end{tabular} 
& \begin{tabular}[c]{@{}c@{}}Cap: 71\%\\ CLIPort: 0\%\end{tabular} 
& \begin{tabular}[c]{@{}c@{}}Single-Arm\\2F Gripper\end{tabular} 
& \begin{tabular}[c]{@{}c@{}}Table-top Manipulation \\ (pick-place)\end{tabular} 
&  \begin{tabular}[c]{@{}c@{}}Code Generation Error ;\\ Imprecise Pose Error;\\ Wrong Object Error;\end{tabular}  \\

\begin{tabular}[c]{@{}c@{}}Instruct2Act (\cite{huang2023instruct2act})\end{tabular} 
& VLAKP 
& NA & NA & NA 
& VIMA 
& \multicolumn{2}{c}{\begin{tabular}[c]{@{}c@{}}Instruct2Act: 84.4\%\\ Gato: 40.8\%\\ Flamingo: 45.5\%\end{tabular}} 
& \begin{tabular}[c]{@{}c@{}}Single-Arm\\Varies tools\end{tabular} 
& Table-top Manipulation 
& Task Order Failure;\\

VoxPoser (\cite{huang2023voxposer})
& VLAKP 
& NA & NA & NA 
& Self-Creation 
& \begin{tabular}[c]{@{}c@{}}VoxPoser: 76.9\%\\ CaP: 46.5\%\end{tabular} 
& \begin{tabular}[c]{@{}c@{}}VoxPoser: 69.6\%\\ CaP: 39.6\%\end{tabular} 
& \begin{tabular}[c]{@{}c@{}}Single-Arm\\2F Gripper\end{tabular} 
& \begin{tabular}[c]{@{}c@{}}Table-top Manipulation\\ (sweep; push; turn on;\\ open; pick-place)\end{tabular} 
& \begin{tabular}[c]{@{}c@{}}Motion Planning Error;\\Imprecise Trajectory\end{tabular} \\

ReKep (\cite{huang2024rekep})
& VLAKP 
& NA & NA & NA 
& Self-Creation 
& \multicolumn{2}{c}{\begin{tabular}[c]{@{}c@{}}ReKep: 44.3\%\\ VoxPoser: 10.0\%\end{tabular}} 
& \begin{tabular}[c]{@{}c@{}}Single-Arm;\\Dual-Arm;\\2F Gripper\end{tabular} 
& \begin{tabular}[c]{@{}c@{}}Pour Tea;\\Recycle Can;\\Stow Book;\\Tape Box;\\Fold Garment;\\Pack Shoes;\\Collab. Folding\end{tabular} 
& \begin{tabular}[c]{@{}c@{}}Motion Planning Error;\\Incorrect Keypoints\end{tabular} \\

CLIPort (\cite{shridhar2021cliport})
& VLAKP 
& \begin{tabular}[c]{@{}c@{}}1000 Demo. per Task\\ (Total 10 Tasks)\end{tabular} 
& NA & NA 
& CLIPort 
& \begin{tabular}[c]{@{}c@{}}CLIPort: 87.7\%\end{tabular} 
& \begin{tabular}[c]{@{}c@{}}CLIPort: 57.1\%\end{tabular} 
& \begin{tabular}[c]{@{}c@{}}UR5e;\\Suction Gripper.\end{tabular} 
& \begin{tabular}[c]{@{}c@{}}Table-top Manipulation\\ (pick-up; sweep;\\ align rope; etc.)\end{tabular} 
& \begin{tabular}[c]{@{}c@{}}Wrong Object Error;\\Grasp Fail Error\end{tabular} \\

PerAct (\cite{shridhar2023perceiver})
& VLAKP 
& \begin{tabular}[c]{@{}c@{}}100 Demo. per Task\\ (Total 18 Tasks)\end{tabular} 
& NA & NA 
& RLBench 
& \multicolumn{2}{c}{PerAct: 42.7\%} 
& \begin{tabular}[c]{@{}c@{}}Franka Panda;\\2F Gripper\end{tabular} 
& \begin{tabular}[c]{@{}c@{}}Table-top Manipulation\\ (pick-place; open-close;\\ sweep; screw; insert;)\end{tabular} 
& Grasp Fail Error \\

Act3D (\cite{gervet2023act3d})
& VLAKP 
& \begin{tabular}[c]{@{}c@{}}100 Demo. per Task\\ (Total 18 Tasks)\end{tabular} 
& NA & NA 
& RLBench 
& \multicolumn{2}{c}{\begin{tabular}[c]{@{}c@{}}Act3D: 65\%\\ PerAct: 42.7\%\end{tabular}} 
& \begin{tabular}[c]{@{}c@{}}Franka Panda;\\2F Gripper\end{tabular} 
& \begin{tabular}[c]{@{}c@{}}Table-top Manipulation\\ (pick-place; open-close;\\ sweep; screw; insert;)\end{tabular} 
& \begin{tabular}[c]{@{}c@{}}Imprecise Pose Error\end{tabular} \\

ChainedDiffuser (\cite{xian2023chaineddiffuser})
& VLAKP 
& \begin{tabular}[c]{@{}c@{}}100 Demo. per Task\\(Total 10 Tasks)\end{tabular} 
& NA & NA 
& Self-Creation 
& \multicolumn{2}{c}{\begin{tabular}[c]{@{}c@{}}ChainedDiffuser: 80.9\%\\ Act3D: 21\%\end{tabular}} 
& \begin{tabular}[c]{@{}c@{}}Franka Panda;\\2F Gripper\end{tabular} 
& \begin{tabular}[c]{@{}c@{}}Continuous Interactions \\ Manipulation Task\\(unplug charger; wipe desk;\\ open-close; books on shelf)\end{tabular} 
& \begin{tabular}[c]{@{}c@{}}Imprecise Pose Error\end{tabular} \\

3D Diffuser Actor (\cite{ke20243d})
& VLAKP 
& \begin{tabular}{@{}c@{}}RLBench:\\ 100 Demo. per Task\\(Total 18 Tasks)\\CALVIN:\\ Six Hours per Env\\(Total 4 Envs)\end{tabular} 
& NA & NA 
& \begin{tabular}{@{}c@{}}RLBench;\\CALVIN\end{tabular} 
& \multicolumn{2}{c}{\begin{tabular}{@{}c@{}}RLBench(Multi-View): \\ 3D Diffuser Actor: 81.3\%\\Act3D: 63.2\%\\PerAct: 49.4\% \\ RLBench(Single-View):\\ 3D Diffuser Actor: 78.4\%\\Act3D: 65.3\%\\GNFactor: 31.7\% \\ CALVIN: \\ 3D Diffuser Actor: 66.96\% \\ GR-1: 61.2\% \\ SuSIE:53.8\% \\ RoboFlamingo: 49.5\% \\ RT-1: 18\% \\ ChainedDiffuser: 16.8\%\end{tabular}} 
& \begin{tabular}{@{}c@{}}Franka Panda;\\2F Gripper\end{tabular} 
& \begin{tabular}{@{}c@{}}Table-top Manipulation\\(pick-place; open-close; \\ sweep; screw; insert; \\ push; lift; turn on/off)\end{tabular} 
& \begin{tabular}{@{}c@{}}Intermediate\\Task Fail Error\end{tabular} \\

GNFactor (\cite{ze2023gnfactor})
& VLAKP 
& \begin{tabular}{@{}c@{}}20 Demo. per Task\\(Total 10 Tasks)\end{tabular} 
& NA & NA 
& RLBench 
& \begin{tabular}{@{}c@{}}GNFactor: 31.7\%\\PerAct: 22.7\%\end{tabular} 
& \begin{tabular}{@{}c@{}}GNFactor: 28.3\%\\PerAct: 18.0\%\end{tabular} 
& \begin{tabular}{@{}c@{}}Franka Panda;\\2F Gripper\end{tabular} 
& \begin{tabular}{@{}c@{}}Table-top Manipulation\\(pick-place; open-close; \\ sweep; screw; insert; push)\end{tabular} 
& NR \\

DNAct (\cite{yan2024dnact})
& VLAKP 
& \begin{tabular}{@{}c@{}}50 Demo. per Task\\(Total 10 Tasks)\end{tabular} 
& NA & NA 
& RLBench 
& \begin{tabular}{@{}c@{}}DNAct: 59.6\%\\GNFactor: 43.3\%\\PerAct: 35.6\%\end{tabular} 
& \begin{tabular}{@{}c@{}}DNAct: 52.3\%\\GNFactor: 30.9\%\\PerAct: 29.8\%\end{tabular} 
& \begin{tabular}{@{}c@{}}Franka Panda;\\2F Gripper\end{tabular} 
& \begin{tabular}{@{}c@{}}Table-top Manipulation\\(pick-place; open-close;\\ sweep; screw; insert; push)\end{tabular} 
& NR \\

VoxAct-B (\cite{liu2024voxact})
& VLAKP 
& \begin{tabular}{@{}c@{}}100 Demo. per Task\\(Total 4 Tasks)\end{tabular} 
& NA & NA 
& Self-Creation 
& \multicolumn{2}{c}{\begin{tabular}{@{}c@{}}VoxAct-B: 49\%\\VoxPoser: 11\%\\Act: 27.6\%\\Diffusion Policy: 8\%\end{tabular}} 
& \begin{tabular}{@{}c@{}}Dual-Arm;\\2F Gripper\end{tabular} 
& \begin{tabular}{@{}c@{}}Asymmetric Bimanual \\ Manipulation\\(open jar; open drawer; \\ put item in drawer;\\ handover)\end{tabular} 
& \begin{tabular}{@{}c@{}}Imprecise Pose Error;\\Collisions;\\Dual-Arm Collaboration Error;\\Motion Planning Error\end{tabular} \\

LEO (\cite{huang2023embodied})
& VLAKP 
& NA & \begin{tabular}{@{}c@{}}100K Demo. per Task\\(Total 3 Tasks)\end{tabular} & NA 
& CLIPort 
& \begin{tabular}{@{}c@{}}LEO: 87.2\%\\CLIPort: 91.1\%\end{tabular} 
& \begin{tabular}{@{}c@{}}LEO: 63.4\%\\CLIPort: 59.6\%\end{tabular} 
& \begin{tabular}{@{}c@{}}UR5e;\\Suction Gripper\end{tabular} 
& \begin{tabular}{@{}c@{}}Table-top Manipulation\\(pick-place)\end{tabular} 
& NR \\

ManiFoundation ((\cite{huang2023embodied})) 
& VLAKP 
& NA & \begin{tabular}{@{}c@{}}200K Rigid, Clothes,\\Deformation Data\end{tabular} & NA 
& Self-Creation 
& \multicolumn{2}{c}{ManiFoundation: 88.6\%} 
& \begin{tabular}{@{}c@{}}Varies Single-Arm;\\Varies 2F Gripper;\\Varies Dexterous Hand\end{tabular} 
& \begin{tabular}{@{}c@{}}Rope Rearrangement;\\Breakfast Preparation;\\Cloth Folding\end{tabular} 
& NR \\

RT-1 (\cite{rt1})
& VLADP 
& NA 
& \begin{tabular}[c]{@{}c@{}}130K Demo.\end{tabular} 
& NA 
& RT-1 Evaluation 
& \begin{tabular}[c]{@{}c@{}}RT-1: 97\%\\Gato: 65\%\end{tabular} 
& \begin{tabular}[c]{@{}c@{}}RT-1: 76\%\\Gato: 52\%\end{tabular} 
& \begin{tabular}[c]{@{}c@{}}Everyday Robots;\\2F Gripper\end{tabular} 
& \begin{tabular}[c]{@{}c@{}}Pick-Place;\\Knock;\\Open-Close\end{tabular} 
& NR \\

Gato (\cite{reed2022generalist})
& VLADP 
& NA 
& \begin{tabular}[c]{@{}c@{}}387K Simulation Demo.;\\15.7K Real Robot Demo.\end{tabular} 
& NA 
& Self-Creation 
& \multicolumn{2}{c}{\centering Gato: 75.6\%} 
& \begin{tabular}[c]{@{}c@{}}Sawyer Robot;\\2F Gripper\end{tabular} 
& Stacking 
& NR \\

RoboCat (\cite{bousmalis2023robocat})
& VLADP 
& NA 
& \begin{tabular}[c]{@{}c@{}}Internet Image Data;\\Real Robot Data;\\Simulation Data\end{tabular} 
& \begin{tabular}[c]{@{}c@{}}1000 Demo.\end{tabular} 
& Self-Creation 
& \multicolumn{2}{c}{\centering\begin{tabular}[c]{@{}c@{}} Fine-tune:\\ RoboCat: 80.5\% \end{tabular} } 
& \begin{tabular}[c]{@{}c@{}}Varies Single-Arm\\(Sawyer; Panda; KUKA);\\2F Gripper;\\3F Dexterous Hand\end{tabular} 
& \begin{tabular}[c]{@{}c@{}}Stack;\\Lift;\\Insert\end{tabular} 
& Imprecise Pose Error \\

RoboAgent (\cite{bharadhwaj2023roboagent})
& VLADP
& NA
& \begin{tabular}[c]{@{}c@{}}7500 Demo.\end{tabular}
& NA
& Self-Creation
& \begin{tabular}[c]{@{}c@{}}RoboAgent: 81.67\% \\ RT-1: 22.5\%\end{tabular}
& \begin{tabular}[c]{@{}c@{}}RoboAgent: 48.25\% \\ RT-1: 5.6\%\end{tabular}
& \begin{tabular}[c]{@{}c@{}}Franka Panda;\\2F Gripper\end{tabular}
& \begin{tabular}[c]{@{}c@{}}Making Tea;\\Cleaning Up;\\Serving Soup;\\Baking Prep;\\Stowing Bowl;\\Heating Soup\end{tabular}
& NR \\

MOO (\cite{stone2023open})
& VLADP
& NA
& \begin{tabular}[c]{@{}c@{}}130K Demo.;\\Augmented Data\end{tabular}
& NA
& RT-1 Evaluation
& \begin{tabular}[c]{@{}c@{}}MOO(111M): 98\% \\ RT-1: $\sim$99\%\end{tabular}
& \begin{tabular}[c]{@{}c@{}}MOO(111M): 79\% \\ RT-1: $\sim$40\%\end{tabular}
& \begin{tabular}[c]{@{}c@{}}Everyday Robot;\\2F Gripper\end{tabular}
& \begin{tabular}[c]{@{}c@{}}Pick-Place;\\Knock;\\Open-Close\end{tabular}
& NR \\

\hline
\end{tabular}
}
\end{threeparttable}
\end{table*}

\begin{table*}[t!]
\Large
\begin{threeparttable}
\resizebox{\textwidth}{!}{
\rowcolors{3}{mygray}{white} 
\begin{tabular}{c c c c c c c c c c c}
\hline
\textbf{Method} 
& \textbf{\begin{tabular}[c]{@{}c@{}}Policy  Type\end{tabular}} 
& \multicolumn{3}{c}{\textbf{Dataset}} 
& \textbf{Benchmark} 
& \multicolumn{2}{c}{\textbf{Success Rate}} 
& \textbf{Embodiment Config} 
& \textbf{Manipulation Task} 
& \textbf{Failure Mode} \\
& 
& \textbf{From-Scratch} 
& \textbf{Pre-Train} 
& \textbf{Fine-Tune} 
& 
& \textbf{Seen} 
& \textbf{UnSeen} 
& 
& 
& \\
\hline

\begin{tabular}{@{}c@{}}Q-Transformer (\cite{q-transformer})\end{tabular} 
& VLADP 
& NA 
& \begin{tabular}[c]{@{}c@{}}115000 \\ Successful Demo.;\\185000 \\ Failed Episodes\end{tabular} 
& NA 
& RT-1 Evaluation
& \multicolumn{2}{c}{\centering \begin{tabular}[c]{@{}c@{}}Q-Trans.: 56\% \\ RT-1: 25\%\end{tabular}} 
& \begin{tabular}[c]{@{}c@{}}Everyday Robot;\\2F Gripper\end{tabular} 
& \begin{tabular}[c]{@{}c@{}}Pick-Place;\\Move;\\Open-Close\end{tabular} 
& NR \\

\begin{tabular}{@{}c@{}}RT-2 (\cite{rt2})\end{tabular} 
& VLADP 
& NA 
& \begin{tabular}[c]{@{}c@{}}Internet \\ Image-Language\\Pairs Data;\\130K Demo.\end{tabular} 
& NA 
& RT-1 Evaluation
& \begin{tabular}[c]{@{}c@{}}RT-2: $\sim$90\% \\ RT-1: $\sim$90\%\end{tabular}
& \begin{tabular}[c]{@{}c@{}}RT-2: 62\% \\ RT-1: 32\%\end{tabular}
& \begin{tabular}[c]{@{}c@{}}Everyday Robot;\\2F Gripper\end{tabular} 
& \begin{tabular}[c]{@{}c@{}}Pick-Place;\\Knock;\\Open-Close\end{tabular} 
& NR \\

\begin{tabular}{@{}c@{}}OpenVLA (\cite{kim2024openvla})\end{tabular} 
& VLADP 
& NA 
& \begin{tabular}[c]{@{}c@{}}970K Demo. \\ from Open-X;\end{tabular} 
& \begin{tabular}[c]{@{}c@{}}50 Demo. per Task\\(Total 10 Tasks) \end{tabular} 
& \begin{tabular}[c]{@{}c@{}}BridgeData V2;\\ RT-1 Evaluation;\\ LIBERO \end{tabular}
& \begin{tabular}[c]{@{}c@{}}Out-of-Box:\\OpenVLA: 88\%;\\RT-1X: 32.0\%;\\RT-2X: 72.0\%;\\Octo: 44.0\%\\ Fine-tune:\\ OpenVLA: 83.7\%; \\ Octo: 83.1\%; \\ DP: 79.7\%\end{tabular}
& \begin{tabular}[c]{@{}c@{}}Out-of-Box:\\OpenVLA: 82.9\%;\\RT-1X: 34.3\%;\\RT-2X: 82.9\%;\\Octo: 14.3\% \\ Fine-tune:\\ OpenVLA: 53.7\%; \\ Octo: 51.1\%; \\ DP: 50.5\%\end{tabular} 
& \begin{tabular}[c]{@{}c@{}}Single-Arm;\\2F Gripper\end{tabular} 
& \begin{tabular}[c]{@{}c@{}}Pick-Place;\\Knock;\\Open-Close;\\Wipe Table\end{tabular} 
& NR \\

\begin{tabular}{@{}c@{}}Octo (\cite{team2024octo})\end{tabular} 
& VLADP 
& NA 
& \begin{tabular}[c]{@{}c@{}}800K Demo. \\ From Open-X;\end{tabular} 
& \begin{tabular}[c]{@{}c@{}}$\sim$100 Demo. per Task \\(Total 6 Tasks)\end{tabular} 
& Self-Creation 
& \multicolumn{2}{p{18.25em}}{\centering \begin{tabular}[c]{@{}c@{}}Out-of-Box:\\Octo: $\sim$70\%;\\RT-1X: $\sim$40\%;\\RT-2X: $\sim$70\% \\ Fine-tune:\\Octo: 72\%\end{tabular}} 
& \begin{tabular}[c]{@{}c@{}}Varies Single-Arm\\(WidowX; BridgeV2; UR5);\\Dual-Arm;\\2F Gripper\end{tabular} 
& \begin{tabular}[c]{@{}c@{}}Pick-Place; \\ Wipe a Table \\ with a Cloth;\\ Open-Close;\\ Handover;\\ Insert\end{tabular} 
& Imprecise Pose Error \\

\begin{tabular}{@{}c@{}}HPT (\cite{wang2024scaling})\end{tabular} 
& VLADP 
& NA 
& \begin{tabular}[c]{@{}c@{}}300K Demo.;\\Simulation Data;\\Human Video Data;\end{tabular} 
& $\sim$50 Traj. 
& Simpler 
& \multicolumn{2}{p{18.25em}}{\centering \begin{tabular}[c]{@{}c@{}}Fine-tune:\\HPT: 46.7\%;\\Octo: 21.7\%\end{tabular}} 
& \begin{tabular}[c]{@{}c@{}}Everyday Robot;\\2F Gripper\end{tabular} 
& \begin{tabular}[c]{@{}c@{}}Pick-Place;\\Open-Close;\\Sweep Leftover;\\Fill Water;\\Switch Insertion;\\Scoop Food\end{tabular} 
& \begin{tabular}[c]{@{}c@{}}Imprecise Pose Error\end{tabular} \\

\begin{tabular}{@{}c@{}}SuSIE (\cite{black2023zero})\end{tabular} 
& VLADP 
& NA 
& \begin{tabular}[c]{@{}c@{}}BridgeData V2\\(60K Demo.);\\Something-Something\\(75K Video Clips)\end{tabular} 
& NA 
& \begin{tabular}[c]{@{}c@{}}BridgeData V2;\\ CALVIN\end{tabular} 
& \begin{tabular}[c]{@{}c@{}}SuSIE: 87\%;\\RT-2X: 43\%;\\MOO: 47\%\end{tabular} 
& \begin{tabular}[c]{@{}c@{}}SuSIE: 69\%;\\RT-2X: 37.5\%;\\MOO: 7.5\%\end{tabular} 
& \begin{tabular}[c]{@{}c@{}}Single-Arm;\\2F Gripper\end{tabular} 
& \begin{tabular}[c]{@{}c@{}}Pick-Place;\\Fold Cloth;\\Open-Close;\\Sweep into Pile\end{tabular} 
& \begin{tabular}[c]{@{}c@{}}Grasp Fail Error;\\Grasp Slip Error\end{tabular} \\

\begin{tabular}{@{}c@{}}GR-1 (\cite{wu2023unleashing})\end{tabular} 
& VLADP 
& NA 
& \begin{tabular}{@{}c@{}}Ego4D Video \\ (8M Frames); \\ 20K Demo.\end{tabular} 
& NA 
& Self-creation 
&  \begin{tabular}{@{}c@{}}GR-1: 84.2\%;\\RT-1: 48.9\%\end{tabular} 
&  \begin{tabular}{@{}c@{}}GR-1: 40.1\%;\\RT-1: 18\%\end{tabular} 
&  \begin{tabular}{@{}c@{}}Single-Arm;\\2F Gripper\end{tabular} 
&   \begin{tabular}{@{}c@{}}Pick-Place;\\ Open-Close;\\ Lift;\\ Turn on/off \end{tabular} 
&  Wrong Object Error \\

\begin{tabular}{@{}c@{}}GR-2 (\cite{cheang2024gr})\end{tabular} 
& VLADP 
& NA 
& \begin{tabular}{@{}c@{}}38M Text-Video Data;\\40K Demo.\end{tabular} 
& NA 
& Self-creation 
&  \begin{tabular}{@{}c@{}}GR-1: $\sim$50\%;\\GR-2: $\sim$80\%\end{tabular} 
&  \begin{tabular}{@{}c@{}}GR-1: $\sim$20\%;\\GR-2: $\sim$70\%\end{tabular} 
&  \begin{tabular}{@{}c@{}}Single-Arm;\\2F Gripper\end{tabular} 
&  Pick-Place 
&  Wrong Object Error \\

\begin{tabular}{@{}c@{}}LAPA (\cite{ye2024latent})\end{tabular} 
& VLADP 
& NA 
& \begin{tabular}{@{}c@{}}3M Traj;\\10K Demo.\end{tabular} 
&  NA 
&  Self-creation 
&  \multicolumn{2}{p{18.25em}}{\centering \begin{tabular}{@{}c@{}} \\LAPA:50.1\%; \\ OpenVLA:43.9\%;\end{tabular}}
&  \begin{tabular}{@{}c@{}}Franka Panda;\\14 DOF bi-manual Robot\end{tabular} 
&  \begin{tabular}{@{}c@{}}Pick-Place;\\Knock Over;\\Cover with Towel\end{tabular} 
&  Imprecise Pose Error \\

\begin{tabular}{@{}c@{}}PI0 (\cite{black2410pi0})\end{tabular} 
& VLADP 
& NA 
&  \begin{tabular}{@{}c@{}}Open-X;\\PI Dataset;\end{tabular} 
& \begin{tabular}{@{}c@{}}5--100 Hours per Task\\(Total 5 Tasks)\end{tabular} 
& Self-creation 
&  \multicolumn{2}{p{18.25em}}{\centering \begin{tabular}{@{}c@{}}Out-of-Box:\\ PI0: $\sim$90\%; \\ OpenVLA: $\sim$35\%;\\ Fine-tune:\\ PI0: $\sim$80\%;\\ DP: $\sim$30\%\end{tabular}}  
&  \begin{tabular}{@{}c@{}}Single-Arm;\\Dual-Arm;\\Mobile-Manipulators\end{tabular} 
&  \begin{tabular}{@{}c@{}}Laundry Folding;\\Clearing a Table;\\Putting Dishes \\ in a Microwave;\\Stacking Eggs \\ into a Carton;\\Assembling a Box;\\Bagging Groceries\end{tabular} 
&   NR \\

\begin{tabular}{@{}c@{}}RDT-1B (\cite{liu2024rdt})\end{tabular} 
& VLADP 
& NA 
&  \begin{tabular}{@{}c@{}}1M+ Demo.\end{tabular} 
&  \begin{tabular}{@{}c@{}}6K+ Demo.\end{tabular} 
&  Self-creation 
&  \multicolumn{2}{p{18.25em}}{\centering \begin{tabular}{@{}c@{}} Fine-tune: \\ RDT-1B: $\sim$70\%;\\ACT: $\sim$10\%;\\OpenVLA: $\sim$2\%;\\Octo: 2\%	\end{tabular}} 
&  \begin{tabular}{@{}c@{}}Mobile ALOHA\end{tabular} 
&  \begin{tabular}{@{}c@{}}Wash Cup;\\Pour Water;\\Handover;\\Fold Shorts;\\Robot Dog\end{tabular} 
&  NR \\

\begin{tabular}{@{}c@{}}Go-1 (\cite{bu2025agibot})\end{tabular} 
& VLADP 
& NA 
& \begin{tabular}{@{}c@{}}Ego4D;\\Open-X;\\AgiBot World\\(140K Traj.)\end{tabular} 
&  NA 
&  Self-creation 
&  \multicolumn{2}{p{18.25em}}{\centering \begin{tabular}{@{}c@{}} Out-of-Box:\\ GO-1: 78\%;\\ RDT-1B: 46\% \end{tabular}} 
&  \begin{tabular}{@{}c@{}}Dual-Arm\end{tabular} 
&  \begin{tabular}{@{}c@{}}Restock Bag;\\Table Bussing;\\Pour Water;\\Restock Beverage;\\Fold Shorts;\\Wipe Table\end{tabular} 
&  NR \\

\begin{tabular}{@{}c@{}}OpenVLA-OFT (\cite{kim2025fine})\end{tabular} 
& VLADP 
& NA 
& \begin{tabular}{@{}c@{}}970K Demo.\\ from Open-X\end{tabular} 
& \begin{tabular}{@{}c@{}}20--300 Demo. per Task\\(Total: 4 Tasks) \end{tabular} 
&  Self-creation 
&  \multicolumn{2}{p{18.25em}}{\centering \begin{tabular}{@{}c@{}} Fine-tune: \\Openvla-oft: 87.8\%;\\	PI0: 83.9\%;\\	RDT-1B: 78.4\%;\\ DP: 77.5\%;\\ Act: 72.3\%	\end{tabular}}
&  \begin{tabular}{@{}c@{}}ALOHA\end{tabular} 
&  \begin{tabular}{@{}c@{}}Fold Shorts;\\Fold Shirt;\\Scoop X into Bowl;\\Put X into Pot\end{tabular} 
&  NR \\

\begin{tabular}{@{}c@{}}Helix (\cite{helix2025})\end{tabular} 
& VLADP 
& NA 
& \begin{tabular}{@{}c@{}}$\sim$500 Hours Demo.\end{tabular} 
& NA 
& NR 
& \multicolumn{2}{p{18.25em}}{\centering \begin{tabular}{@{}c@{}}NR\end{tabular}} 
& \begin{tabular}{@{}c@{}}Figure 02\\Humanoid Robot\end{tabular} 
& \begin{tabular}{@{}c@{}}Pick-Place;\\Open the Refrigerator;\\Organize the Items;\\Handover\end{tabular} 
&  NR \\

\begin{tabular}{@{}c@{}}Groot-N1 (\cite{bjorck2025gr00t})\end{tabular} 
& VLADP 
& NA 
&  \begin{tabular}{@{}c@{}}Real World Datasets;\\Synthetic Dataset;\\Human Video Datasets (592.9M)\end{tabular} 
& \begin{tabular}{@{}c@{}}Human Teleoperation Data  \end{tabular} 
&  Self-creation 
& \multicolumn{2}{p{18.25em}}{\centering \begin{tabular}{@{}c@{}} Fine-tune:\\ GR00T-N1:76.8\%; \\	 DP:46.2\% \end{tabular}}	
&  \begin{tabular}{@{}c@{}}GR-1 Humanoid Robot\end{tabular} 
&  \begin{tabular}{@{}c@{}}Pick-Place;\\Machinery Packing;\\Mesh Cup Pouring;\\Cylinder Handover\end{tabular} 
&  NR \\

\hline
\end{tabular}
}
\begin{tablenotes}
        \footnotesize
        \item[1] "NA" stands for "Not Application."
        \item[2] "NR" stands for "Not Reported."
\end{tablenotes}

\end{threeparttable}
\end{table*}

\begin{table*}[htbp!]
\centering
\captionsetup{justification=justified,singlelinecheck=false}
\caption{\textbf{Representative Policy Works in Sec. \ref{sec:policy}.} ‘Input Modality’ refers to the type of information that is fed into the model. ‘Architecture’ refers to the structural framework employed by the method or the integration of multiple foundational models. ‘Train Objective’ refers to the training loss used by the model. ‘Output Modality’ refers to the type of information produced by the model. ‘Inference Frequency’ refers to the frequency of actions generated by the model. ‘Computational Resources’ refers to the computational power utilized during the training and inference stages. 
}
\label{tab:comparasion_2}
\tiny
\begin{threeparttable}
\rowcolors{3}{mygray}{white}
\begin{tabular}{l c l l c l c l c}
\toprule
\textbf{Method} 
& \textbf{Policy Type}
& \textbf{Input Modality} 
& \textbf{Architecture} 
& \textbf{Train Objective} 
& \textbf{Model Size} 
& \textbf{Output Modality} 
& \textbf{Inference Frequency} 
& \textbf{Computation Resource} \\
\midrule

\begin{tabular}[c]{@{}c@{}}Code as Policy \\(\cite{liang2023code})\end{tabular} 
& VLAC
& Language 
& InstructGPT 
& NA 
& 175B 
& Code 
& NR 
& NA \\

\begin{tabular}[c]{@{}c@{}}Instruct2Act \\(\cite{huang2023instruct2act})\end{tabular} 
& VLAKP
& \begin{tabular}[c]{@{}c@{}}Language; Image\end{tabular}
& SAM; CLIP
& NA
& 995M
& \begin{tabular}[c]{@{}c@{}}Key Poses\end{tabular}
& NR
& \begin{tabular}[c]{@{}c@{}}Inference:\\1 NVIDIA 3090Ti\end{tabular} \\

\begin{tabular}{@{}c@{}}VoxPoser \\(\cite{huang2023instruct2act})\end{tabular} 
& VLAKP
& \begin{tabular}[c]{@{}c@{}}Language; RGBD Image\end{tabular}
& \begin{tabular}[c]{@{}c@{}}OWL-ViT; SAM; XMEM; GPT-4\end{tabular}
& NA
& NR
& \begin{tabular}[c]{@{}c@{}}Key Poses\end{tabular}
& NR
& NR \\

\begin{tabular}{@{}c@{}}ReKep \\(\cite{huang2024rekep}) \end{tabular} 
& VLAKP
& \begin{tabular}[c]{@{}c@{}}Language; RGBD Image\end{tabular}
& \begin{tabular}[c]{@{}c@{}}DINO; SAM; GPT-4O\end{tabular}
& NA
& NR
& \begin{tabular}[c]{@{}c@{}}Key Poses\end{tabular}
& NR
& NR \\

\begin{tabular}{@{}c@{}}CLIPort \\(\cite{shridhar2021cliport}) \end{tabular} 
& VLAKP
& \begin{tabular}[c]{@{}c@{}}Language; RGBD Image\end{tabular}
& Two-Stream Architecture
& Cross Entropy
& NR
& \begin{tabular}[c]{@{}c@{}}Key Poses\end{tabular}
& NR
& \begin{tabular}[c]{@{}c@{}}Inference: \\ 1 NVIDIA P100\end{tabular}  \\

\begin{tabular}{@{}c@{}}PerAct \\(\cite{shridhar2023perceiver})\end{tabular} 
& VLAKP
& \begin{tabular}[c]{@{}c@{}}Language; Voxel Grid\end{tabular}
& \begin{tabular}[c]{@{}c@{}}PerceiverIO Transformer\end{tabular}
& Cross Entropy
& 33.2M
& \begin{tabular}[c]{@{}c@{}}Key Poses\end{tabular}
& NR
& \begin{tabular}[c]{@{}c@{}}Train:\\ 8 NVIDIA V100 * 384h\end{tabular} \\

\begin{tabular}{@{}c@{}}Act3D \\(\cite{gervet2023act3d})\end{tabular} 
& VLAKP
& \begin{tabular}[c]{@{}c@{}}Language; Multi-view Images\end{tabular}
& \begin{tabular}[c]{@{}c@{}}3D Feature Field Transformer\end{tabular}
& \begin{tabular}[c]{@{}c@{}}Cross Entropy; Regression\end{tabular}
& NR
& \begin{tabular}[c]{@{}c@{}}Key Poses\end{tabular}
& NR
& \begin{tabular}[c]{@{}c@{}}Train:\\ 8 NVIDIA 32GB V100 * 120h\end{tabular} \\

\begin{tabular}{@{}c@{}}ChainedDiffuser \\ (\cite{xian2023chaineddiffuser})\end{tabular} 
& VLAKP
& \begin{tabular}[c]{@{}c@{}}Language; Multi-view Images\end{tabular}
& \begin{tabular}[c]{@{}l@{}}3D Feature Field Transformer;\\Diffusion Policy\end{tabular}
& \begin{tabular}[c]{@{}c@{}}Cross Entropy; Diffusion\end{tabular}
& NR
& \begin{tabular}[c]{@{}c@{}}Key Poses\end{tabular}
& NR
& \begin{tabular}[c]{@{}c@{}}Train:\\ 4 NVIDIA A100 * 120h\end{tabular} \\

\begin{tabular}{@{}c@{}}3D Diffuser Actor \\(\cite{ke20243d}) \end{tabular}
& VLAKP
& \begin{tabular}[c]{@{}c@{}}Language; RGBD Image\end{tabular}
& \begin{tabular}[c]{@{}c@{}}Encoders; Diffusion Model\end{tabular}
& \begin{tabular}[c]{@{}c@{}}Cross Entropy; Diffusion\end{tabular}
& NR
& \begin{tabular}[c]{@{}c@{}}Key Poses\end{tabular}
& 1.67 Hz
& \begin{tabular}[c]{@{}c@{}}Inference:\\ 1 NVIDIA 2080 Ti\end{tabular} \\

\begin{tabular}{@{}c@{}}GNFactor \\ (\cite{ze2023gnfactor}) \end{tabular}
& VLAKP
& \begin{tabular}[c]{@{}c@{}}Language; RGBD Image\end{tabular}
& \begin{tabular}[c]{@{}l@{}}Encoders;\\Perceiver Transformer\end{tabular}
& Cross Entropy
& 41.7M
& \begin{tabular}[c]{@{}c@{}}Key Poses\end{tabular}
& NR
& \begin{tabular}[c]{@{}c@{}}Train:\\ 2 NVIDIA RTX3090 * 48h\end{tabular} \\

\begin{tabular}{@{}c@{}}DNAct \\(\cite{yan2024dnact}) \end{tabular}
& VLAKP
& \begin{tabular}[c]{@{}c@{}}Language; Point Cloud\end{tabular}
& \begin{tabular}[c]{@{}c@{}}Encoders; Diffusion Model\end{tabular}
& \begin{tabular}[c]{@{}c@{}}Cross Entropy; Diffusion\end{tabular}
& 11.1M
& \begin{tabular}[c]{@{}c@{}}Key Poses\end{tabular}
& NR
& \begin{tabular}[c]{@{}c@{}}Train:\\ 2 NVIDIA RTX3090 * 12h\end{tabular} \\

\begin{tabular}{@{}c@{}}VoxAct-B \\(\cite{liu2024voxact})\end{tabular}
& VLAKP
& \begin{tabular}[c]{@{}c@{}}Language; RGBD Image\end{tabular}
& \begin{tabular}[c]{@{}l@{}}VLM;\\Perceiver Transformer\end{tabular}
& Cross Entropy
& NR
& \begin{tabular}[c]{@{}c@{}}Key Poses\end{tabular}
& NR
& \begin{tabular}[c]{@{}c@{}}Train:\\ 1 NVIDIA 3080 * 48h\end{tabular} \\

\begin{tabular}{@{}c@{}}LEO \\ (\cite{huang2023embodied})\end{tabular}
& VLAKP
& \begin{tabular}{@{}c@{}}Language; Egocentric Image;\\3D Observation\end{tabular}
& \begin{tabular}{@{}c@{}}Encoders; Vicuna\end{tabular}
& Cross Entropy
& 7B
& \begin{tabular}{@{}c@{}}Key Poses\end{tabular}
& NR
& \begin{tabular}{@{}c@{}}Train:\\ 4 NVIDIA A100\end{tabular} \\

\begin{tabular}{@{}c@{}}Mani\\Foundation \\(\cite{xu2024manifoundation})\end{tabular} 
& VLAKP
& \begin{tabular}{@{}c@{}} Language;\\ Object Point Cloud \\ \& Physical Properties; \\ Hand Point Cloud\end{tabular}
& \begin{tabular}{@{}c@{}}VLM; ManiFoundation\end{tabular}
& Regression
& NR
& \begin{tabular}{@{}c@{}}Contact Points;\\ Force\end{tabular}
& NR
& NR \\

\begin{tabular}{@{}c@{}}RT-1 \\ (\cite{rt1}) \end{tabular} 
& VLADP
& \begin{tabular}{@{}c@{}}Language; Image\end{tabular}
& \begin{tabular}[c]{@{}l@{}}FiLM; Token Learner;\\Transformer Model\end{tabular}
& \begin{tabular}{@{}c@{}}Cross Entropy\end{tabular}
& 35M
& \begin{tabular}{@{}c@{}}Dense Poses\end{tabular}
& \begin{tabular}{@{}c@{}}3HZ\end{tabular}
& NR
\\

\begin{tabular}{@{}c@{}}Gato \\(\cite{reed2022generalist})\end{tabular} 
& VLADP 
& \begin{tabular}{@{}c@{}}Prompt; Image; Proprio.\end{tabular} 
& \begin{tabular}{@{}c@{}}Patch Embedding \\ Transformer Model\end{tabular} 
& \begin{tabular}{@{}c@{}}Cross Entropy\end{tabular} 
& 1.2B 
& \begin{tabular}{@{}c@{}}Dense Poses\end{tabular} 
& \begin{tabular}{@{}c@{}}20HZ\end{tabular} 
& \begin{tabular}{@{}c@{}}Train:\\ 16*16 TPU v3 slice * 96h\\Inference:\\1 NVIDIA RTX3090s\end{tabular} 
\\

\begin{tabular}{@{}c@{}}RoboCat \\ (\cite{bousmalis2023robocat}) \end{tabular} 
& VLADP 
& \begin{tabular}{@{}c@{}}Prompt; Image; Proprio.\end{tabular} 
& \begin{tabular}[c]{@{}l@{}}Patch Embedding \\ Transformer Model;\\VQGAN\end{tabular} 
& \begin{tabular}{@{}c@{}}Cross Entropy\end{tabular} 
& 1.2B 
& \begin{tabular}{@{}c@{}}Dense Poses;\\Hindsight Image\end{tabular} 
& NR 
& NR 
\\

\begin{tabular}{@{}c@{}}RoboAgent \\ (\cite{bharadhwaj2023roboagent})\end{tabular} 
& VLADP
& \begin{tabular}{@{}c@{}}Language; Image; Proprio.\end{tabular}
& \begin{tabular}[c]{@{}l@{}}FiLM; Transformer Encoder;\\Transformer Decoder\end{tabular}
& \begin{tabular}{@{}c@{}}Cross Entropy\end{tabular}
& NR
& \begin{tabular}{@{}c@{}}Dense Poses\end{tabular}
& NR 
&  \begin{tabular}{@{}c@{}}Train:\\ 1 NVIDIA 2080Ti * 48h\end{tabular}
\\

\begin{tabular}{@{}c@{}}MOO \\ (\cite{stone2023open})\end{tabular} 
& VLADP
& \begin{tabular}{@{}c@{}}Language; Mask Image\end{tabular}
& \begin{tabular}[c]{@{}l@{}}FiLM; \\ Token Learner;\\Transformer Model\end{tabular}
& \begin{tabular}{@{}c@{}}Cross Entropy\end{tabular}
& \begin{tabular}{@{}c@{}}111M;\\10.2M;\\2.37M\end{tabular}
& \begin{tabular}{@{}c@{}}Dense Poses\end{tabular}
& NR
& NR
\\

\begin{tabular}{@{}c@{}}Q-Transformer \\ (\cite{q-transformer})\end{tabular} 
& VLADP 
& \begin{tabular}{@{}c@{}}Language; Image\end{tabular} 
& \begin{tabular}[c]{@{}l@{}}FiLM; \\ Transformer Model\end{tabular} 
& \begin{tabular}{@{}c@{}}TD-Learning\end{tabular} 
& NR 
& \begin{tabular}{@{}c@{}}Dense Poses\end{tabular} 
& \begin{tabular}{@{}c@{}}3 Hz\end{tabular} 
& NR
\\

\begin{tabular}{@{}c@{}}RT-2 \\ (\cite{rt2})\end{tabular} 
& VLADP & 
\begin{tabular}{@{}c@{}}Language; Image\end{tabular} & 
\begin{tabular}[c]{@{}l@{}}Encoder; GPT-Style \\ Transformer; Action Head\end{tabular} & 
\begin{tabular}{@{}c@{}}Cross Entropy\end{tabular} & 
\begin{tabular}{@{}c@{}}12B;\\55B\end{tabular} & 
\begin{tabular}{@{}c@{}}Dense Poses\end{tabular} & 
\begin{tabular}{@{}c@{}}1-3HZ (55B);\\5HZ (5B)\end{tabular} & 
\begin{tabular}{@{}c@{}}Inference:\\ Multi-TPU Cloud Service\end{tabular} \\

\begin{tabular}{@{}c@{}}OpenVLA\\ (\cite{kim2024openvla})\end{tabular} 
& VLADP
 &\begin{tabular}[c]{@{}l@{}}Language; Image\end{tabular} & 
\begin{tabular}[c]{@{}l@{}}Encoders; GPT-Style \\ Transformer; \\ Action Head\end{tabular} & 
\begin{tabular}[c]{@{}l@{}}Cross Entropy\end{tabular} & 
7B & 
\begin{tabular}[c]{@{}l@{}}Dense Poses\end{tabular} & 
6Hz & 
\begin{tabular}{@{}c@{}}Train:\\ 64 NVIDIA A100 * 336h; \\ Inference:\\ 1 NVIDIA RTX 4090\end{tabular} \\

\begin{tabular}{@{}c@{}}Octo\\ (\cite{team2024octo})\end{tabular} &  VLADP & 
\begin{tabular}[c]{@{}l@{}}Language; Image; Proprio.\end{tabular} & 
\begin{tabular}[c]{@{}l@{}}Tokenizers;\\ Transformer Backbone; \\ Readout Heads; \\Diffusion Action Head\end{tabular} & 
Diffusion & 
\begin{tabular}[c]{@{}l@{}}10M; \\ 27M; \\ 93M\end{tabular} & 
\begin{tabular}[c]{@{}l@{}}Dense Poses\end{tabular} & 
5--15Hz & 
\begin{tabular}{@{}c@{}}Train:\\ 1 TPU v4-128 pod * 14h; \\ Fine-tune:\\ 1 NVIDIA A5000 * 5h\end{tabular} \\

\begin{tabular}{@{}c@{}}HPT\\ (\cite{wang2024scaling})\end{tabular}
&  VLADP & 
\begin{tabular}[c]{@{}l@{}}Proprio.; Image\end{tabular} & 
\begin{tabular}[c]{@{}l@{}}Varies Tokenization Stems; \\ Transformer Trunk; \\ Varies Action Heads\end{tabular} & 
Regression & 
1B & 
\begin{tabular}[c]{@{}l@{}}Dense Poses\end{tabular} & 
33Hz & 
\begin{tabular}{@{}c@{}}Fine-tune:\\ 1 NVIDIA RTX 2080Ti * 4h; \\ Inference:\\ 1 NVIDIA RTX 3070\end{tabular} \\

\begin{tabular}{@{}c@{}}SuSIE\\ (\cite{black2023zero})\end{tabular} &  VLADP & 
\begin{tabular}[c]{@{}l@{}}Language; Image\end{tabular} & 
\begin{tabular}[c]{@{}l@{}}Goal Generation Model; \\ Goal-Reaching Model\end{tabular} & 
Diffusion & 
NR & 
\begin{tabular}[c]{@{}l@{}}Dense Poses\end{tabular} & 
NR & 
\begin{tabular}{@{}c@{}}Train:\\ 1 v4-64 TPU pod * 17h \\ 1 v4-8 TPU VM * 15h\end{tabular} \\

\begin{tabular}{@{}c@{}}GR-1\\ (\cite{wu2023unleashing})\end{tabular} &  VLADP & 
\begin{tabular}[c]{@{}l@{}}Language; Image; Proprio.\end{tabular} & 
\begin{tabular}[c]{@{}l@{}}Encoders; GPT-style\\ Transformer; \\ Action Head \end{tabular} & 
Regression & 
195M & 
\begin{tabular}{@{}c@{}}Dense Poses\end{tabular} & 
NR & 
NR \\

\begin{tabular}{@{}c@{}}GR-2\\ (\cite{cheang2024gr})\end{tabular} &  VLADP & 
\begin{tabular}[c]{@{}l@{}}Language; Image; Proprio.\end{tabular} & 
\begin{tabular}[c]{@{}l@{}}Encoders; GPT-style\\ Transformer; \\ Action Head \& VQGAN\end{tabular} & 
Regression & 
230M & 
\begin{tabular}{@{}c@{}}Dense Poses; \\ Hindsight Image\end{tabular} & 
NR & 
NR \\

\begin{tabular}{@{}c@{}}LAPA \\ (\cite{ye2024latent})\end{tabular} &  VLADP & 
\begin{tabular}[c]{@{}l@{}}Language; Image\end{tabular} & 
\begin{tabular}[c]{@{}l@{}}Encoders; GPT-style\\ Transformer; \\ Action Head\end{tabular} & 
Cross Entropy & 
7B & 
Dense Poses & 
NR & 
\begin{tabular}{@{}c@{}}Train:\\ 8 NVIDIA H100 * 34h\end{tabular} \\

\begin{tabular}{@{}c@{}}PI0 (\cite{black2410pi0})\end{tabular}  & VLADP & 
\begin{tabular}[c]{@{}l@{}}Language; Image; Proprio.\end{tabular} & 
\begin{tabular}[c]{@{}l@{}}Encoders; GPT-style\\ Transformer; \\ Flow Matching Model\end{tabular} & 
Diffusion & 
3.3B & 
Dense Poses & 
13Hz & 
\begin{tabular}{@{}c@{}}Inference:\\ 1 NVIDIA RTX 4090\end{tabular} \\

\begin{tabular}{@{}c@{}}RDT-1B \\ (\cite{liu2024rdt}) \end{tabular} & VLADP & 
\begin{tabular}[c]{@{}l@{}}Language; Image; Proprio.\end{tabular} & 
\begin{tabular}[c]{@{}l@{}}DiT Block; MLP\end{tabular} & 
Diffusion & 
1B & 
Dense Poses & 
381Hz & 
\begin{tabular}{@{}c@{}}Train:\\ 48 NVIDIA 80GB H100 * 720h;\\ Fine-tune: \\48 NVIDIA 80GB H100 * 72h; \\ Inference:\\ 1 NVIDIA 24GB RTX 4090\end{tabular} \\

\begin{tabular}{@{}c@{}}Go-1 \\ (\cite{bu2025agibot})\end{tabular} & VLADP & 
\begin{tabular}[c]{@{}l@{}}Language; Image\end{tabular} & 
\begin{tabular}[c]{@{}l@{}}Encodes; GPT-style\\ Transformer;\\ Latent Planner; Action Expert\end{tabular} & 
Diffusion & 
2B & 
Dense Poses & 
NR & 
NR \\

\begin{tabular}{@{}c@{}}OpenVLA-OFT\\ (\cite{bu2025agibot})\end{tabular} & VLADP & 
\begin{tabular}[c]{@{}l@{}}Language; Image; Proprio.\end{tabular} & 
\begin{tabular}[c]{@{}l@{}}FiLM; GPT-style\\ Transformer; \\ Action Head\end{tabular} & 
Regression & 
7B & 
Dense Poses & 
77.9Hz & 
\begin{tabular}{@{}c@{}}Fine-tune:\\ 8 NVIDIA 80GB H100; \\ Inference:\\ 1 NVIDIA A100\end{tabular} \\

\begin{tabular}{@{}c@{}}Helix \\ (\cite{helix2025})\end{tabular}  & VLADP & 
\begin{tabular}[c]{@{}l@{}}Language; Image; Proprio.\end{tabular} & 
\begin{tabular}[c]{@{}l@{}}System1 (fast) + System2 (slow)\end{tabular} & 
Regression & 
7B & 
Dense Poses & 
200Hz & 
NR \\

\begin{tabular}{@{}c@{}}Groot-N1 \\ (\cite{bjorck2025gr00t})\end{tabular} & VLADP & 
\begin{tabular}[c]{@{}l@{}}Language; Image; Proprio.\end{tabular} & 
\begin{tabular}[c]{@{}l@{}}System 1 (fast) + System 2 (slow)\end{tabular} & 
Diffusion & 
2.2B & 
Dense Poses & 
120Hz & 
\begin{tabular}{@{}c@{}}Train:\\ 1024 NVIDIA H100 * 49h;\\ Fine-tune:\\ 1 NVIDIA A6000; \\ Inference:\\ 1 NVIDIA L40\end{tabular} \\

\bottomrule
\end{tabular}
\begin{tablenotes}
        \footnotesize
        \item[1] "NA" stands for "Not Application."
        \item[2] "NR" stands for "Not Reported."
      \end{tablenotes}
\end{threeparttable}
\end{table*}

\begin{table*}[htbp!]
\centering
\captionsetup{justification=justified,singlelinecheck=false}
\caption{\textbf{Representative Hierarchy of Skills Works in Sec. 5.} `Instruction Type' refers to the modality of instructions given to the model. `Foundation Models' refers to the type of foundation model used by the method. `Manipulation Task' refers to the task or benchmark designed for each method experiment. `Horizon Steps' refers to the maximum number of steps in the examples provided in the method paper or website.  `Success Rate' refers to the average execution success rate of all tasks for each method and the baseline. `Failure Modes' refers to the common failure cases of the method, as indicated in the paper. There are three main types of Failure Modes: Wrong Object Error: The method fails to recognize different objects. Spatial Relations Error: The method fails to reason about the spatial relations between objects for task planning. Task Order Failure: The method fails to generate the correct temporal order of actions.}
\fontsize{2pt}{2.5pt}\selectfont
\label{tab:comparise_5}
\begin{threeparttable}
\resizebox{\textwidth}{!}{
\rowcolors{3}{mygray}{white} 
\begin{tabular}{c c c c c c c}
\toprule[0.1mm] 
\textbf{Method} 
& \textbf{\begin{tabular}[c]{@{}c@{}}Instruction\\Type\end{tabular}} 
& \textbf{\begin{tabular}[c]{@{}c@{}}Foundation\\Models\end{tabular}} 
& \textbf{\begin{tabular}[c]{@{}c@{}}Manipulation\\Task\end{tabular}} 
& \textbf{\begin{tabular}[c]{@{}c@{}}Horizon\\Steps\end{tabular}} 
& \textbf{\begin{tabular}[c]{@{}c@{}}Success\\Rate\end{tabular}} 
& \textbf{\begin{tabular}[c]{@{}c@{}}Failure\\Modes\end{tabular}} \\
\midrule[0.1mm]

VLaMP (\cite{patel2023pretrained})
& Video 
& GPT-2 
& Cooking; Assembly; etc. 
& 8 
& 47.9\% 
& NR \\

SeeDo (\cite{wang2024vlm})
& Video 
& GPT-4o 
& \begin{tabular}[c]{@{}c@{}}Vegetable Organization; \\ Garments Organization; \\Wooden Block Stacking \end{tabular} 
& 6 
& 36.3\% 
& \begin{tabular}[c]{@{}c@{}}Wrong Object Error;\\ Spatial Relations Error;\\ Task Order Failure \end{tabular} \\

SayCan (\cite{ahn2022can})
& Language 
& PaLM 
& \begin{tabular}[c]{@{}c@{}}Bring Something;\\Throw away Something; etc.  \end{tabular}
& 16 
& 67\% 
& Wrong Object Error; \\

GD (\cite{huang2023grounded})
& Language 
& PaLM / InstructGPT 
& \begin{tabular}[c]{@{}c@{}}Bring Something;\\ Throw away Something; etc. \end{tabular}
& 8 
&  \begin{tabular}[c]{@{}c@{}}GD: 44\%;\\ SayCan: 25\% \end{tabular}
& Task Order Failure \\

TidyBot (\cite{huang2023grounded})
& Language 
& GPT-3 
& Household Organization 
& 10 
& 85\% 
& Wrong Object Error \\

ReAct (\cite{yao2022react})
& Language 
& PaLM 
& ALFWorld 
& more than 50 steps 
& 71\% 
& Task Order Failure \\

LLM-Planner (\cite{song2023llm})
& Language 
& GPT-3 
& ALFRED 
& 7 
&  \begin{tabular}[c]{@{}c@{}}LLM-Planner: 30\%;\\ SayCan: 23.5\% \end{tabular}
& Wrong Object Error \\

VILA (\cite{hu2023look})
& Language 
& GPT-4V 
& \begin{tabular}[c]{@{}c@{}}Stack Plates Steadily; \\  Bring Something; \\ Take Out Something; \\ Prepare Art Class; etc. \end{tabular}
& 6 
& \begin{tabular}[c]{@{}c@{}}VILA: 80\%;\\ GD: 20\%;\\ SayCan: 13\% \end{tabular}
& \begin{tabular}[c]{@{}c@{}} Wrong Object Error;\\ Spatial Relations Error  \end{tabular} \\

LLM+P (\cite{liu2023llm})
& Language 
& GPT-4 
& \begin{tabular}[c]{@{}c@{}}Rearrangement; \\  Create Cocktails; \\ Build Towers; etc. \end{tabular} 
& NR 
& 46\% 
& Task Order Failure \\

\bottomrule[0.1mm]
\end{tabular}
}
 \begin{tablenotes}
        \footnotesize
        \item[1] "NR" stands for "Not Reported."
  \end{tablenotes}
\end{threeparttable}
\end{table*}




\clearpage
\bibliographystyle{SageH}
\bibliography{ref} 
\end{document}